\RequirePackage{fix-cm}
\documentclass[smallextended]{svjour3_no_journal_info}       % onecolumn (second format)
\smartqed  % flush right qed marks, e.g. at end of proof
\usepackage[letterpaper,top=3cm,bottom=3cm,left=3cm,right=3cm,marginparwidth=1.75cm]{geometry}
\usepackage{lipsum}
\usepackage{amsfonts}
\usepackage{graphicx,wrapfig}
\usepackage{epstopdf}
\usepackage{algorithmic}
\ifpdf
  \DeclareGraphicsExtensions{.eps,.pdf,.png,.jpg}
\else
  \DeclareGraphicsExtensions{.eps}
\fi

\usepackage[utf8]{inputenc} % allow utf-8 input
\usepackage[T1]{fontenc}    % use 8-bit T1 fonts
\usepackage{url}            % simple URL typesetting
\usepackage{booktabs}       % professional-quality tables
\usepackage{amsfonts}       % blackboard math symbols
\usepackage{nicefrac}       % compact symbols for 1/2, etc.
\usepackage{microtype}
\usepackage{mathtools}
\usepackage{natbib}
\usepackage{graphicx}
\usepackage{subcaption}
\usepackage{amsmath,amssymb,amsfonts,amsxtra,bm}
\usepackage{enumerate}
\usepackage{enumitem}
\usepackage{hyperref}       % hyperlinks
\usepackage{tcolorbox}
\usepackage{algorithm,algorithmic}
\usepackage{booktabs, 
            makecell, multirow, tabularx} 

\clearpage{}%
\newcommand{\Xc}{\mathcal{X}}

\newcommand{\Cc}{\mathcal{C}}

\newcommand{\Sc}{\mathcal{S}}
\newcommand{\Zc}{\mathcal{Z}}

\newcommand{\Nc}{\mathcal{N}}
\newcommand{\Uc}{\mathcal{U}}
\newcommand{\Oc}{\mathcal{O}}
\newcommand{\gr}{\nabla}
\newcommand{\E}{\mathbb{E}}

\newcommand{\reals}{\mathbb{R}}

\newcommand{\ip}[2]{\langle {#1},\, {#2} \rangle}
\newcommand{\norm}[1]{\| {#1} \|}
\newcommand{\lnorm}[1]{\left\| {#1} \right\|}

\DeclareMathOperator{\Proj}{Proj}

\DeclareMathOperator*{\argmin}{argmin}

\newtheorem{assumption}{Assumption}
\makeatletter
\newcommand{\leqnomode}{\tagsleft@true}
\newcommand{\reqnomode}{\tagsleft@false}

\newcommand{\blue}[1]{{\color{black}#1}}

\title{On Penalty-based Bilevel Gradient Descent Method
\thanks{This work was presented in part at International Conference on Machine Learning (ICML) 2023 \citep{shen2023penalty}.\\ The work of Q. Xiao, H. Shen, and T. Chen was   supported by National
Science Foundation (NSF) MoDL-SCALE project 2134168, NSF project 2412486, the Cisco Research Award, and an Amazon Research Award.}
}

\titlerunning{On Penalty-based Bilevel Gradient Descent Method}        % if too long for running head

\author{
        Han Shen\and
        Quan Xiao \and
 Tianyi Chen%etc.
}

\authorrunning{Han Shen, Quan Xiao, Tianyi Chen} % if too long for running head

\institute{Han Shen, Quan Xiao \and Tianyi Chen \at
                Department of Electrical, Computer, and Systems Engineering\\
              Rensselaer Polytechnic Institute \\
              \email{shenhanhs@gmail.com; quanx1808@gmail.com; chentianyi19@gmail.com}           %  \\
}
 
\date{Received: August 29, 2023 / Accepted: December 21, 2024}

\begin{document}

\maketitle

% REQUIRED
\begin{abstract}
Bilevel optimization enjoys a wide range of applications in emerging machine learning and signal processing problems such as hyper-parameter optimization, image reconstruction, meta-learning, adversarial training, and reinforcement learning. However, bilevel optimization problems are traditionally known to be difficult to solve. Recent progress on bilevel algorithms mainly focuses on bilevel optimization problems through the lens of the implicit-gradient method, where the lower-level objective is either strongly convex or unconstrained. In this work, we tackle a challenging class of bilevel problems through the lens of the penalty method. We show that under certain conditions, the penalty reformulation recovers the (local) solutions of the original bilevel problem. Further, we propose the penalty-based bilevel gradient descent (PBGD) algorithm and establish its finite-time convergence for the constrained bilevel problem with lower-level constraints yet without lower-level strong convexity. Experiments on synthetic and real datasets showcase the efficiency of the proposed PBGD algorithm. The code for implementing this algorithm is publicly available on \href{https://github.com/hanshen95/penalized-bilevel-gradient-descent}{GitHub}.

\end{abstract}

\section{Introduction}

Bilevel optimization plays an increasingly important role in   machine learning \citep{liu2021investigating}, image processing \citep{crockett2022bilevel} and communications \citep{chen2023learning}. Specifically, in machine learning, it has a wide range of applications including hyper-parameter optimization \citep{maclaurin2015gradient,franceschi2018bilevel},   meta-learning \citep{finn2017maml,rajeswaran2019meta}, reinforcement learning \citep{cheng2022adversarial} and adversarial learning \citep{jiang2021learning}. 

Define $f:\mathbb{R}^{d_x} \times \reals^{d_y}\mapsto \reals$ and $g:\mathbb{R}^{d_x} \times \reals^{d_y}\mapsto \reals$. We consider the following bilevel  problem:
\begin{align}
    \mathcal{BP}: ~~~~\min_{x,y}f(x,y)~~\,\,\,\,\,{\rm s.t.}\,\,&x\in \Cc,~y \in \Sc(x)\coloneqq \arg\min_{y\in\Uc(x)}g(x,y) \nonumber
\end{align}
where $\Cc\subseteq \mathbb{R}^{d_x}$ is a non-empty and closed set, and $\Uc(x), \Sc(x)$  are non-empty and closed sets given any $x\in \Cc$. We call $f$ and $g$ respectively as the upper-level and lower-level objectives.

\begin{wrapfigure}{R}{0.5\textwidth}
    % \vspace*{-0.2cm}
    \centering
    \includegraphics[width=.4\textwidth]{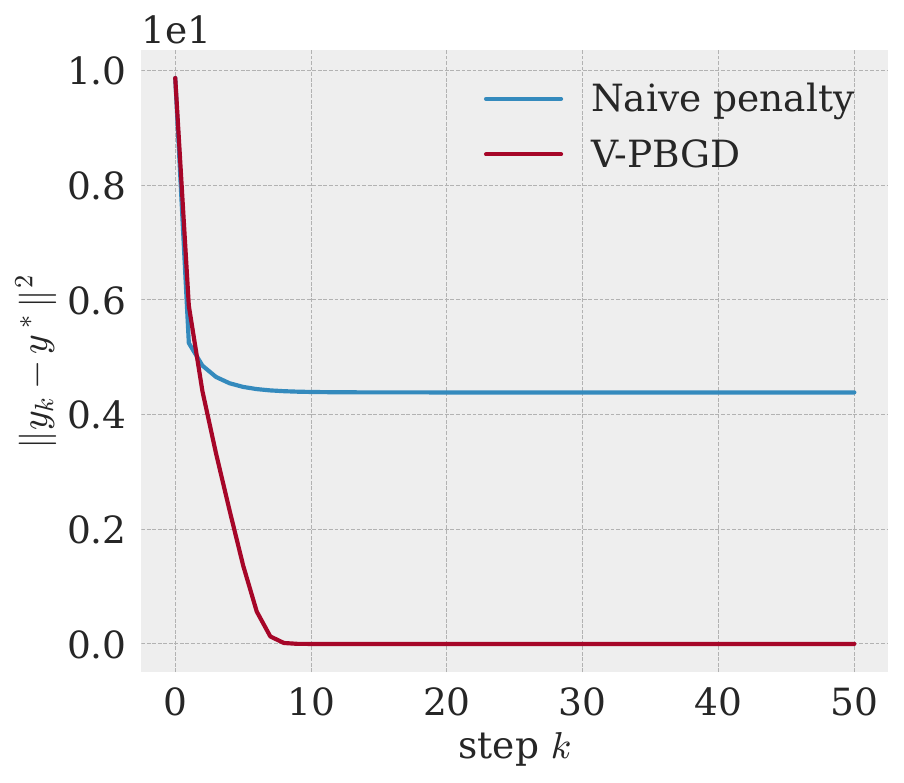}
    \vspace*{-0.2cm}
    \caption{"Naive" penalty yields suboptimal points while the proposed algorithm finds the solution.}
    \label{fig:intro toy}
 \vspace*{-0.4cm}
\end{wrapfigure}

The bilevel optimization problem $\mathcal{BP}$ can be difficult to solve due to the coupling between the upper-level and lower-level problems through the solution set $\Sc(x)$. Even for the simpler case where $g(x,\cdot)$ is strongly convex, and the upper- and lower-level problems are unconstrained, i.e., $\Uc(x)=\reals^{d_y}$ and $\Cc=\reals^{d_x}$, it was not until recently that the iteration and sample complexity of solving this problem has become partially understood. Under the strong convexity of $g(x,\cdot)$, the lower-level solution set $\Sc(x)$ is a singleton. In this case, $\mathcal{BP}$ reduces to minimizing an implicitly defined objective function $f(x,\Sc(x))$, the gradient of which can be calculated with the implicit function theory \citep{dontchev2009implicit} or the implicit gradient (IG) method \citep{pedregosa2016hyper,ghadimi2018approximation}. 
It has been later shown by \citep{chen2021tighter} that the stochastic IG method converges almost as fast as the (stochastic) gradient-descent method. 
However, existing IG methods cannot handle the non-strong convexity of $g(x,\cdot)$ due to the lack of implicit gradients and thus can not be applied to complicated bilevel problems. 

To overcome the above challenges, recent work aims to develop gradient-based methods for bilevel problems without lower-level strong convexity. A prominent branch of algorithms are based on the iterative differentiation method; see e.g., \citep{franceschi2017forward,liu2021towards}. In this case, the lower-level solution set $\Sc(x)$ is replaced by the output of an iterative optimization algorithm with a differentiable update rule that solves the lower-level problem (e.g., gradient descent (GD)) which allows for explicit differentiation over the entire optimization trajectory. 
However, these methods are typically restricted to the unconstrained case since the lower-level algorithm with the projection operator is difficult to differentiate.
% Though not restricted by strong-convexity, these methods generally lack finite-time convergence guarantee without restrictive assumptions on the iterative mapping. 
Furthermore, the algorithm usually has high memory and computational costs when the number of lower-level iterations is large.

On the other hand, it is tempting to penalize certain optimality metrics of the lower-level problem (e.g., $\|\nabla_y g(x,y)\|^2$) to the upper-level objective, leading to a single-level optimization problem. The high-level idea is that minimizing the optimality metric guarantees the lower-level optimality $y\in\Sc(x)$ and as long as the optimality metric admits simple gradient evaluation, the penalized objective can be optimized via gradient-based algorithms. However, as we will show in the next example, GD for a straightforward penalization mechanism may not lead to the desired solution of the original problems.

% \vspace{-0.2cm}
\begin{example}\label{exp:intro}
Consider the following special case of $\mathcal{BP}$ with only one variable $y$ in both levels:
\vspace{-0.1cm}
\begin{align}\label{eq:exp intro}
    &\min_{x,y\in\reals}~ f(x,y)~\coloneqq \sin^2(y-\frac{2\pi}{3})\,\,\,\,\,\,{\rm s.t.}\,\,\,\,~~y\in \arg\min_{y\in\reals} g(x,y)\coloneqq y^2+2\sin^2 y.
\end{align}
The only solution of \eqref{eq:exp intro} is $y^*\!=\!0$.
In this example, it can be checked that $(\nabla_y g(x,y))^2=(2y+2\sin(2y))^2=0$ if and only if $y\in\arg\min_{y\in\reals} g(x,y)$ and thus $(\nabla_y g(x,y))^2=0$ is a lower-level optimality metric.
Penalizing $f(x,y)$ with the penalty function $(\nabla_y g(x,y))^2$ and a penalty constant $\gamma>0$ gives $\min f(x,y)+4\gamma (y+\sin(2y))^2$. For any $\gamma$, $y=\frac{2\pi}{3}$ is a local solution of the penalized problem, but it is neither a global solution nor a local solution of the original problem \eqref{eq:exp intro}.
\end{example}

In Figure \ref{fig:intro toy}, we show that, for {Example \ref{exp:intro}}, the ``naive" penalty method, which solves $\min f(x,y)+\gamma (y+\sin(2y))^2$ via gradient descent, can get stuck at some sub-optimal points, but our V-PBGD method (introduced in this paper) successfully converges to the optimum.
To tackle such issues, it is crucial to study the relation between the bilevel problem and its penalized problem. Specifically, \emph{what impact do different penalty terms, penalization constants and problem properties have on this relation?} Through studying this relation, we aim to develop an efficient penalty-based bilevel GD method for $\mathcal{BP}$.

\textbf{Our contributions.} In this work, we consider the following penalty reformulation of $\mathcal{BP}$, given by
\begin{equation}
    \mathcal{BP}_{\gamma p}:~\min_{x,y}f(x,y)+\gamma p(x,y),~~~~~{\rm s.t.}~~~~~x\in\Cc,~y\in\Uc(x) \nonumber
\end{equation}
where $p(x,y)$ is a certain penalty term that will be specified in Section \ref{sec:NP}.
Our first result shows that under certain generic conditions on $p(x,y)$, one can recover approximate global (local) solutions of $\mathcal{BP}$ by solving $\mathcal{BP}_{\gamma p}$ globally (locally). Further, we show that these generic conditions hold without the strong convexity or even convexity of $g(x,\cdot)$. Then we propose a penalized bilevel GD (PBGD) method, extend to its stochastic version, and then establish its finite-time convergence when the lower-level is unconstrained (e.g., $\Uc(x)\!=\!\reals^{d_y}$) in Section \ref{sec:NP}. Building upon the results in the unconstrained setting, we extend the algorithm and its analysis to the more challenging bilevel problems, where the constraint set $\Uc(x)\!=\!\Uc$ is a compact convex set in Section \ref{sec:CP}, and where the penalty function is nonsmooth in Section \ref{sec:nonsmooth penalization}. We summarize the convergence results of our algorithm in Table \ref{tab:table2} and compare them with several related works in Table \ref{tab:table1}. Finally, we showcase the performance, computation and memory efficiency of the proposed algorithm in comparison with several competitive baselines in Section \ref{sec:simulation}.
\begin{table*}[t]
\fontsize{10}{9}\selectfont
\centering
%    \vspace{0.2cm}
    \begin{tabular}{c|c|c|  c}
    \hline
    \hline
    &Section \ref{sec:NP}  & Section \ref{sec:CP}  & Section \ref{sec:nonsmooth penalization} \\
    \hline
    Upper-level constraint &\multicolumn{3}{c}{$\Cc$ closed and convex} \\
    \hline
    Lower-level constraint & $\Uc(x)=\reals^{d_y}$ &$\Uc(x)=\Uc$ is closed convex&$\Uc(x)=\reals^{d_y}$\\
    \hline
    $\text{Assumption}^\dag$ on $f$& \multicolumn{3}{c}{$f(x,\cdot)$ is Lipschitz-continuous} \\
    \hline
    $\text{Assumption}^\dag$ on $g(x,\cdot)$&$\text{PL}^\ddag$ &$\text{QG+convex/EB}^\ddag$ & $\text{PL}^\ddag,~\|\nabla_y g(x,\cdot)\|$ convex\\
    \hline
    Constant $\gamma$&$\Omega(\epsilon^{-0.5})$&$\Omega(\epsilon^{-0.5})$&$\Omega(1)$\\
    \hline
    Iteration complexity &$\Oc(\epsilon^{-1.5})$&$\Oc(\epsilon^{-1.5})$&$\Oc(\epsilon^{-1})$\\
    \hline
    \hline
    \end{tabular}
%    \vspace{-0.2cm}
    \caption{Comparison of the main theorems in this paper. Here $\epsilon$ is the accuracy. $^\dag$The Lipschitz-smooth assumption on $f,g$ are neglected in this table; $^\ddag$PL short for Polyak-Łojasiewicz function, QG short for quadratic growth, EB short for error bound. These notions will be defined in each section.} 
    \label{tab:table2}
%     \vspace{-0.2cm}
\end{table*}
\begin{table*}[t]
\fontsize{10}{9}\selectfont
\centering
%    \hspace{-0.2cm}
    \begin{tabular}{c|c|c|c|  c}
    \hline
    \hline
    &{V-PBGD} (\textbf{ours})   & BOME  & IAPTT-GM & AiPOD \\
    \hline
    Upper-level constraint&Yes&No&Yes& Yes\\
    \hline
    Lower-level   constraint& \!\! $\Uc(x)\!=\!\reals^{d_x} \text{ or }\Uc$ \!\! &No&$\Uc(x)\!=\!\Uc$&  \!\! equality constraint \!\! \\
    \hline
   \!\!\!\!  Lower-level non-strongly-convex   \!\!\!\! &Yes&Yes&Yes& No \\
    \hline
    Non-singleton $\Sc(x)$&Yes&No&Yes& No\\
    \hline
    First-order &Yes&Yes&No& No\\
    \hline
    Convergence & finite-time  & finite-time & asymptotic& finite-time\\
    \hline
    \hline
    \end{tabular}
%    \vspace{-0.2cm}
    \caption{Comparison of this work (V-PBGD) and IAPTT-GM \citep{liu2021towards}, BOME \citep{ye2022bome}, AiPOD \citep{xiao2022alternating}. 
    In the table, $\Uc$ is a convex compact set. } 
    \label{tab:table1}
     \vspace{0.2cm}
\end{table*}

\subsection{Related works}
The study of bilevel optimization problems can be dated back to the early work on game theory   \citep{stackelberg}. 
Since its introduction, it has inspired a rich literature \citep{colson2007overview,vicente1994bilevel,vicente1994descent,falk1995bilevel,luo1996mathematical}. 
Since the seminal work \citep{bennett2006model}, the gradient-based bilevel optimization methods have gained growing popularity in the machine learning area; see, e.g., \citep{sabach2017jopt,franceschi2018bilevel,liu2020icml}. 
 Many gradient-based methods belong to the class of IG methods \citep{pedregosa2016hyper}. The finite-time convergence was first established in \citep{ghadimi2018approximation} for the unconstrained strongly-convex lower-level problem. Later, the convergence was improved in \citep{hong2020ac,ji2021bilevel,chen2022stable,chen2021tighter,khanduri2021near,shen2022single, Li2022fully,sow2022convergence}. Recent works extend IG to constrained strongly-convex lower-level problems; see, e.g., the equality-constrained IG method \citep{xiao2022alternating} and a 2nd-derivative-free approach \citep{giovannelli2022inexact}; and lower-level problems satisfying the Polyak-Lojasiewicz (PL) condition \citep{xiao2023}.
 
 Another branch of methods are based on the iterative differentiation (ITD) methods \citep{maclaurin2015gradient,franceschi2017forward,nichol2018firstorder,shaban2019truncated}. 
Later, \citep{liu2021towards} proposes an ITD method with initialization optimization and shows asymptotic convergence.
Another work \citep{liu2022general} develops an ITD method where each lower-level iteration uses a combination of upper-level and lower-level gradients. 
Recently, the iterative differentiation of non-smooth lower-level algorithms has been studied in \citep{bolte2022automatic}.
 The ITD methods generally lack finite-time guarantee unless restrictive assumptions are made for the iterative update \citep{grazzi2020iteration,ji2022will}.  
 
Recently, bilevel optimization methods have also been studied in composition optimization \citep{wang2017stochastic}, distributed learning \citep{tarzanagh2022fednest,lu2022decentralized,yang2022decentralized,chen2023decentralized}, corset selection \citep{zhou2022probabilistic}, overparametrized setting \citep{vicol2022implicit}, multi-block min-max \citep{hu2022multi}, and game theory \citep{arbel2022non}. Several acceleration methods have been proposed to improve the complexity \citep{khanduri2021near,yang2021provably,huang2021enhanced,dagreou2022framework}. The works \citep{liu2021value} and \citep{mehra2021penalty} propose penalty-based methods respectively with log-barrier and gradient norm penalty, and establish their asymptotic convergence. Other works \citep{gao2022value,ye2023difference} develop methods based on the difference-of-convex algorithm. In preparing our final version, a concurrent work \citep{chen2023bilevel} studies the bilevel problem with convex lower-level objectives and proposes a zeroth-order  optimization method with finite-time convergence to the Goldstein stationary point.
Another concurrent work \citep{lu2023first} proposes a penalty method for the bilevel problem with a convex lower-level objective $g(x,\cdot)$. It shows convergence to a weak Karush–Kuhn–Tucker (KKT) point of the bilevel problem while does not study the relation between the bilevel problem and its penalized problem.

The relation between the bilevel problem and its penalty reformulation has been first studied in the seminal work \citep{jane1997exact} under the calmness condition paired with other conditions such as the 2-Hölder continuity, which may be difficult to satisfy. A recent work \citep{ye2022bome} proposes a novel first-order method that is termed BOME. By assuming the constant rank constraint qualification (CRCQ), \citep{ye2022bome} shows convergence of BOME to a KKT point of the bilevel problem. However, it is unclear when CRCQ can be satisfied and the convergence relies on restrictive assumptions like the uniform boundedness of $\|\nabla g\|$, $\|\nabla f\|$, $|f|$ and $|g|$. It is also difficult to argue when the KKT point is a solution of the bilevel problem under lower-level non-convexity. Besides, based on the exact penalty theorem \citep{clarke1990optimization}, deriving the necessary condition for the optimality of bilevel problem \citep{ye1995optimality,ye2010new,dempe2012karush,dempe2006optimality,ye2020constraint} is also related to our work, which is of independent interest in the optimization community.   
% In this work, we focus on the PL condition that grows prominent in machine learning and other standard conditions in gradient-based optimization.

\textbf{Notations.}
We use $\|\cdot\|$ to denote the $l^2$-norm.
Given $r>0$ and $z\in\reals^d$, define $\Nc(z,r)\coloneqq \{z'\in\reals^{d}:\|z-z'\|\leq r\}$. Given vectors $x$ and $y$, we use $(x,y)$ to indicate the concatenated vector of $x,y$. 
Given a non-empty closed set $\Sc \subseteq \reals^{d}$, define the distance of $y\in\reals^{d}$ to the set $\Sc$ as
$d_{\Sc}(y)\coloneqq \min_{y' \in \Sc}\|y-y'\|$. We use $\Proj_{\Zc}$ to denote the projection to the set $\Zc$.

\section{Penalty Reformulation of Bilevel Problems}\label{sec:generic}
This section studies the relation between the solutions of the bilevel problem $\mathcal{BP}$ and those of its penalty reformulation $\mathcal{BP}_{\gamma p}$ by posting certain generic conditions on the penalty function $p(x,y)$. 

Since $\Sc(x)$ is closed, $y\in\Sc(x)$ is equivalent to $d_{\Sc(x)}(y)=0$. We therefore rewrite $\mathcal{BP}$ as
\begin{align}\label{eq:BP rewrite}
    \min_{x,y}f(x,y)\,\,\,\,\,{\rm s.t.}\,\,&x\in \Cc,~d^2_{\Sc(x)}(y)=0. 
\end{align}
The squared distance $d^2_{\Sc(x)}(y)$ is non-differentiable, and thus penalizing it to the upper-level objective is computationally intractable. Instead, we consider its upper bounds defined as follows.
\begin{definition}[Squared-distance bound function]\label{def:SDB}
A function $p:\reals^{d_x}\times \reals^{d_y}\mapsto \reals$ is a $\rho$-squared-distance-bound if there exists $\rho>0$ such that for any $x\in \Cc,y\in \Uc(x)$, it holds 
\begin{subequations}
    \begin{align}
    &p(x,y)\geq 0,~\rho p(x,y) \geq d_{\Sc(x)}^2(y)\\
    &p(x,y)=0~\text{ if and only if }~d_{\Sc(x)}(y)=0.
\end{align}
\end{subequations}
\end{definition}

Suppose $p(x,y)$ is a squared-distance bound function. {\color{black} Given $\epsilon\!>\!0$, we define the following problem:} 
\begin{align}
    \mathcal{BP}_\epsilon:~\min_{x,y}f(x,y)\,\,\,\,\,{\rm s.t.}\,\,&x\in \Cc,~y\in\Uc(x),~p(x,y)\leq \epsilon.\nonumber
\end{align}
It is clear that the constrained problem $\mathcal{BP}_\epsilon$ above with $\epsilon=0$ recovers the original bilevel problem $\mathcal{BP}$.

For $\epsilon>0$, we will show that $\mathcal{BP}_\epsilon$ is an $\epsilon$-approximate problem of the original bilevel problem $\mathcal{BP}$, which is rigorously defined as follows.
 
\begin{definition}[An $\epsilon$-approximate problem]\label{def.epsilon_app}
    We say the problem $\mathcal{BP}_\epsilon$ is an $\epsilon$-approximate problem of the bilevel problem $\mathcal{BP}$ if the following two conditions are met: 
    \begin{enumerate}
    	\item [i)] any feasible solution $(x,y)$ of $\mathcal{BP}_\epsilon$ satisfies $d_{\Sc(x)}^2(y)\leq \rho\epsilon$ and $x\in\Cc,~y\in\Uc(x)$; and, 
    	\item [ii)] $|f^*-f_\epsilon^*|\leq L\sqrt{\rho\epsilon}$, where $f^*$ and $f_\epsilon^*$ are respectively the optimal objective values of $\mathcal{BP}$ and $\mathcal{BP}_\epsilon$.
    \end{enumerate}
\end{definition}

% since $p(x,y)$ is an upperbound of $d_{\Sc(x)}^2(y)$.
% We start by considering the relation among the global solutions of $\mathcal{BP}_\epsilon$ and $\mathcal{BP}_{\gamma p}$.

Before we introduce the result, we additionally give the following definition and assumption.
\begin{definition}[Lipschitz continuity]\label{def:Lipschitz continuity}
Given $L>0$, a function $\ell:\reals^{d}\mapsto \reals^{d'}$ is said to be $L$-Lipschitz continuous on $\Zc\subseteq \reals^d$ if it holds for any $z,z'\in \Zc$ that
$\|\ell(z)-\ell(z')\| \leq L\|z-z'\|$. A function $\ell$ is said to be $L$-Lipschitz-smooth if its gradient is $L$-Lipschitz continuous. 
\end{definition}

\begin{assumption}\label{asp:Lipschitz continuity}
 There exists $L\!>\!0$ that given any $x\in \Cc$, $f(x,\cdot)$ is $L$-Lipschitz continuous on $\Uc(x)$.
\end{assumption}
The above assumption is standard and has been made in several other works studying bilevel optimization; see, e.g., \citep{ghadimi2018approximation,chen2021tighter,chen2023bilevel}. 
With this assumption, we can connect $\mathcal{BP}_\epsilon$ and $\mathcal{BP}$ in the following lemma.
\begin{lemma}[Relation between $\mathcal{BP}_\epsilon$ and $\mathcal{BP}$]\label{lem:epsilon approx}
Assume $p(x,y)$ in $\mathcal{BP}_\epsilon$ is a $\rho$-squared-distance-bound function and Assumption \ref{asp:Lipschitz continuity} holds, then $\mathcal{BP}_\epsilon$ is an $\epsilon$-approximate problem of $\mathcal{BP}$.
\end{lemma}
\begin{proof}
Since $p(x,y)$ is a $\rho$-squared-distance bound, it is immediate that any feasible solution of $\mathcal{BP}_\epsilon$ satisfies $x\in\Cc,y\in\Uc(x)$ and $d_{\Sc(x)}(y)^2\leq \rho \epsilon$.

Next we prove $|f^*-f_\epsilon^*|\leq L\sqrt{\rho \epsilon}$. Let $(x_\epsilon,y_\epsilon)$ be a global solution of $\mathcal{BP}_\epsilon$ with $f_\epsilon^*=f(x_\epsilon,y_\epsilon)$. Since $\Sc(x_\epsilon)$ is non-empty and closed, one can find $\bar{y}_\epsilon\in\Sc(x_\epsilon)$ such that $d_{\Sc(x_\epsilon)}(y_\epsilon)=\|\Bar{y}_\epsilon-y_\epsilon\|$. Then it holds  
    \begin{align}\label{eq:idk100}
        f(x_\epsilon,\Bar{y}_\epsilon)-f(x_\epsilon,y_\epsilon)
        &\leq L\|y_\epsilon-\bar{y}_\epsilon\|=L d_{\Sc(x_\epsilon)}(y_\epsilon) \leq L\sqrt{\rho \epsilon}
    \end{align}
where the last inequality follows as $p(x,y)$ is a $\rho$-squared-distance bound and thus $d_{\Sc(x_\epsilon)}(y_\epsilon)^2\leq \rho \epsilon$. 
Let $(x^*,y^*)$ be a global solution of $\mathcal{BP}$ so that $f^*=f(x^*,y^*)$.
Since $(x_\epsilon,\bar{y}_\epsilon)$ is feasible for $\mathcal{BP}$, we have $f(x_\epsilon,\bar{y}_\epsilon) \geq f^*$. Since $(x^*,y^*)$ is feasible for $\mathcal{BP}_{\epsilon}$, we have $f_\epsilon^*=f(x_\epsilon,y_\epsilon) \leq f^*$. The inequalities  $f(x_\epsilon,\bar{y}_\epsilon) \geq f^*$, $f(x_\epsilon,y_\epsilon) \leq f^*$ and \eqref{eq:idk100} imply $|f^*-f_\epsilon^*|\leq L\sqrt{\rho \epsilon}$, which justifies the definition of an $\epsilon$-approximate problem in Definition \ref{def.epsilon_app}.
\qed
\end{proof}

Next, we consider the relation between global solutions of the penalized problem $\mathcal{BP}_{\gamma p}$ and those of the original bilevel problem $\mathcal{BP}$. 
We first introduce the definition of an $\epsilon$-global-minimum point.
\begin{definition}[$\epsilon$-global-minimum]\label{def_approx_global}
    Given a feasible set $\Zc\subseteq \reals^d$ and a function  $\ell:\reals^d \mapsto \reals$, for the constrained problem defined as
    \begin{equation}
        \min \ell(z),~~~\,\,\,{\rm s.t.}~~~\,\,\,z\in\Zc,\nonumber
    \end{equation}
    we say a point $\hat{z}\in\Zc$ is an $\epsilon$-global-minimum point of this problem if $\ell(\hat{z})\leq \ell(z)+\epsilon$ for any $z\in\Zc$.
\end{definition}
 
To establish the relation between global solutions, a crucial step is to guarantee that a solution of $\mathcal{BP}_{\gamma p}$ denoted as $(x_\gamma,y_\gamma)$ is feasible for $\mathcal{BP}_\epsilon$, e.g., $p(x_\gamma,y_\gamma)$ is small. Under Assumption \ref{asp:Lipschitz continuity}, the growth of $f(x_\gamma,\cdot)$ is controlled. Then an important intuition is that increasing $\gamma$ in $\mathcal{BP}_{\gamma p}$ likely makes $p(x_\gamma,\cdot)$ more dominant, and thus decreases $p(x_\gamma,y_\gamma)$.
With this intuition, we introduce the theorem as follows.

\begin{theorem}[Relation on global solutions]\label{the:BP_gamr_epsilon}
Assume $p(x,y)$ is a $\rho$-squared-distance-bound function and Assumption \ref{asp:Lipschitz continuity} holds with $L>0$. Given $\epsilon_1>0$, any global solution of $\mathcal{BP}$ is an $\epsilon_1$-global-minimum point of $\mathcal{BP}_{\gamma p}$ with any $\gamma \geq \gamma^*=\frac{L^2 \rho}{4}\epsilon_1^{-1}$. Conversely, given $\epsilon_2\geq0$, if $(x_\gamma,y_\gamma)$ achieves $\epsilon_2$-global-minimum of $\mathcal{BP}_{\gamma p}$ with $\gamma > \gamma^*$, $(x_\gamma,y_\gamma)$ is $\epsilon_2$-global-minimum of $\mathcal{BP}_{\epsilon_\gamma}$ with some $\epsilon_\gamma \leq (\epsilon_1+\epsilon_2)/(\gamma\!-\!\gamma^*)$.
\end{theorem}
\begin{proof}
First we prove from $\mathcal{BP}$ to $\mathcal{BP}_{\gamma p}$.
Given any $x\in \Cc$ and $y\in \Uc(x)$, since $\Sc(x)$ is closed and non-empty, we can find the projection of $y$ on $\Sc(x)$ as $y_x\in\arg\min_{y'\in\Sc(x)}\|y'-y\|$; that is $d_{\Sc(x)}(y)=\|y_x-y\|$. By Lipschitz continuity assumption on $f(x,\cdot)$, given any $ x\in \Cc$, it holds for any $y\in\Uc(x)$ that
\begin{align}
    f(x,y)-f(x,y_x) &\geq -L d_{\Sc(x)}(y)\quad \text{by }d_{\Sc(x)}(y)=\|y_x-y\|.\nonumber
\end{align}
Then it follows that
\begin{align}\label{eq:idk2}
    f(x,y)+\gamma^* p(x,y)-f(x,y_x) &\geq -L d_{\Sc(x)}(y) + \gamma^* p(x,y) \nonumber\\
    &\geq -L d_{\Sc(x)}(y) + \frac{\gamma^*}{\rho} d_{\Sc(x)}^2(y) \nonumber\\
    &\geq \min_{z\in\mathbb{R}_{\geq 0}}-Lz + \frac{\gamma^*}{\rho} z^2 = -\epsilon_1 \quad\text{with }\gamma^*=\frac{L^2 \rho}{4}\epsilon_1^{-1}.
\end{align}
Since $y_x \in \Sc(x)$ (thus $y_x\in \Uc(x)$) and $x\in \Cc$, we have that $(x,y_x)$ is feasible for $\mathcal{BP}$. Let $f^*$ be the optimal objective value for $\mathcal{BP}$, we know $f(x,y_x) \geq f^*$. This along with \eqref{eq:idk2} indicates
\begin{align}\label{eq:global calmness 1}
    f(x,y)+\gamma^* p(x,y)-f^*\geq -\epsilon_1,~~\forall x\in \Cc,~y\in \Uc(x).
\end{align}
Let $(x^*,y^*)$ be a global solution of $\mathcal{BP}$ so that $f(x^*,y^*)=f^*$. Since $y^*\in\Sc(x^*)$, it follows that $p(x^*,y^*)=0$. By \eqref{eq:global calmness 1}, we have
\begin{align}\label{eq:global calmness 1 below}
    f(x^*,y^*)+\gamma^*p(x^*,y^*) \leq f(x,y)+\gamma^* p(x,y) +\epsilon_1,~~\forall x\in \Cc,~\forall y\in \Uc(x).
\end{align}
Inequality \eqref{eq:global calmness 1 below} along with the fact that the global solution of $\mathcal{BP}$ is feasible for $\mathcal{BP}_{\gamma p}$ prove that the global solution of $\mathcal{BP}$ achieves $\epsilon_1$-global-minimum for $\mathcal{BP}_{\gamma p}$.

Now for the converse part, we prove from $\mathcal{BP}_{\gamma p}$ to $\mathcal{BP}$. Since $(x_\gamma,y_\gamma)$ achieves $\epsilon_2$-global-minimum, it holds for any $x,y$ feasible for $\mathcal{BP}_{\gamma p}$ that
\begin{align}\label{eq:optimality0}
    f(x_\gamma,y_\gamma) + \gamma p(x_\gamma,y_\gamma) -\epsilon_1 \leq f(x,y) + \gamma p(x,y) -\epsilon_1+\epsilon_2.
\end{align}
In \eqref{eq:optimality0}, choosing $(x,y)=(x^*,y^*)$, which is a global solution of $\mathcal{BP}$, yields 
\begin{align*}
    f(x_\gamma,y_\gamma) + \gamma p(x_\gamma,y_\gamma) -\epsilon_1 &\leq f(x^*,y^*) -\epsilon_1+\epsilon_2\quad\text{since }p(x^*,y^*)=0 \nonumber\\
    &\leq  f(x_\gamma,y_\gamma) + \gamma^* p(x_\gamma,y_\gamma)+\epsilon_2\quad\text{by \eqref{eq:global calmness 1}.}
\end{align*}
Then we have
$$(\gamma-\gamma^*)p(x_\gamma,y_\gamma) \leq \epsilon_1+\epsilon_2\Rightarrow p(x_\gamma,y_\gamma) \leq (\epsilon_1+\epsilon_2)/(\gamma-\gamma^*).$$
Define $\epsilon_\gamma=p(x_\gamma,y_\gamma)$, then $\epsilon_\gamma \leq (\epsilon_1+\epsilon_2)/(\gamma-\gamma^*)$. By \eqref{eq:optimality0}, it holds for any feasible $(x,y)$ for $\mathcal{BP}_{\epsilon_\gamma}$that
\begin{align*}
    f(x_\gamma,y_\gamma) + \gamma p(x_\gamma,y_\gamma) \leq f(x,y) + \gamma p(x,y)+\epsilon_2\Rightarrow f(x_\gamma,y_\gamma) -f(x,y)\leq \gamma(p(x,y)-\epsilon_\gamma)+\epsilon_2 \leq \epsilon_2,
\end{align*}
where the last inequality follows from the feasibility of $(x,y)$.
This along with the fact that $(x_\gamma,y_\gamma)$ is feasible for problem $\mathcal{BP}_{\epsilon_\gamma}$prove that $(x_\gamma,y_\gamma)$ is achieves $\epsilon_2$-global-minimum of problem $\mathcal{BP}_{\epsilon_\gamma}$. \qed
\end{proof}

In Example \ref{exp:intro}, $\|\nabla_y g(x,y)\|^2$ is   a squared-distance-bound and the above theorem regarding global solutions holds. However, as illustrated in Example \ref{exp:intro}, a penalized problem with any $\gamma>0$ always admits a local solution that is meaningless to the original problem. In fact, the relationship between {\em local solutions} is more intricate than that between {\em global ones}. Nevertheless, we prove in the following theorem that under some verifiable conditions, the local solutions of $\mathcal{BP}_{\gamma p}$ are local solutions of $\mathcal{BP}_\epsilon$.
\begin{theorem}[Relation on local solutions]\label{the:BPgamr_epsilon_local_relax}
Assume the penalty function $p(x,\cdot)$ used in $\mathcal{BP}_{\gamma p}$ is continuous given any $x \in \Cc$ and $p(x,y)$ is $\rho$-squared-distance-bound function.
Given $\gamma>0$, let $(x_\gamma,y_\gamma)$ be a local solution of $\mathcal{BP}_{\gamma p}$ on $\Nc((x_\gamma,y_\gamma),r)$.
Assume $f(x_\gamma,\cdot)$ is $L$-Lipschitz-continuous on $\Nc(y_\gamma,r)$, 
and either one of the following holds
\begin{enumerate}[label=(\roman*)]
 \item There exists $\Bar{y}\in\Nc(y_\gamma,r)$ such that $\Bar{y}\in \Uc(x_\gamma)$ and $p(x_\gamma,\Bar{y})\leq \epsilon$ for some $\epsilon\geq 0$. Define $\Bar{\epsilon}_\gamma=\frac{L^2 \rho}{\gamma^2}\!+\!2\epsilon$.
    \item The set $\Uc(x_\gamma)$ is convex and the function $p(x_\gamma,\cdot)$ is convex. Define $\Bar{\epsilon}_\gamma=\frac{L^2 \rho}{\gamma^2}$.
\end{enumerate}
Then $(x_\gamma,y_\gamma)$ is a local solution of $\mathcal{BP}_{\epsilon_\gamma}$ with any  constant $\epsilon_\gamma$ satisfying $\epsilon_\gamma \leq \Bar{\epsilon}_\gamma$.
\end{theorem}
The proof of Theorem \ref{the:BPgamr_epsilon_local_relax} can be found in Appendix \ref{app.subsec.a2}. We provide a remark below.

\begin{remark}[Connecting with subsequent bilevel settings.]
In (i) of Theorem \ref{the:BPgamr_epsilon_local_relax}, we need an approximate global minimizer of $p(x_\gamma,\cdot)$; and, in (ii), we assume $p(x_\gamma,\cdot)$ is convex. Broadly speaking, these conditions essentially require $\min_{\Uc(x)} p(x,\cdot)$ to be globally solvable. Such a requirement is natural since finding a feasible point in $\Sc(x)$ is possible only if one can solve for $p(x,\cdot)=0$ on $\Uc(x)$. While they appear to be abstract, we will show how Conditions (i) and (ii) in Theorem \ref{the:BPgamr_epsilon_local_relax} can be verified in the subsequent sections.
Specifically, in Proposition \ref{pro:NP solution}, we verify Condition (i) by proving $p(x_\gamma,y_\gamma)=\mathcal{O}(1/\gamma^2)$ using the stationary condition of $(x_\gamma,y_\gamma)$ along with the so-called PL condition of $g(x,\cdot)$ \citep{karimi2016linear}; Proposition \ref{pro:CP solution} verifies Condition (i) with a similar idea extended to the constrained lower-level case; and, Proposition \ref{exact-penalty-nonsmooth} uses Condition (ii) directly for the non-smooth penalty function case.
\end{remark}

\section{Solving Bilevel Problems with Non-convex Lower-level Objectives}\label{sec:NP}
To develop bilevel algorithms with non-asymptotic convergence, in this section, we first consider $\mathcal{BP}$ with an {\em unconstrained} lower-level problem ($\Uc(x)\!=\!\reals^{d_y}$ in this case), given by

{\centering
\begin{tcolorbox}[width=0.8\textwidth]
\leqnomode
\vspace{-0.3cm}
\begin{align}
    \!\!\mathcal{UP}:\min_{x,y}f(x,y)\,\,\,\,\,\,~~{\rm s.t.}\,\,\,\,\,x \in \Cc,~y \in \arg\min_{y\in \reals^{d_y}}g(x,y)\nonumber
\end{align}
\vspace*{-0.4cm}
\reqnomode
\end{tcolorbox}}
\noindent where we assume $\Cc$ is a closed convex set and $f,g$ are continuously differentiable.

\subsection{Candidate penalty terms}
Following Section \ref{sec:generic}, to reformulate $\mathcal{UP}$, we first seek a squared-distance bound function $p(x,y)$ that satisfies Definition \ref{def:SDB}.
For a non-convex lower-level function $g(x,\cdot)$, an interesting property is the Polyak-Łojasiewicz (PL) inequality which is defined in the next assumption.

\begin{assumption}[Polyak-Lojasiewicz functions]\label{def:pl}
The lower-level function $g(x,\cdot)$ satisfies the $\frac{1}{\mu}$-PL inequality; that is, there exists $\mu>0$ such that given any $x\in\Cc$, it holds for any $y\in\reals^{d_y}$ that
\begin{equation}
    \|\nabla_y g(x,y)\|^2 \geq \frac{1}{\mu} (g(x,y)-v(x))~\text{ where $v(x)\coloneqq\min_{y\in\reals^{d_y}} g(x,y)$.}
\end{equation}
\end{assumption}

Taking policy optimization in reinforcement learning as an example, it has been proven in \cite[Lemma 8\&9]{mei2020softmax} that the non-convex discounted return objective satisfies the PL inequality under certain policy parameterization.
Moreover, recent studies have found that over-parameterized neural networks can lead to losses that satisfy the PL inequality \citep{liu2022loss}.

Under the PL inequality, we consider the following potential penalty functions:
\begin{center}
\begin{tcolorbox}[width=0.6\textwidth]
\vspace{-0.4cm}
\begin{subequations}\label{eq:penalty-all}
\begin{align}
\label{eq:g-v}
  &p(x,y)\!=\!g(x,y)-v(x) \\
\label{eq:gry g}
  &p(x,y)\!=\!\|\nabla_y g(x,y)\|^2.
\end{align}
\end{subequations}
\end{tcolorbox}
\end{center}
The next lemma shows that the above penalty functions are squared-distance bound functions.
\begin{lemma}[$\mu$-squared-distance-bound functions]\label{lem:pl penalty}
    Consider the following assumptions:
    \begin{enumerate}[label=(\roman*)]
         \item Suppose Assumption \ref{def:pl} holds, and there exists $L_g$ such that $\forall x\in\Cc$, $g(x,\cdot)$ is $L_g$-Lipschitz-smooth.
         \item Suppose Assumption \ref{def:pl} holds with {\color{black} PL constant} $\frac{1}{\sqrt{\mu}}$.
    \end{enumerate}
    Then \eqref{eq:g-v} and \eqref{eq:gry g} are $\mu$-squared-distance-bound functions respectively under (i) and (ii).
\end{lemma}
The proof of Lemma \ref{lem:pl penalty} can be found in Appendix \ref{app.subsec.b1}.

\subsection{Penalty reformulation and its optimality}
Given a squared-distance bound $p(x,y)$ and constants $\gamma\!>\!0$ and $\epsilon>0$, define the penalized problem and the $\epsilon$-approximate bilevel problem of $\mathcal{UP}$ respectively as

\begin{minipage}{.49\linewidth}
\begin{align}
    \mathcal{UP}_{\gamma p}:~&\min_{x,y}~ F_\gamma(x,y)\,\coloneqq f(x,y)+\gamma p(x,y)\nonumber\\
    &{\rm s.t.}~\,\,x \in \Cc.\nonumber
\end{align}
\end{minipage}
\begin{minipage}{.49\linewidth}
\begin{align}
    \mathcal{UP}_{\epsilon}:~&\min_{x,y} ~f(x,y)\nonumber\\
    &{\rm s.t.}~\,\,x \in \Cc,~p(x,y)\leq \epsilon.\nonumber
\end{align}
\end{minipage}\vspace{0.4cm}

% which is a special case of $\mathcal{BP}_{\gamma p}$ with $\Uc(x)=\reals^{d_y}$.
It remains to show that the solutions of the penalized reformulation $\mathcal{UP}_{\gamma p}$ are meaningful to the original bilevel problem $\mathcal{UP}$. Starting with the global solutions, we give the following proposition.
\begin{proposition}[Relation on global solutions]\label{pro:NP global solution}
Under Assumption \ref{asp:Lipschitz continuity}, 
let either of (a) and (b) hold:
\begin{enumerate}[label=(\alph*)]
    \item Condition (i) in Lemma \ref{lem:pl penalty} holds; and, choose $p(x,y)=g(x,y)-v(x)$; and, 
    \item Condition (ii) in Lemma \ref{lem:pl penalty} holds; and, choose $p(x,y)=\|\nabla_y g(x,y)\|^2$.
\end{enumerate}
Suppose $\gamma \geq L\sqrt{\mu \delta^{-1}}$ with some $\delta>0$. If $(x_\gamma,y_\gamma)$ be a global solution of the penalized problem $\mathcal{UP}_{\gamma p}$, then $(x_\gamma,y_\gamma)$ is a global solution of the original problem $\mathcal{UP}_{\epsilon_\gamma}$ with $\epsilon_\gamma\leq \delta$.
\end{proposition}
Proposition \ref{pro:NP global solution} follows directly from Theorem \ref{the:BP_gamr_epsilon} with $\epsilon_1=L\sqrt{\rho\delta}/2$, $\gamma\geq2\gamma^*=L\sqrt{\mu \delta^{-1}}$ and $\epsilon_2=0$. By Proposition \ref{pro:NP global solution}, the global solution of $\mathcal{UP}_{\gamma p}$ solves an approximate bilevel problem of $\mathcal{UP}$. However, since $\mathcal{UP}_{\gamma p}$ is generally non-convex, it is also important to consider the local solutions.
Following Theorem \ref{the:BPgamr_epsilon_local_relax}, the next proposition captures the relation on the local solutions.

\begin{proposition}[Relation on local solutions]\label{pro:NP solution}
Under Assumption \ref{asp:Lipschitz continuity}, 
let either of the following hold:
\begin{enumerate}[label=(\alph*)]
    \item Condition (i) in Lemma \ref{lem:pl penalty} holds; with some $\delta>0$, choose 
    \begin{equation*}
        p(x,y)=g(x,y)-v(x)\quad{\rm and}\quad\gamma \geq L\sqrt{3\mu\delta^{-1}}.
    \end{equation*}
    \item Condition (ii) in Lemma \ref{lem:pl penalty} holds; with some $\delta>0$, choose
    \begin{equation*}
    p(x,y)=\|\nabla_y g(x,y)\|^2 \quad{\rm and}\quad
    \gamma \geq \max\left\{L\sqrt{2\mu\delta^{-1}},L\sigma^{-1}\sqrt{\delta^{-1}}\right\}
    \end{equation*}
    %$p(x,y)=\|\nabla_y g(x,y)\|^2$ and $\gamma \geq \max\{L\sqrt{2\mu\delta^{-1}},L\sqrt{\delta^{-1}}/\sigma\}$, 
    where $\sigma>0$ is the lower-bound of the singular values of $\nabla_{yy}g(x,y)$  on $\{(x,y)\!\in\!\Cc\!\times\!\reals^{d_y}\!:y\notin \!\Sc(x)\}$. 
\end{enumerate}
    If $(x_\gamma,y_\gamma)$ is a local solution of $\mathcal{UP}_{\gamma p}$, then it is a local solution of $\mathcal{UP}_{\epsilon_\gamma}\!$ with some constant $\epsilon_\gamma \!\leq\! \delta$.
\end{proposition}
\begin{proof}
    We prove the proposition from the two conditions separately. 
 Since the lower-level unconstrained problems $\mathcal{UP}$ and $\mathcal{UP}_{\epsilon}$ are respectively special cases of the general bilevel problems $\mathcal{BP}$ and $\mathcal{BP}_{\epsilon}$, we leverage Theorem \ref{the:BPgamr_epsilon_local_relax} to prove the relation on local solutions. Specifically, we aim to prove Condition (i) in Theorem \ref{the:BPgamr_epsilon_local_relax} holds for the two cases in this proposition. For each case, we will use its respective stationary condition of $\mathcal{BP}_{\gamma p}$, along with some error bounds to prove Condition (i) in Theorem \ref{the:BPgamr_epsilon_local_relax} holds.  

\textbf{Proof of Case (a).}  Since $(x_\gamma,y_\gamma)$ is a local solution of $\mathcal{UP}_{\gamma p}$, it follows that $y_\gamma$ is a local solution of $\mathcal{UP}_{\gamma p}$ with $x=x_\gamma$. By the first-order stationary condition and Assumption \ref{asp:Lipschitz continuity}, it therefore holds that
$$\nabla_y f(x_\gamma,y_\gamma)+\gamma \nabla_y g(x_\gamma,y_\gamma)=0 \Rightarrow \|\nabla_y g(x_\gamma,y_\gamma)\| \leq L/\gamma.$$

Since $g(x_\gamma,\cdot)$ satisfies $1/\mu$-PL inequality from Condition (i) in Lemma \ref{lem:pl penalty}, it holds that 
\begin{align}
    \|\nabla_y g(x_\gamma,y_\gamma)\|^2 \geq \frac{1}{\mu}p(x_\gamma,y_\gamma)=\frac{1}{\mu}(g(x_\gamma,y_\gamma)-v(x_\gamma)).\nonumber
\end{align}
The above two inequalities imply $p(x_\gamma,y_\gamma) \leq \frac{L^2 \mu}{\gamma^2}$. Further notice that $\mathcal{UP}$ and $\mathcal{UP}_{\gamma p}$ are respectively the special cases of $\mathcal{BP}$ and $\mathcal{BP}_{\gamma p}$ with $\Uc(x)=\reals^{d_y}$; and $p(x,y)$ is a squared-distance-bound function by Lemma \ref{lem:pl penalty}, then the result directly follows from Theorem \ref{the:BPgamr_epsilon_local_relax} where Condition $(i)$ is met with $\Bar{y}=y_\gamma$, $\rho=\mu$ and $\epsilon=\frac{L^2 \mu}{\gamma^2}$ with $\gamma \geq L\sqrt{3\mu\delta^{-1}}$.

\textbf{Proof of Case (b).} Suppose $y_\gamma \notin \Sc(x_\gamma)$. Since $(x_\gamma,y_\gamma)$ is a local solution of $\mathcal{UP}_{\gamma p}$, $y_\gamma$ is a local solution of $\mathcal{UP}_{\gamma p}$ with $x=x_\gamma$. By the first-order stationary condition and Assumption \ref{asp:Lipschitz continuity}, it holds that
$$\nabla_y f(x_\gamma,y_\gamma)+2\gamma \nabla_{yy} g(x_\gamma,y_\gamma)\nabla_y g(x_\gamma,y_\gamma)=0 \Rightarrow \|\nabla_{yy} g(x_\gamma,y_\gamma)\nabla_y g(x_\gamma,y_\gamma)\| \leq L/2\gamma$$
which along with the assumption that the singular values of $\nabla_{yy} g(x,y)$ on $\{x\in\Cc,y\in\Uc(x):y\notin \Sc(x)\}$ are lower bounded by $\sigma>0$ gives
\begin{align}\label{eq:idk72}
    p(x_\gamma,y_\gamma)=\|\nabla_y g(x_\gamma,y_\gamma)\|^2 \leq \frac{\|\nabla_{yy} g(x_\gamma,y_\gamma)\nabla_y g(x_\gamma,y_\gamma)\|^2}{\sigma^2} \leq L^2/(4\gamma^2\sigma^2).
\end{align}
When $y_\gamma \in\Sc(x_\gamma)$, we know $p(x_\gamma,y_\gamma)=0$ and thus \eqref{eq:idk72} still holds.
Further notice that $\mathcal{UP}$ and $\mathcal{UP}_{\gamma p}$ are special cases of $\mathcal{BP}$ and $\mathcal{BP}_{\gamma p}$ with $\Uc(x)=\reals^{d_y}$; and $p(x,y)$ is a squared-distance-bound function by Lemma \ref{lem:pl penalty}, then the result directly follows from Theorem \ref{the:BPgamr_epsilon_local_relax} where Condition $(i)$ is met with $\Bar{y}=y_\gamma$, $\rho=\mu$ and $\epsilon=L^2/(4\gamma^2\sigma^2)$ with $\gamma \geq \max\{L\sqrt{2\mu\delta^{-1}},L\sqrt{\delta^{-1}}/\sigma\}$. \qed
\end{proof}
Proposition \ref{pro:NP solution} explains the observations in Figure \ref{fig:intro toy} and Example \ref{exp:intro}. When using the penalty function $p(x,y)\!=\!\|\nabla_y g(x,y)\|^2$, the convergent suboptimal local solution $y=\frac{2\pi}{3}$ mentioned in Example \ref{exp:intro} yields $\nabla_{yy} g(x,y)=0$ which violates the condition on the singular values of $\nabla_{yy} g(x,y)$ in (b) of Proposition \ref{pro:NP solution}. On the other hand, it can be checked that Condition (a) of Proposition \ref{pro:NP solution} holds in Example \ref{exp:intro}.

Propositions \ref{pro:NP global solution} and \ref{pro:NP solution} suggest $\gamma=\Omega(\delta^{-0.5})$ to achieve $\epsilon_\gamma \leq \delta$. Next we show this bound is also tight.

\begin{corollary}[Lower bound on the penalty constant]\label{cor:gamma tight}
In Proposition \ref{pro:NP global solution} or \ref{pro:NP solution}, to guarantee $\epsilon_\gamma =\mathcal{O}(\delta)$,
the lower-bound on the penalty constant $\gamma=\Omega(\delta^{-0.5})$ is tight.
\end{corollary}
\begin{proof}
Consider the following special case of $\mathcal{UP}$:
\begin{equation}
    \min_{y\in\reals} f(x,y)=y,\quad {\rm s.t.}\quad y\in\arg\min_{y\in\reals}g(x,y)=y^2.
\end{equation}
In the example, the two penalty terms $\frac{1}{4}\|\nabla_y g(x,y)\|^2$ and $g(x,y)-v(x)$ coincide to be $p(x,y)=y^2$.
In this case, solutions of $\mathcal{UP}$ and $\mathcal{UP}_{\gamma p}$ are respectively $0$ and $-\frac{1}{2\gamma}$. Thus $-\frac{1}{2\gamma}$ is a solution of  $\mathcal{CP}_{\epsilon_\gamma}$ with $\epsilon_\gamma=1/(4\gamma^2)$. To ensure $\epsilon_\gamma =\mathcal{O}(\delta)$, $\gamma=\Omega(\delta^{-0.5})$ is required in this example. Then the poof is complete by the fact that the assumptions in Proposition \ref{pro:NP global solution} and \ref{pro:NP solution} hold in this example. \qed
\end{proof}

Note that in addition to the relation between local/global solutions, we can further establish the stationary relations of $\mathcal{UP}$ and $\mathcal{UP}_{\gamma p}$ with $p(x,y)$ in \eqref{eq:g-v} and \eqref{eq:gry g}. Following \citep[Theorem 1]{xiao2023}, the $\epsilon$-stationary point of $\mathcal{UP}$ is defined as the $\epsilon$-KKT point of the gradient-based constrained reformulated problem 
$\min_{x,y} f(x,y), \text{ s.t. }\nabla_y g(x,y)=0$.  Specifically, the $\epsilon$-stationary point of $\mathcal{UP}$ is a pair $(x,y)$ such that there exists $w\in\mathbb{R}^{d_y}$, together with $(x,y)$, satisfying the following conditions
\begin{subequations}
\begin{align}
    \text{Stationarity: }&\quad \|\nabla_x f(x,y)+\nabla_{xy}g(x,y)w\|\leq \epsilon\label{kkt1-x-m}\\
    &\quad\|\nabla_y f(x,y)+\nabla_{yy}g(x,y)w\|\leq \epsilon\label{kkt1-y-m}\\
    \text{Feasibility: }&\quad\|\nabla_y g(x,y)\|\leq \epsilon. \label{kkt1-low-m} 
\end{align}
\end{subequations}

Building upon the above definition of $\epsilon$-stationary point of $\mathcal{UP}$, we will establish the relation on the stationary points of $\mathcal{UP}$ and those of $\mathcal{UP}_{\gamma p}$ in the next proposition. 
\begin{proposition}[Relation on stationary points]\label{pro:SP}
Suppose Assumption \ref{asp:Lipschitz continuity}, Condition (i) in Lemma \ref{lem:pl penalty}, 
and either of the following holds: 
\begin{enumerate}[label=(\alph*)]
    \item Suppose the gradient $\nabla_y g(x,y)$ is Lipschitz smooth with $L_{g,2}$, and choose 
    \begin{equation*}
        p(x,y)=g(x,y)-v(x)\quad{\rm and}\quad\gamma = \Omega(\delta^{-0.5}).
    \end{equation*}
    % $p(x,y)=g(x,y)-v(x)$ and $\gamma \geq L\sqrt{3\mu\delta^{-1}}$ with some $\delta>0$.
    \item Choose the candidate penalty term 
    \begin{equation*}
    p(x,y)=\|\nabla_y g(x,y)\|^2 \quad{\rm and}\quad
    \gamma = \Omega(\delta^{-0.5}).
    \end{equation*}
    %$p(x,y)=\|\nabla_y g(x,y)\|^2$ and $\gamma \geq \max\{L\sqrt{2\mu\delta^{-1}},L\sqrt{\delta^{-1}}/\sigma\}$, 
    
    % Choose $\gamma \geq \max\{L\sqrt{2\mu\delta^{-1}},L\sqrt{\delta^{-1}}/\sigma\}$ with some $\delta>0$.
\end{enumerate}
    If $(x_\gamma,y_\gamma)$ is an $\delta$-stationary point of $\mathcal{UP}_{\gamma p}$, then it is an $\epsilon_\gamma$-stationary point of $\mathcal{UP}$ with some $\epsilon_\gamma ={\cal O}(\delta)$. 
\end{proposition}
The proof of Proposition \ref{pro:SP} is deferred in Appendix \ref{app.subsec.b3}. The smoothness condition of $\nabla_y g(x,y)$ is widely assumed in bilevel literature \citep{chen2021tighter,ji2021bilevel,hong2020ac,Li2022fully,chen2022stable,dagreou2022framework}. In particular, the stationary relation in Proposition \ref{pro:SP} does not depend on any assumed constraint qualification (CQ) conditions made in the bilevel optimization literature \citep{gong2021automatic, ye2022bome, dempe2012bilevel}.

As a summary, Propositions \ref{pro:NP global solution} and \ref{pro:NP solution} imply that $\mathcal{UP}$ and $\mathcal{UP}_{\gamma p}$ are related in the sense that one can globally/locally solve an approximate bilevel problem of $\mathcal{UP}$ by globally/locally solving the penalized problem $\mathcal{UP}_{\gamma p}$ instead. 
 Furthermore, Proposition \ref{pro:SP} shows one can recover a stationary point of $\mathcal{UP}$ by finding a stationary point of $\mathcal{UP}_{\gamma p}$.
A natural approach to solving the penalized problem $\mathcal{UP}_{\gamma p}$ is the projected gradient descent method. At each iteration $k$, we assume access to $h_k$ which is either $\nabla p(x_k,y_k)$ or its estimate if $\nabla p$ cannot be exactly evaluated. We then update $(x_k,y_k)$ with $\nabla F_\gamma(x_k,y_k)$ evaluated using $h_k$. The process is summarized in Algorithm \ref{alg:PBGD}.

\begin{algorithm}[t]
% \setstretch{1.2}
\caption{PBGD: Penalized bilevel gradient descent - Meta version}
% \vspace{0.2cm}
\begin{algorithmic}[1]
\STATE Select $(x_1,y_1) \in \Zc\coloneqq\Cc \times\Uc(x)$.
Select step size $\alpha$, penalty constant $\gamma$ and iteration number $K$.

\FOR{$k=1$ {\bfseries to} $K$}
\STATE Compute $h_k=\nabla p(x_k,y_k)$ or its estimate.
\STATE $(x_{k+1},y_{k+1})=\Proj_{\Zc}\Big((x_k,y_k)-\alpha \big(\nabla f(x_k,y_k)+\gamma h_k\big)\Big)$.
\ENDFOR
\end{algorithmic}
\label{alg:PBGD}
\end{algorithm}
When $p(x,y)=\|\nabla_y g(x,y)\|^2$, $\nabla p(x,y)$ can be exactly evaluated. In this case, Algorithm \ref{alg:PBGD} is a standard projected gradient method and its convergence property directly follows from the existing literature under some Lipchitz condition on $\nabla p(x,y)$. One caveat is that $\nabla p(x,y)$ involves the second-order information of $g(x,y)$, which may be costly.
In the next subsection, we focus on the penalty function $p(x,y)=g(x,y)-v(x)$ and discuss how $\mathcal{UP}_{\gamma p}$ can be solved via only first-order information.

\subsection{Fully first-order PBGD with function value gap}
We consider solving $\mathcal{UP}_{\gamma p}$ with $p(x,y)$ chosen as the function value gap \eqref{eq:g-v}.
To solve $\mathcal{UP}_{\gamma p}$ with the gradient-based method, the obstacle is that $\nabla p(x,y)$ requires $\nabla v(x)$. On one hand, $v(x)$ is not necessarily smooth. Even if $v(x)$ is differentiable,
$\nabla v(x) \neq \nabla_x g(x,y^*)~\text{in general, where }y^*\in \Sc(x)$.
However, it is possible to compute $\nabla v(x)$ efficiently under some relatively mild conditions.

\begin{lemma}[{\cite[Lemma A.5]{nouiehed2019solving}}]\label{lem:grad vx}
Assume Assumption \ref{def:pl} holds, and $g$ is $L_g$-Lipschitz-smooth. Then $\nabla v(x)=\nabla_x g(x,y^*)$ for any $y^*\in \Sc(x)$, and $v(x)$ is $(L_g \!+\!L_g^2 \mu)$-Lipschitz-smooth.
\end{lemma}

Under the conditions in Lemma \ref{lem:grad vx}, $\nabla v(x)$ can be evaluated directly at any optimal solution of the lower-level problem. This suggests one find a lower-level optimal solution $y^*\in\Sc(x)$, and evaluate the penalized gradient $\nabla F_\gamma (x,y)$ with $\nabla v(x) = \nabla_x g(x,y^*)$. Following this idea, given outer iteration $k$ and $x_k$, we run $T_k$ steps of inner GD update to solve the lower-level problem: 
\begin{subequations}
\begin{align}\label{eq:omega noconvex update}
    \omega_{t+1}^{(k)}&=\omega_t^{(k)} - \beta \nabla_y g(x_k,\omega_t^{(k)}),~t=1,\dots,T_k
\end{align}
where $\omega_1^{(k)}=y_k$.
Update \eqref{eq:omega noconvex update} yields an approximate lower-level solution $\hat{y}_k=\omega_{T_k+1}^{(t)}$. Then we can approximate $\nabla F_\gamma(x_k,y_k)$ with $\hat{y}_k$ and update $(x_k,y_k)$ via:
\label{eq:nonconvex update}
\begin{align}\label{eq:xy nonconvex update}
    (x_{k+1},y_{k+1})&=\Proj_{\Zc} \Big((x_k,y_k)- \alpha \big(\nabla f(x_k,y_k) + \gamma(\nabla g(x_k,y_k)-\overline{\nabla}_x g(x_k,\hat{y}_k)\big)\Big)
\end{align}
\end{subequations}
where $\Zc= \Cc \times \reals^{d_y}$ and $\overline{\nabla}_x g(x,y)\coloneqq (\nabla_x g(x,y),\mathbf{0})$ with $\mathbf{0}\in\reals^{d_y}$.
The update is summarized in Algorithm \ref{alg:V-PBGD}, which is a function value gap-based special case of the generic penalty-based bilevel PBGD method (Algorithm \ref{alg:PBGD}) with $h_k = \nabla g(x_k,y_k)-\overline{\nabla}_x g(x_k,\hat{y}_k)$. 
% As will be shown later, $h_k$ is an efficient estimator of $\nabla p(x_k,y_k)$.

Notice that only first-order information is required in update \eqref{eq:nonconvex update}, which is in contrast to the implicit gradient methods or some iterative differentiation methods where higher-order derivatives are required; see, e.g., \citep{ghadimi2018approximation,franceschi2017forward,liu2021towards}. In modern machine learning applications, this could substantially save computational cost since the dimension of the parameter is often large, making higher-order derivatives particularly costly. 

\begin{algorithm}[t]
% \setstretch{1.2}
\caption{V-PBGD: Function value gap-based fully first-order PBGD}
% \vspace{0.2cm}
\begin{algorithmic}[1]
\STATE Select $(x_1,y_1)\in \Zc=\Cc \times \reals^{d_y}$. Select step sizes $\alpha,\beta$, constant $\gamma$, iteration numbers $T_k$ and $K$.

\FOR{$k=1$ {\bfseries to} $K$}
\STATE Obtain the auxiliary variable $\hat{y}_k=\omega_{T_k+1}^{(k)}$ by running $T_k$ steps of inner GD update \eqref{eq:omega noconvex update}.
\STATE Use $\hat{y}_k$ to approximate $\nabla v(x_k)$ via $\nabla_x g(x_k,\hat{y}_k)$ and update $(x_k,y_k)$ following \eqref{eq:xy nonconvex update}.
% \FOR{$t=1$ {\bfseries to} $T_k$}
% \SATE Update $\omega_{t}^{(k)}$ following \eqref{eq:omega noconvex update}.
% \ENDFOR
\ENDFOR

% \STATE Output $\{x_m,y_m\}$ with $m \in \arg\min_{k\in\{1,...,K\}}\|(x_{k+1},y_{k+1})-(x_k,y_k)\|$.
\end{algorithmic}
\label{alg:V-PBGD}
\end{algorithm}

\subsection{Analysis of PBGD with function value gap}
We first introduce the following regularity assumption commonly made in the convergence analysis of the gradient-based bilevel optimization methods \citep{chen2021tighter,grazzi2020iteration}.
\begin{assumption}[smoothness]\label{asp:in theorem}
There exist   constants $L_f$ and $L_g$ such that $f(x,y)$ and  $g(x,y)$ are respectively $L_f$-Lipschitz-smooth and $L_g$-Lipschitz-smooth in $(x,y)$.
\end{assumption}

Define the projected gradient of $\mathcal{UP}_{\gamma p}$ at $(x_k,y_k)\in\Zc$ as
\begin{align}\label{eq:nonconvex proximal grad}
    G_\gamma(x_k,y_k) \coloneqq \frac{1}{\alpha}\big((x_k,y_k)-(\Bar{x}_{k+1},\Bar{y}_{k+1})\big),
\end{align}
where $(\Bar{x}_{k+1},\Bar{y}_{k+1})\coloneqq \Proj_{\Zc}((x_k,y_k)-\alpha \nabla F_\gamma (x_k,y_k))$.
This definition \eqref{eq:nonconvex proximal grad} is commonly used as the convergence metric for the projected gradient methods. 
It is known that given a convex $\Zc$, $G_\gamma(x,y)\!=\!0$ if and only if $(x,y)$ is a stationary point of $\mathcal{UP}_{\gamma p}$ \citep{ghadimi2016mini}.
We provide the following theorem on the convergence of V-PBGD.
\begin{theorem}\label{the:nonconvex xy convergence}
Consider V-PBGD (Algorithm \ref{alg:V-PBGD}). 
Suppose Assumptions \ref{asp:Lipschitz continuity},\ref{def:pl} and \ref{asp:in theorem} hold.
Select $\omega_1^{(k)}=y_k$ and choose the constants as
\begin{align}
&\alpha\in (0, (L_f + \gamma (2L_g + L_g^2 \mu))^{-1}],~\beta\in (0,L_g^{-1}],~\gamma \geq L\sqrt{3\mu\delta^{-1}},~T_k = \Omega(\log(\alpha k)).\nonumber
\end{align}
% $$, $\beta\leq 1/L_g$. Given $c_\beta=1-\beta/2\mu$, let $T_k \geq \max\{-\log_{c_\beta}(16L_g^2),-2\log_{c_\beta}(2\alpha k)\}$. 
  i) With $C_f\!=\!\inf_{(x,y)\in\Zc} f(x,y)$, it holds that
\begin{equation}
    \frac{1}{K}\sum_{k=1}^K \|G_\gamma(x_k,y_k)\|^2 \leq \frac{18\big(F_\gamma(x_1,y_1)-C_f\big)}{\alpha K}+\frac{10 L^2 L_g^2}{K}.\nonumber
\end{equation}
ii) Suppose $\lim_{k\rightarrow\infty}(x_k,y_k)=(x^*,y^*)$, then $(x^*,y^*)$ is a stationary point of $\mathcal{UP}_{\gamma p}$. If $(x^*,y^*)$ is a local/global solution of $\mathcal{UP}_{\gamma p}$, it is a local/global solution of  $\mathcal{UP}_{\epsilon_\gamma}$ with some $\epsilon_\gamma\leq\delta$.
\end{theorem}
\begin{proof}
{\color{black}
We will first show the linear convergence of the auxiliary variable $\omega_t^{(k)}$, and then show the convergence of the main variables $(x_k,y_k)$ under the fast convergent $\omega_t^{(k)}$. The intuition is that $\omega_t^{(k)}$ converges fast under the smooth PL condition, which decreases the gradient estimation error tailored to the $\nabla g(x_k,y_k)-\bar{\nabla}_x g(x_k,\hat{y}_k)$ in the outer loop. With small enough error, the V-PBGD algorithm then shows similar convergence behavior as the projected GD algorithm.}

\textbf{Convergence of $\omega$.} We first provide the convergence of the sequence $\{\omega_t^{(k)}\}$ at some outer iteration $k$.
We omit index $k$ since this proof holds given any $k$. By Lipschitz smoothness of $g(x,\cdot)$, it holds that
\begin{align}
    g(x,\omega_{t+1})  &\leq g(x,\omega_t) -\beta\|\gr_y g(x,\omega_t)\|^2+\frac{L_g \beta^2}{2}\|\gr_y g(x,\omega_t)\|^2 \nonumber\\
    &\leq g(x,\omega_t)-\frac{\beta}{2}\|\gr_y g(x,\omega_t)\|^2\quad\text{since}~L_g \beta \leq 1.\nonumber
\end{align}
By $\frac{1}{\mu}$-PL condition of $g(x,\cdot)$, we further have
\begin{align}
    g(x,\omega_{t+1})-v(x) &\leq g(x,\omega_t)-v(x)-\frac{\beta}{2 \mu}\big(g(x,\omega_t)-v(x)\big)\nonumber\\
 &\leq \big(1-\frac{\beta}{2 \mu}\big)\big(g(x,\omega_t)-v(x)\big).\nonumber
\end{align}
Iteratively applying the above inequality for $t=1,\dots,T$ yields
\begin{align}\label{eq:idkgx}
    g(x,\omega_{T+1})-v(x) \leq \big(1-\frac{\beta}{2 \mu}\big)^T \big(g(x,\omega_1)-v(x)\big).
\end{align}

By \citep[Theorem 2]{karimi2016linear}, the $\frac{1}{\mu}$-PL condition of $g(x,\cdot)$ also implies the error bound of $g(x,\cdot)$, which leads to   
$$g(x,\omega_{T+1})-v(x) \geq \frac{1}{\mu} d_{S(x)}^2(\omega_{T+1}),~x\in \Cc$$
This error bound along with \eqref{eq:idkgx} yield
\begin{align}\label{eq:nonconvex omega convergence}
    d_{S(x_k)}^2(\hat{y}_k) \leq \mu\big(1-\frac{\beta}{2 \mu}\big)^{T_k} \big(g(x_k,\omega_1^{(k)})-v(x_k)\big).
\end{align}

Notice the term $g(x_k,\omega_1^{(k)})-v(x_k)$ in \eqref{eq:nonconvex omega convergence} depends on the drifting variable $x_k$. If $\omega_1^{(k)}$ is not carefully chosen, $g(x_k,\omega_1^{(k)})-v(x_k)$ can grow unbounded with $k$ and hence hinder the convergence. To prevent this, we choose $\omega_1^{(k)}=y_k$ in the analysis. Since $g(x,\cdot)$ is $\frac{1}{\mu}$-PL for any $x\in \Cc$, it holds that
\begin{align}\label{eq:omega1 upper-bound}
    g(x_k,\omega_1^{(k)})-v(x_k) \leq \frac{1}{\mu}\|\gr_y g(x_k,\omega_1^{(k)})\|^2 
    &= \frac{1}{\mu}\lnorm{\frac{y_k-y_{k+1}-\alpha \nabla_y f(x_k,y_k)}{\alpha\gamma}}^2 \nonumber\\ 
    &\leq \frac{2}{\mu \gamma^2 \alpha^2} \|y_{k+1}-y_k\|^2+\frac{2 L^2}{\mu\gamma^2}
\end{align}
where we have used Young's inequality and the condition that $f(x_k,\cdot)$ is $L$-Lipschitz-continuous. Later we will show that the inexact gradient descent update \eqref{eq:xy nonconvex update} decreases $\|(x_{k+1},y_{k+1})-(x_k,y_k)\|$ and therefore upper-bounds $g(x_k,\omega_1^{(k)})-v(x_k)$. 

\textbf{Convergence of $(x,y)$.} Next we give the convergence proof of the main sequence $\{(x_k,y_k)\}$.
In this proof, we write $z=(x,y)$. Update \eqref{eq:xy nonconvex update} can be written as
$$z_{k+1}=\Proj_{\Zc}\big(z_k-\alpha \hat{\gr}F_\gamma (z_k;\hat{y}_k)\big)$$
where $\hat{\gr}F_\gamma (z_k;\hat{y}_k\big)\coloneqq \nabla f(z_k) + \gamma(\nabla g(z_k)-\overline{\nabla}_x g(x_k,\hat{y}_k)\big)$.

By the assumptions made in this theorem and Lemma \ref{lem:grad vx}, $F_\gamma$ is $L_\gamma$-Lipschitz-smooth with $L_\gamma = L_f + \gamma (2L_g + L_g^2 \mu)$.
Then by Lipschitz-smoothness of $F_\gamma$, it holds that
\begin{align}\label{eq:00}
    F_\gamma(z_{k+1}) 
    &\leq F_\gamma(z_k) + \ip{\gr F_\gamma(z_k)}{z_{k+1}-z_k} + \frac{L_\gamma}{2}\|z_{k+1}-z_k\|^2 \nonumber\\
    &\stackrel{\alpha\leq \frac{1}{L_\gamma}}{\leq} F_\gamma(z_k) + \ip{\hat{\gr}F_\gamma (z_k;\hat{y}_k)}{z_{k+1}-z_k} + \frac{1}{2\alpha}\|z_{k+1}-z_k\|^2 +\ip{\gr F_\gamma(z_k)-\hat{\gr}F_\gamma (z_k;\hat{y}_k)}{z_{k+1}-z_k}.
\end{align}
Consider the second term in the RHS of \eqref{eq:00}.
By Lemma \ref{lem:support 1}, $z_{k+1}$ can be written as
$$z_{k+1}=\arg\min_{z\in\Zc}\ip{\hat{\gr}F_\gamma (z_k;\hat{y}_k)}{z}+\frac{1}{2\alpha}\|z-z_k\|^2.$$
By the first-order optimality condition of the above problem, it holds that
\begin{align}
    \ip{\hat{\gr}F_\gamma (z_k;\hat{y}_k)+\frac{1}{\alpha}(z_{k+1}-z_k)}{z_{k+1}-z} \leq 0,~\forall z \in \Zc.\nonumber
\end{align}
Since $z_k\in\Zc$, we can choose $z=z_k$ in the above inequality and obtain
\begin{align}\label{eq:bb00}
    \ip{\hat{\gr}F_\gamma (z_k;\hat{y}_k)}{z_{k+1}-z_k} \leq -\frac{1}{\alpha}\|z_{k+1}-z_k\|^2.
\end{align}
Consider the last term in the RHS of \eqref{eq:00}. By Young's inequality, we first have
\begin{align}\label{eqn.app.proof.b8}
    \ip{\gr F_\gamma(z_k)-\hat{\gr}F_\gamma (z_k;\hat{y}_k)}{z_{k+1}-z_k} \leq \alpha \norm{\gr F_\gamma(z_k)-\hat{\gr}F_\gamma (z_k;\hat{y}_k)}^2 + \frac{1}{4\alpha}\norm{z_{k+1}-z_k}^2
\end{align}
where the first term in the above inequality can be bounded as 
\begin{align}\label{eq:error nonconvex}
    \norm{\gr F_\gamma(z_k)-\hat{\gr}F_\gamma (z_k;\hat{y}_k)}^2
    &= \gamma^2 \norm{\nabla v(x_k) - \overline{\nabla}_x g(x_k,\hat{y}_k)}^2 \nonumber\\
    &= \gamma^2 \norm{\nabla g(x_k,y^*)|_{y^*\in\Sc(x_k)} - \overline{\nabla}_x g(x_k,\hat{y}_k)}^2\quad\text{by Lemma \ref{lem:grad vx}},\nonumber\\
    &\leq \gamma^2 L_g^2 d^2_{\Sc(x_k)}(\hat{y}_k)\quad\text{by choosing $y^*\in\arg\min_{y'\in\Sc(x_k)}\norm{y'-\hat{y}_k}$}, \nonumber\\
    &\leq \gamma^2 L_g^2 \mu\big(1-\frac{\beta}{2 \mu}\big)^{T_k} \big(g(x_k,\omega_1^{(k)})-v(x_k)\big) \quad\text{by \eqref{eq:nonconvex omega convergence}},\nonumber\\
    &\leq \big(1-\frac{\beta}{2 \mu}\big)^{T_k} \big(\frac{2 L_g^2}{\alpha^2}\norm{z_{k+1}-z_k}^2+2L^2 L_g^2\big) \quad\text{by \eqref{eq:omega1 upper-bound}},\nonumber\\
    &\leq \frac{1}{8\alpha^2}\norm{z_{k+1}-z_k}^2+\frac{ L^2 L_g^2}{2\alpha^2 k^2} 
\end{align}
where the last inequality requires $T_k \geq \max\{-\log_{c_\beta}(16L_g^2),-2\log_{c_\beta}(2\alpha k)\}~\text{with}~c_\beta=1-\frac{\beta}{2\mu}$.

Plugging the inequality \eqref{eq:error nonconvex} into \eqref{eqn.app.proof.b8} yields
\begin{align}\label{eq:bb01}
    \ip{\gr F_\gamma(z_k)-\hat{\gr}F_\gamma (z_k;\hat{y}_k)}{z_{k+1}-z_k} \leq \frac{3}{8\alpha}\norm{z_{k+1}-z_k}^2+\frac{ L^2 L_g^2 }{2\alpha k^2}.
\end{align}
Substituting \eqref{eq:bb01} and \eqref{eq:bb00} into \eqref{eq:00} and rearranging the resulting inequality yield
\begin{align}\label{eq:trash00}
\frac{1}{8\alpha}\|z_{k+1}-z_k\|^2 \leq F_\gamma(z_k)-F_\gamma(z_{k+1}) + \frac{ L^2 L_g^2 }{2\alpha k^2}.
\end{align}
With $\Bar{z}_{k+1}$ defined in \eqref{eq:nonconvex proximal grad}, we have
\begin{align}\label{eqn.app-proof.b12}
    \norm{\Bar{z}_{k+1}-z_k}^2 &\leq 2\norm{\Bar{z}_{k+1}-z_{k+1}}^2+2\|z_{k+1}-z_k\|^2 \nonumber\\
    &\leq 2\alpha^2 \norm{\gr F_\gamma(z_k)-\hat{\gr}F_\gamma (z_k;\hat{y}_k)}^2\!+\!2\|z_{k+1}\!-\!z_k\|^2 \nonumber\\
    &\leq \frac{9}{4}\|z_{k+1}-z_k\|^2 + \frac{ L^2 L_g^2 }{k^2}
\end{align}
where the second inequality uses non-expansiveness of $\Proj_{\Zc}$ and the last one follows from \eqref{eq:error nonconvex}.

Together \eqref{eq:trash00} and \eqref{eqn.app-proof.b12} imply
\begin{align}
    \norm{\Bar{z}_{k+1}-z_k}^2 \leq 18\alpha\big(F_\gamma(z_k)-F_\gamma(z_{k+1})\big)+\frac{10 L^2 L_g^2 }{k^2}.\nonumber
\end{align}
Since $f(z) \geq C_f$ for any $z\in\Zc$ and $g(x,y)-v(x) \geq 0$, $F_\gamma(z) \geq C_f$ for any $z\in\Zc$.
Taking a telescope sum of the above inequality and using $G_\gamma(z_k)=\frac{1}{\alpha}(z_k-\Bar{z}_{k+1})$ yield
\begin{align}
    \sum_{k=1}^K \|G_\gamma(z_k)\|^2 \leq \frac{18\big(F_\gamma(z_1)-C_f\big)}{\alpha}+\sum_{k=1}^K\frac{10 L^2 L_g^2 }{k^2}\nonumber
\end{align}
which along with the fact $\sum_{k=1}^K \frac{1}{k^2}\leq \int_{1}^K \frac{1}{x^2} dx=1-\frac{1}{K}$ implies
\begin{align}\label{eq:result nonconvex}
    \sum_{k=1}^K \|G_\gamma(z_k)\|^2 \leq \frac{18\big(F_\gamma(z_1)-C_f\big)}{\alpha}+10 L^2 L_g^2 .
\end{align}
This proves (i) in Theorem \ref{the:nonconvex xy convergence}. 

Suppose $\lim_{k \rightarrow \infty}z_k=z^*$. Since $\nabla F_\gamma(z)$ is continuous, $G_\gamma(z)$ is continuous and thus $\lim_{k \rightarrow \infty}G_\gamma(z_k)=G_\gamma(z^*)$. By \eqref{eq:result nonconvex}, $G_\gamma(z^*)=0$, that is $z^* = \Proj_{\Zc}\big(z^*-\alpha\nabla F_\gamma(z^*)\big)$. This further implies
$$\ip{\gr F_\gamma (z^*)}{z^*-z} \leq 0,~\forall z \in \Zc$$
which indicates $z^*$ is a stationary point of $\mathcal{UP}_{\gamma p}$. If $z^*$ is a local/global solution, it follows from Proposition \ref{pro:NP global solution} and \ref{pro:NP solution} that the rest of the result holds. \qed
\end{proof}

Theorem \ref{the:nonconvex xy convergence} implies an iteration complexity of $\widetilde{\mathcal{O}}(\gamma\epsilon^{-1})$ to find an $\epsilon$-stationary-point of $\mathcal{UP}_{\gamma p}$. This recovers the iteration complexity of the projected GD method \citep{nesterov2013gradient} with a smoothness constant of $\Theta(\gamma)$. If we choose $\delta=\epsilon$ and thus $\gamma={\cal O}(\epsilon^{-0.5})$, then we get an iteration complexity of $\widetilde{\mathcal{O}}(\epsilon^{-1.5})$. 
In (ii) of Theorem \ref{the:nonconvex xy convergence}, under no stronger conditions needed for the projected GD method to yield meaningful solutions, the V-PBGD algorithm finds a local/global solution of the approximate $\mathcal{UP}$.

\begin{remark}[Convergence to the local minima.]
Note that the non-asymptotic guarantee in Theorem \ref{the:nonconvex xy convergence} is established for the stationary-point convergence of $\mathcal{UP}_{\gamma p}$. Then together with the relation on stationary points in Proposition \ref{pro:SP}, one can gauge the $\epsilon$-stationary-point convergence to the original bilevel problem $\mathcal{UP}$. To ensure convergence to the local or even global minima of $\mathcal{UP}$, additional assumptions are needed beyond Theorem \ref{the:nonconvex xy convergence}. The key ingredients towards the convergence to the local minima of $\mathcal{UP}$ require the additional argument on escaping saddle points of the nonconvex loss landscape of $\mathcal{UP}_{\gamma p}$, which itself is an active research area in the recent years. Fortunately, recent advances have shown that random initialization or perturbation during the optimization trajectory of (projected) GD can almost surely ensure convergence to the local minima of $\mathcal{UP}_{\gamma p}$ and thus those of $\mathcal{UP}$; see e.g., \citep{lee2016gradient,jin2017escape,davis2022proximal}.
\end{remark}

\subsection{Extension to the stochastic version of PBGD}\label{app.subsec.b4}
In addition to the deterministic PBGD algorithm, a stochastic version of the V-PBGD algorithm and its convergence analysis are provided in this subsection.

With random variables $\xi$ and $\psi$, we assume access to $\nabla g(x,y;\psi)$ and $\nabla f(x,y;\xi)$ which are respectively stochastic versions of $\nabla g(x,y)$ and $\nabla f(x,y)$. 
Following the idea of V-PBGD, given iteration $k$ and $x_k$, we first solve the lower-level problem with the stochastic gradient descent (SGD) method: 
\begin{equation}
\omega_{t+1}^{(k)}=\omega_t^{(k)} - \beta_t \nabla_y g(x_k,\omega_t^{(k)};\psi_t^{(k)})\quad\text{for}\quad t=1,\dots,T_k.
\end{equation}
Then we choose the approximate lower-level solution $\hat{y}_k=\omega_{i}^{(k)}$ where $i$ is drawn from a step-size weighted distribution specified by $\mathbf{P}(i=t)=\beta_t/\sum_{t=1}^{T_k} \beta_t,~t=1,...,T_k$.
Given $\hat{y}_k$ and the batch size $M$, $(x_k,y_k)$ is updated with the approximate stochastic gradient of $F_\gamma(x_k,y_k)$ as follows:
$$(x_{k+1},y_{k+1})\!=\!\Proj_{\Zc} \Big((x_k,y_k)-\frac{\alpha_k}{M}\sum_{i=1}^M \big(\nabla f(x_k,y_k;\xi_k^i) \!+\!\gamma(\nabla g(x_k,y_k;\psi_k^i)\!-\!\overline{\nabla}_x g(x_k,\hat{y}_k;\psi_k^i)\big)\Big).$$
The update is summarized in Algorithm \ref{alg:V-PBSGD}.
\begin{algorithm}[t]
% \setstretch{1.2}
\caption{V-PBSGD: Function value gap-based penalized bilevel SGD}
% \vspace{0.2cm}
\begin{algorithmic}[1]
\STATE Select $(x_1,y_1)\in \Zc=\Cc \times \reals^{d_y}$. Select $\gamma$,$K,T_k$,$\alpha_k,\beta_t$ and $M$.
\FOR{$k=1$ {\bfseries to} $K$}
\STATE Choose $\omega_1^{(k)}=y_k$, do $\omega_{t+1}^{(k)}=\omega_t^{(k)} - \beta_t \nabla_y g(x_k,\omega_t^{(k)};\psi_t^{(k)})~\text{for}~t=1,\dots,T_k$
\STATE Choose $\hat{y}_k=\omega_{i}^{(k)}$, $i \sim \mathbf{P}$ where $\mathbf{P}(i=t)=\beta_t/\sum_{t=1}^{T_k} \beta_t,~t=1,...,T_k$.
\STATE $(x_{k+1},y_{k+1})\!=\!\Proj_{\Zc} \Big((x_k,y_k)\!-\!  \frac{\alpha_k}{M}\sum_{i=1}^M \big(\nabla f(x_k,y_k;\xi_k^i) \!+\!\gamma(\nabla g(x_k,y_k;\psi_k^i)\!-\!\overline{\nabla}_x g(x_k,\hat{y}_k;\psi_k^i)\big)\Big)\!.$
% \FOR{$t=1$ {\bfseries to} $T_k$}
% \SATE Update $\omega_{t}^{(k)}$ following \eqref{eq:omega noconvex update}.
% \ENDFOR
\ENDFOR
% \STATE Output $\{x_m,y_m\}$ with $m \in \arg\min_{k\in\{1,...,K\}}\|(x_{k+1},y_{k+1})-(x_k,y_k)\|$.
\end{algorithmic}
\label{alg:V-PBSGD}
\end{algorithm}

We make the following assumption commonly used in the analysis for SGD methods.
\begin{assumption}\label{asp:stochastic grad}
There exists constant $ c>0$ such that given any $k$, the stochastic gradients in Algorithm \ref{alg:V-PBSGD} are unbiased and have variance bounded by $c$.
\end{assumption}
With the above assumption, we provide the convergence result as follows.
\begin{theorem}
Consider V-PBSGD (Algorithm \ref{alg:V-PBSGD}). Assume Assumption \ref{asp:stochastic grad} and those in Theorem \ref{the:nonconvex xy convergence} hold. Choose $\alpha_k =\alpha\leq (L_f + \gamma (2L_g + L_g^2 \mu))^{-1}$, $\beta_t =1/(L_g\sqrt{t})$ and $T_k=T$ for any $k$. It holds that
\begin{align}
    \frac{1}{K}\sum_{k=1}^K\E\norm{G_\gamma (x_k,y_k)}^2 = \mathcal{O}\Big(\frac{1}{\alpha K}\Big)+\mathcal{O}\Big(\frac{\gamma^2  c^2}{M}\Big)+\mathcal{O}\Big(\frac{\gamma^2 \ln T}{\sqrt{T}}\Big).
\end{align}
\end{theorem}
\begin{proof}
\textbf{Convergence of $\omega$ for a given outer-loop index $k$.} We omit the superscription $(k)$ of $\omega_t^{(k)}$ and $\psi_t^{(k)}$ since the proof holds for any outer-loop index $k$. We write $\E_t[\cdot]$ as the conditional expectation given the filtration of samples before iteration $(k,t)$.
By the $L_g$-Lipschitz-smoothness of $g(x,\cdot)$, it holds 
\begin{align}
\label{eq:idk78}
    \E_t[g(x_k,\omega_{t+1})]
    &\leq g(x_k,\omega_t)+\ip{\gr_y g(x_k,\omega_t)}{\E_t[\omega_{t+1}-\omega_t]}+\frac{L_g}{2}\E_t\norm{\omega_{t+1}-\omega_t}^2 \nonumber\\
    &\leq g(x_k,\omega_t)-\beta_t\norm{\gr_y g(x_k,\omega_t)}^2+\frac{L_g \beta_t^2}{2}\E_t\norm{\gr_y g(x_k,\omega_t;\psi_t)}^2
\end{align}
which follows $\gr_y g(x_k,\omega_t;\psi_t)$ is unbiased. 

The last term of \eqref{eq:idk78} can be bounded as
\begin{align}
    \E_t\norm{\gr_y g(x_k,\omega_t;\psi_t)}^2 
    &= \E_t\norm{\gr_y g(x_k,\omega_t;\psi_t)-\gr_y g(x_k,\omega_t)+\gr_y g(x_k,\omega_t)}^2 \nonumber\\
    &= \E_t\norm{\gr_y g(x_k,\omega_t;\psi_t)-\nabla g(x_k,\omega_t)}^2+\norm{\gr_y g(x_k,\omega_t}^2 \nonumber\\
    &\leq  c^2 + \norm{\gr_y g(x_k,\omega_t)}^2.
\end{align}
Substituting the above inequality back to \eqref{eq:idk78} yields
\begin{align}\label{eq:idk79}
\E_t[g(x_k,\omega_{t+1})] &\leq g(x_k,\omega_t)-\frac{\beta_t}{2}\norm{\gr_y g(x_k,\omega_t)}^2+\frac{L_g  c^2}{2}\beta_t^2. \nonumber\\
&\leq g(x_k,\omega_t)-\frac{\beta_t}{2\mu^2}d^2_{\Sc(x_k)}(\omega_t)+\frac{L_g  c^2}{2}\beta_t^2
\end{align}
where the first inequality requires $\beta_t \leq L_g^{-1}$ and the last one follows from the fact that Lipschitz-smooth $1/\mu$-PL function $g(x,\cdot)$ satisfies $1/\mu^2$-error bound  \citep[Theorem 2]{karimi2016linear}. 

We write $\E_k[\cdot]$ as the conditional expectation given the filtration of samples before iteration $k$. Taking $\E_k$ and a telescope sum over both sides of \eqref{eq:idk79} yields
\begin{align}\label{eq:idk80}
    \sum_{t=1}^{T_k}\beta_t\E_k [d^2_{\Sc(x_k)}(\omega_t^{(k)})] 
    &\leq 2\mu^2(g(x_k,\omega_1^{(k)})-g(x_k,\omega_{T_k+1}^{(k)})) + \frac{2 L_g  c^2 \mu^2}{2}\sum_{t=1}^{T_k}\beta_t^2 \nonumber\\
    &\leq 2\mu^2(g(x_k,\omega_1^{(k)})-v(x_k)) +  L_g  c^2 \mu^2\sum_{t=1}^{T_k}\beta_t^2.
\end{align}

\textbf{Convergence of $(x,y)$.} In this proof, we write $z=(x,y)$. Given $z_k$, define $\Bar{z}_{k+1}=\Proj_{\Zc}(z_k-\alpha_k \nabla F_\gamma(z_k))$. For convenience, we also write
\begin{align}
    \overline{\gr}_k F_\gamma &=\frac{1}{M}\sum_{i=1}^M \big(\nabla f(x_k,y_k;\xi_k^i) \!+\!\gamma(\nabla g(x_k,y_k;\psi_k^i)\!-\!\overline{\nabla}_x g(x_k,\hat{y}_k;\psi_k^i)\big)\nonumber\\
   \hat{\gr} F_\gamma(z_k;\hat{y}_k)&=\E_k[\overline{\gr}_k F_\gamma]=\nabla f(x_k,y_k) \!+\!\gamma(\nabla g(x_k,y_k)\!-\!\overline{\nabla}_x g(x_k,\hat{y}_k)).
\end{align}
By the assumptions made in this theorem, $F_\gamma$ is $L_\gamma$-Lipschitz-smooth with $L_\gamma = L_f + \gamma (2L_g + L_g^2 \mu)$.
Then by Lipschitz-smoothness of $F_\gamma$, it holds that
\begin{align}\label{eq:idk45}
    \E_k[F_\gamma(z_{k+1})]
    &\leq F_\gamma(z_k) + \E_k\ip{\gr F_\gamma(z_k)}{z_{k+1}-z_k} + \frac{L_\gamma}{2}\E_k\|z_{k+1}-z_k\|^2 \nonumber\\
    &\stackrel{L_\gamma \leq \frac{1}{\alpha_k}}{\leq} F_\gamma(z_k) + \E_k\ip{\overline{\gr}_k F_\gamma}{z_{k+1}-z_k} +  \E_k\ip{\gr F_\gamma(z_k)-\overline{\gr}_k F_\gamma}{z_{k+1}-\Bar{z}_{k+1}}\nonumber\\
    &\quad+\E_k\ip{\gr F_\gamma(z_k)-\overline{\gr}_k F_\gamma}{\Bar{z}_{k+1}-z_k}+\frac{1}{2\alpha_k}\E_k\|z_{k+1}-z_k\|^2.
\end{align}
Consider the second term in the RHS of \eqref{eq:idk45}.
By Lemma \ref{lem:support 1}, $z_{k+1}$ can be written as
$$z_{k+1}=\arg\min_{z\in\Zc}\ip{\overline{\gr}_k F_\gamma}{z}+\frac{1}{2\alpha_k}\|z-z_k\|^2.$$
By the first-order optimality condition of the above problem, it holds that
\begin{align}
    \ip{\overline{\gr}_k F_\gamma+\frac{1}{\alpha_k}(z_{k+1}-z_k)}{z_{k+1}-z} \leq 0,~\forall z \in \Zc.\nonumber
\end{align}
Since $z_k\in\Zc$, we can choose $z=z_k$ in the above inequality and obtain
\begin{align}\label{eq:idk46}
    \ip{\overline{\gr}_k F_\gamma}{z_{k+1}-z_k} \leq -\frac{1}{\alpha_k}\|z_{k+1}-z_k\|^2.
\end{align}
The third term in the RHS of \eqref{eq:idk45} can be bounded as
\begin{align}
    \label{eq:idk47}
    \E_k\ip{\gr F_\gamma(z_k)-\overline{\gr}_k F_\gamma}{z_{k+1}-\Bar{z}_{k+1}}
    &\leq \E_k [\norm{\gr F_\gamma(z_k)-\overline{\gr}_k F_\gamma}\norm{z_{k+1}-\Bar{z}_{k+1}}] \nonumber\\
    &\leq \alpha_k\E_k \norm{\gr F_\gamma(z_k)-\overline{\gr}_k F_\gamma}^2 
\end{align}
where the second inequality follows from the non-expansiveness of the projection operator. 

The fourth term in the RHS of \eqref{eq:idk45} can be bounded as
\begin{align}
    \label{eq:idk48}
    \E_k\ip{\gr F_\gamma(z_k)-\overline{\gr}_k F_\gamma}{\Bar{z}_{k+1}-z_k}
    &= \ip{\gr F_\gamma(z_k)-\hat{\gr} F_\gamma (z_k;\hat{y}_k)}{\Bar{z}_{k+1}-z_k} \nonumber\\
    &\leq 2\alpha_k \E_k\norm{\gr F_\gamma(z_k)-\hat{\gr} F_\gamma (z_k;\hat{y}_k)}^2 + \frac{1}{8\alpha_k}\norm{\Bar{z}_{k+1}-z_k}^2
\end{align}
where the last inequality follows from Young's inequality.
In addition, we have
\begin{align}
    &\norm{\Bar{z}_{k+1}-z_k}^2 \nonumber\\
    &\leq 2\E_k\norm{\Bar{z}_{k+1}-z_{k+1}}^2+2\E_k\norm{z_{k+1}-z_k}^2 \nonumber\\
    &\leq 2\alpha_k^2\E_k\norm{\gr F_\gamma(z_k)-\overline{\gr}_k F_\gamma}^2+2\E_k\norm{z_{k+1}-z_k}^2 \nonumber\\
    &\leq 4\alpha_k^2\E_k\norm{\gr F_\gamma(z_k)-\hat{\gr} F_\gamma(z_k;\hat{y}_k)}^2+4\alpha_k^2\E_k\norm{\hat{\gr} F_\gamma(z_k;\hat{y}_k)-\overline{\gr}_k F_\gamma}^2+2\E_k\norm{z_{k+1}-z_k}^2 \nonumber
\end{align}
which after rearranging gives
\begin{align}\label{eq:idk49}
    &\E_k\norm{z_{k+1}-z_k}^2 \nonumber\\
    &\geq \frac{1}{2}\norm{\Bar{z}_{k+1}-z_k}^2 - 2\alpha_k^2\E_k\norm{\gr F_\gamma(z_k)-\hat{\gr} F_\gamma(z_k;\hat{y}_k)}^2-2\alpha_k^2\E_k\norm{\hat{\gr} F_\gamma(z_k;\hat{y}_k)-\overline{\gr}_k F_\gamma}^2.
\end{align}
Substituting \eqref{eq:idk46}--\eqref{eq:idk49} into \eqref{eq:idk45} and rearranging yields
\begin{align}\label{eq:idk50}
    \frac{1}{8\alpha_k}\E_k\|\bar{z}_{k+1}-z_k\|^2\
    &\leq F_\gamma(z_k)- \E_k[F_\gamma(z_{k+1})] +  2\alpha_k\E_k \norm{\gr F_\gamma(z_k)-\overline{\gr}_k F_\gamma}^2\nonumber\\
    &\quad+3\alpha_k \E_k\norm{\gr F_\gamma(z_k)-\hat{\gr} F_\gamma (z_k;\hat{y}_k)}^2.
\end{align}
Under Assumption \ref{asp:stochastic grad}, the third term in the RHS of \eqref{eq:idk50} is bounded by the $\mathcal{O}(1/M)$ dependence of variance as follows
\begin{align}
    \label{eq:idk51}
    \E_k \norm{\gr F_\gamma(z_k)-\overline{\gr}_k F_\gamma}^2
    &\leq \frac{3(2\gamma^2+1) c^2}{M}.
\end{align}
The fourth term in the RHS of \eqref{eq:idk50} can be bounded by
\begin{align}\label{eq:idk54}
    \E_k\norm{\gr F_\gamma(z_k)-\hat{\gr} F_\gamma (z_k;\hat{y}_k)}^2
    &= \gamma^2 \E_k \norm{\nabla v(x_k) -\nabla_x g(x_k,\hat{y}_k)}^2 \nonumber\\
   &\!\!\!\!\!\!\stackrel{\rm Lemma ~\ref{lem:grad vx}}{=} \gamma^2 \E_k \norm{\nabla_x g(x_k,y)|_{y\in\Sc(x_k)} -\nabla_x g(x_k,\hat{y}_k)}^2 \nonumber\\
    &\leq \gamma^2 L_g^2 \E_k[d_{\Sc(x_k)}^2(\hat{y}_k)] \nonumber\\
    &=\gamma^2 L_g^2 \sum_{t=1}^{T}\frac{\beta_t \E_k[d_{\Sc(x_k)}^2(\omega_t^{(k)})]}{\sum_{i=1}^T\beta_i}
\end{align}
where the last equality follows from the distribution of $\hat{y}_k$. 

By \eqref{eq:idk80}, it holds that
\begin{align}\label{eq:idk52}
     \sum_{t=1}^{T}\beta_t\E_k [d^2_{\Sc(x_k)}(\omega_t^{(k)})] 
    &\leq 2\mu^2(g(x_k,\omega_1^{(k)})-v(x_k)) +  L_g  c^2 \mu^2\sum_{t=1}^{T}\beta_t^2.
\end{align}
In the above inequality, we can further bound the initial gap as (cf. $\omega_1^{(k)}=y_k$)
\begin{align}\label{eq:idk53}
    g(x_k,\omega_1^{(k)})-v(x_k) \leq \frac{1}{\mu}\|\gr_y g(x_k,\omega_1^{(k)})\|^2 
    &= \frac{1}{\mu}\left\|\frac{y_k-\Bar{y}_{k+1}-\alpha_k \nabla_y f(x_k,y_k)}{\alpha_k\gamma}\right\|^2 \nonumber\\ 
    &\leq \frac{2}{\mu \gamma^2 \alpha_k^2} \|\Bar{z}_{k+1}-z_k\|^2+\frac{2 L^2}{\mu\gamma^2}
\end{align}
where the first inequality follows from $g(x,\cdot)$ is $1/\mu$-PL; the equality follows from the definition of $\Bar{z}_{k+1}$; and the last one follows from Young's inequality and the Lipschitz continuity of $f(x,\cdot)$.

Substituting \eqref{eq:idk52} and \eqref{eq:idk53} into \eqref{eq:idk54} yields
\begin{align}\label{eq:idk265}
     \E_k\norm{\gr F_\gamma(z_k)-\hat{\gr} F_\gamma (z_k;\hat{y}_k)}^2
    % &\leq \frac{4\mu L_g^2}{\alpha_k^2 \sum_{i=1}^T \beta_i^2}\norm{\Bar{z}_{k+1}-z_k}^2+\frac{4\mu L^2 L_g^2}{\sum_{i=1}^T \beta_i}+\gamma^2 L_g^3  c^2 \mu^2 \frac{\sum_{t=1}^{T}\beta_t^2}{\sum_{t=1}^{T}\beta_t} \nonumber\\
    &\leq \frac{1}{48\alpha_k^2 }\norm{\Bar{z}_{k+1}-z_k}^2+\frac{4\mu L^2 L_g^2}{\sum_{i=1}^T \beta_i}+\gamma^2 L_g^3  c^2 \mu^2 \frac{\sum_{t=1}^{T}\beta_t^2}{\sum_{t=1}^{T}\beta_t}
\end{align}
where we have also used $\sum_{i=1}^T \beta_i^2 \geq 192 \mu L_g^2$ to simplify the first term. This can always be satisfied by a large enough $T$.

Substituting \eqref{eq:idk265} and \eqref{eq:idk51} into \eqref{eq:idk50}, rearranging and taking total expectation yield
\begin{align}
    \frac{1}{16\alpha_k}\E\|\bar{z}_{k+1}-z_k\|^2\
    &\leq \E[F_\gamma(z_k)- F_\gamma(z_{k+1})] +  \frac{6(2\gamma^2+1) c^2}{M}\alpha_k\nonumber\\
    &\quad+\Bigg(\frac{12\mu L^2 L_g^2}{\sum_{i=1}^T \beta_i}+3\gamma^2 L_g^3  c^2 \mu^2 \frac{\sum_{t=1}^{T}\beta_t^2}{\sum_{t=1}^{T}\beta_t}\Bigg)\alpha_k.\nonumber
\end{align}
Using $\|\Bar{z}_{k+1}-z_k\|^2=\alpha_k^2 \|G_\gamma(z_k)\|^2$ in the LHS of the above inequality and taking telescope sum over $k=1,\dots,K$ yields
\begin{align}\label{eq:idk57}
    \sum_{k=1}^K \alpha_k \E\|G_\gamma(z_k)\|^2 = \mathcal{O}(F_\gamma(z_1)-C_f) + \mathcal{O}\Big(\frac{\gamma^2  c^2 }{M}\alpha_k\Big)+\Oc\Bigg(\frac{\gamma^2\sum_{t=1}^{T}\beta_t^2}{\sum_{i=1}^{T}\beta_i}\alpha_k\Bigg).
\end{align}
By the choice of step size, we have in the RHS $\sum_{i=1}^{T}\beta_i \geq \sum_{i=1}^{T}\beta_T = \Theta(\sqrt{T})$ and $\sum_{t=1}^{T}\beta_t^2 \leq 1+\int_{1}^T \frac{\beta_1}{x} dx=\beta_1\ln T + 1$. This proves the result. \qed
\end{proof}

\section{Solving Bilevel Problems with Lower-level Constraints}\label{sec:CP}
In the previous section, we have introduced the PBGD method to solve a class of non-convex bilevel problems with only upper-level constraints.
When the lower-level constraints are involved, it becomes more difficult to develop a gradient-based algorithm with finite-time guarantees.

In this section, under assumptions on the lower-level objective that are weaker than the commonly used strong-convexity assumption, we propose an algorithm with a finite-time convergence guarantee.
Specifically, consider the  special case of $\mathcal{BP}$ with a fixed lower-level constraint set $\Uc(x)=\Uc$:
\begin{center}
\begin{tcolorbox}[width=0.8\textwidth]
\leqnomode
\vspace{-0.3cm}
\begin{align}
    \mathcal{CP}:~~\min_{x,y}f(x,y)\,\,\,\,\,\,{\rm s.t.}\,\,\,\,&x \in \Cc,~y \in \arg\!\min_{y\in \Uc}g(x,y) \nonumber
\end{align}
\vspace*{-0.4cm}
\reqnomode
\end{tcolorbox}
\end{center}
where we assume $\Cc$ and $\Uc$ are convex and compact in this section. 
% Assume $f$ and $g$ are continuously differentiable.

\subsection{Penalty reformulation}
Following Section \ref{sec:generic}, we will seek to reformulate $\mathcal{CP}$ with a suitable penalty function $p(x,y)$. In this section, we consider choosing $p(x,y)$ as the lower-level function value gap $g(x,y)-v(x)$ where $v(x)\coloneqq\min_{y\in \Uc}g(x,y)$.
We first list some assumptions that will be repeatedly used in this section.
\begin{assumption}\label{cond:cp}
Consider the following conditions that do not need to hold simultaneously:
\begin{enumerate}[label=(\roman*)]
    \item There exists $\mu>0$ such that $\forall x\in \Cc$, $g(x,\cdot)$ has $\frac{1}{\mu}$-quadratic-growth, that is, $\forall y\in\Uc$, it holds  that
    $$g(x,y)-v(x) \geq \frac{1}{\mu} d^2_{\Sc(x)}(y).$$
    \item  There exists $\Bar{\mu}>0$ such that $\forall x\in\Cc$, 
    $g(x,\cdot)$ satisfies $\frac{1}{\Bar{\mu}}$-proximal-error-bound, i.e., $\forall y\in\Uc$, it holds  
    $$\frac{1}{\beta}\big\|y - \Proj_{\Uc}\big(y-\beta\nabla_y g(x,y)\big)\big\| \geq \frac{1}{\Bar{\mu}} d_{\Sc(x)}(y)$$  where $\beta$ is a constant step size.  
    \item Given any $x\in\Cc$, $g(x,\cdot)$ is convex.
\end{enumerate}
\end{assumption}
 Now we are ready to introduce the following lemma.

\begin{lemma}\label{lemma:QG penalty}
Assume (i) in Assumption \ref{cond:cp} holds.
Then $g(x,y)-v(x)$ is a $\mu$-squared-distance-bound. 
\end{lemma}
The proof is similar to that of Lemma \ref{lem:pl penalty} and thus is omitted.
Given $\gamma > 0$ and $\epsilon>0$, we can define the penalized problem and the approximate bilevel problem of $\mathcal{CP}$ respectively as:

\begin{minipage}{.49\linewidth}
\begin{align}
    ~~\mathcal{CP}_{\gamma p}\!:~&\min_{x,y} F_\gamma (x,y) \!\coloneqq\! f(x,y)\!+\!\gamma \big(g(x,y)-v(x)\big)\nonumber\\
    &~~~{\rm s.t.}~~~x \in \Cc,~y\in\Uc.\nonumber
\end{align}
\end{minipage}
\begin{minipage}{.49\linewidth}
\begin{align}
    \mathcal{CP}_{\epsilon}\!:~&\min_{x,y}f(x,y)\nonumber\\
    &~~~{\rm s.t.}~~~x \!\in\! \Cc,~y\!\in\!\Uc,~g(x,y)\!-\!v(x)\leq \epsilon.\nonumber
\end{align}
\end{minipage}\vspace{0.4cm}

It remains to show that the solutions of $\mathcal{CP}_{\gamma p}$ are meaningful to $\mathcal{CP}$. In the following proposition, we show that the solutions of $\mathcal{CP}_{\gamma p}$ approximately solve $\mathcal{CP}$.
\begin{proposition}[Relation on the local/global solutions]\label{pro:CP solution}
Assume Assumption \ref{asp:Lipschitz continuity} and either of the following holds:
\begin{enumerate}[label=(\alph*)]
    \item Conditions (i) and (ii) in Assumption \ref{cond:cp} hold. Choose $$\gamma \geq \max\left\{L\sqrt{\mu \delta^{-1}},L\sqrt{3\Bar{\mu} \delta^{-1}},3{L}_{1,g} \Bar{\mu}L\delta^{-1}\right\}$$ with ${L}_{1,g}=\max_{x\in\Cc,y\in\Uc}\|\nabla_y g(x,y)\|$ and some $\delta>0$; 
    \item Conditions (i) and (iii) in Assumption \ref{cond:cp} hold. Choose $\gamma \geq L\sqrt{\mu \delta^{-1}}$ with some $\delta>0$.
\end{enumerate}
If $(x_\gamma,y_\gamma)$ is a local/global solution of $\mathcal{CP}_{\gamma p}$, it is a local/global solution of $\mathcal{CP}_{\epsilon_\gamma}$ with some $\epsilon_\gamma \leq \delta$.
\end{proposition}
\begin{proof}
We prove the proposition from the two conditions separately. 

 {\color{black}
 Notice $\mathcal{CP}$ and $\mathcal{CP}_{\epsilon}$ are respectively special cases of $\mathcal{BP}$ and $\mathcal{BP}_{\epsilon}$, we aim to utilize Theorem \ref{the:BPgamr_epsilon_local_relax} to prove the results. Specifically, we aim to prove condition (a) holds for the two cases in this proposition. The idea of the proof is to adapt Proposition \ref{pro:NP solution} to the constrained case. We will use the constrained stationarity of $(x_\gamma,y_\gamma)$ to show that the projected gradient of $g(x_\gamma, \cdot)$ is small under a large $\gamma$. This combined with the proximal error bound assumption will lead to condition (a) in Theorem \ref{the:BPgamr_epsilon_local_relax}. 
 }

\textbf{Proof of Condition (a).} Suppose (a) holds. Given $x\in\Cc$, define the projected gradient of $g(x,\cdot)$ as
$$G(y;x)=\frac{1}{\beta}\big(y-\Proj_{\Uc}(y-\beta\nabla_y g(x,y))\big).$$

Since $y_\gamma$ is a local solution of $\mathcal{CP}_{\gamma p}$ given $x=x_\gamma$, we have
\begin{equation}
    \frac{1}{\beta}\Big[y_\gamma-\Proj_{\Uc}\Big(y_\gamma-\beta\big(\frac{1}{\gamma}\nabla_y f(x,y_\gamma)+\nabla_y g(x_\gamma,y_\gamma)\big)\Big)\Big]=0.
\end{equation}
% $$\frac{1}{\beta}\Big[y_\gamma-\Proj_{\Uc}\Big(y_\gamma-\beta\big(\frac{1}{\gamma}\nabla_y f(x,y_\gamma)+\nabla_y g(x_\gamma,y_\gamma)\big)\Big)\Big]=0.$$
Then we have
\begin{align}
    \|G(y_\gamma;x_\gamma)\| 
    &= \frac{1}{\beta}\Big\|\Proj_{\Uc}\Big(y_\gamma-\beta\big(\frac{1}{\gamma}\nabla_y f(x,y)+\nabla_y g(x_\gamma,y_\gamma)\big)\Big)-\Proj_{\Uc}(y_\gamma-\beta\nabla_y g(x_\gamma,y_\gamma)) \Big\|  \nonumber\\
    &\leq \frac{1}{\gamma}\|\nabla_y f(x,y)\| \leq \frac{L}{\gamma}.
\end{align}
By the proximal error bound inequality, we further have
$$d_{\Sc(x_\gamma)}(y_\gamma) \leq \Bar{\mu}\|G(y_\gamma;x_\gamma)\| \leq  \frac{\Bar{\mu} L}{\gamma}.$$
Since $g$ is continuously differentiable and $\Cc\times\Uc$ is compact, we can define ${L}_{1,g} = \max_{x\in\Cc,y\in\Uc}\|\nabla_y g(x,y)\|$. Then $g(x,\cdot)$ is ${L}_{1,g}$-Lipschitz-continuous on $\Uc$ given any $x\in\Cc$, which yields
\begin{align}\label{eq:idk90}
    p(x_\gamma,y_\gamma)=g(x_\gamma,y_\gamma)-v(x_\gamma)\leq {L}_{1,g} d_{\Sc(x_\gamma)} \leq \frac{{L}_{1,g}\Bar{\mu} L}{\gamma}.
\end{align}
In addition, Lemma \ref{lemma:QG penalty} holds under condition (a) so $p(x,y)$ is a squared distance bound.
Further notice that $\mathcal{CP}$ and $\mathcal{CP}_{\gamma p}$ are special cases of $\mathcal{BP}$ and $\mathcal{BP}_{\gamma p}$ with $\Uc(x)=\Uc$, then the rest of the result follows from Theorem \ref{the:BP_gamr_epsilon} with $\epsilon_1=L\sqrt{\rho\delta}/2$, $\gamma\geq2\gamma^*=L\sqrt{\mu \delta^{-1}}$, $\epsilon_2=0$ and Theorem \ref{the:BPgamr_epsilon_local_relax} where condition (i) holds with \eqref{eq:idk90}.

\textbf{Proof of Condition (b)}. Under condition (b), Lemma \ref{lemma:QG penalty} holds so $p(x,y)$ is a squared distance bound.
Further notice that $\mathcal{CP}$ and $\mathcal{CP}_{\gamma p}$ are special cases of $\mathcal{BP}$ and $\mathcal{BP}_{\gamma p}$ with $\Uc(x)=\Uc$, then the result follows directly from Theorem \ref{the:BP_gamr_epsilon} with $\epsilon_1=L\sqrt{\rho\delta}/2$, $\gamma\geq2\gamma^*=L\sqrt{\mu \delta^{-1}}$, $\epsilon_2=0$ and Theorem \ref{the:BPgamr_epsilon_local_relax} where condition (ii) holds by the convexity of $g(x,\cdot)$. \qed
\end{proof}

\subsection{PBGD under lower-level constraints}
To study the gradient-based method for solving $\mathcal{CP}_{\gamma p}$, it is crucial to identify when $\nabla v(x)$ exists and can be efficiently evaluated. In the unconstrained lower-level case, we have answered this question by introducing Lemma \ref{lem:grad vx}. However, the proof of Lemma \ref{lem:grad vx} relies on a crucial condition that $\nabla_y g(x,y)=0$ for any $y\in\Sc(x)$ which is not necessarily true under lower-level constraints.
In this context, we introduce a Danskin-type theorem next that generalizes Lemma \ref{lem:grad vx}.

\begin{proposition}[Smoothness of $v(x)$]\label{pro:generalized danskin}
    Assume $g(x,y)$ is $L_g$-Lipschitz-smooth; and either (ii) or both (i) and (iii) in Assumption \ref{cond:cp} hold.
    Then $v(x)=\min_{y\in\Uc}g(x,y)$ is differentiable with the gradient
\begin{align}
    \nabla v(x)=\nabla_x g(x,y^*), ~~\forall y^*\in \Sc(x).
\end{align}
Moreover, there exists a constant $L_v$ such that $v(x)$ is $L_v$-Lipschitz-smooth. 
\end{proposition}
The proof for a more general version of Proposition \ref{pro:generalized danskin} can be found in Appendix \ref{app.subsec.c2}. 
\textcolor{black}{Different from the Danskin theorem in \citep{fiacco2020optimal,giovannelli2024bilevel} that builds on certain constraint quantification, this proposition is built on conditions in Assumption \ref{cond:cp} which is necessary for the solution relation results earlier in Proposition \ref{pro:CP solution}.}
Proposition \ref{pro:generalized danskin} suggests one evaluate $\nabla v(x)$ with any solution of the lower-level problem. Given iteration $k$ and $x_k$, it is then natural to run the projected GD method to find one lower-level solution:
\begin{align}\label{eq:omega constrained ll}
\omega_{t+1}^{(k)}=\Proj_{\Uc}\big(\omega_t^{(k)} - \beta \nabla_y g(x_k,\omega_t^{(k)})\big),~~t=1,\dots,T_k.
\end{align}
We can then calculate $\nabla v(x_k) \approx \nabla_x g(x_k,\hat{y}_k)$ with $\hat{y}_k=\omega_{T_k+1}^{(k)}$ and update $(x_k,y_k)$ following \eqref{eq:xy nonconvex update} with $\Zc=\Cc\times \Uc$. The V-PBGD update for the lower-level constrained bilevel problem is summarized in Algorithm \ref{alg:V-PBGD+}.
Next, we provide the convergence result for Algorithm \ref{alg:V-PBGD+}.
% \begin{figure*}[t]
% \centering
%     \includegraphics[width=0.3\textwidth]{figures/toy_graph.pdf}\hspace{0.8cm}
%     \includegraphics[width=0.3\textwidth]{figures/toy_point_spread.pdf}
%      \caption{Left figure: the graph of $z\!=\!f(x,y)$ and $z\!=\!f(x,y)|_{y\in\Sc(x)}$ ; right figure: the plot of the last iterate points generated by $1000$ runs of V-PBGD with random initial points.}
%     \label{fig:name1}
% \end{figure*}

% \begin{figure}[t]
%     \centering
%         \includegraphics[width=0.47\linewidth]{figures/toy_K_vs_gam.pdf}
%         \includegraphics[width=0.495\linewidth]{figures/toy_d_vs_gam.pdf}
%     \caption{Test of V-PBGD with different constant $\gamma$. The figure is generated by running V-PBGD (Algorithm \ref{alg:V-PBGD}) for $K$ steps such that $\|G_\gamma(x_K,y_K)\|^2 \leq 10^{-4}$ given $\gamma$.}
%     \label{fig:name2}
% \end{figure}

\begin{algorithm}[t]
% \setstretch{1.2}
\caption{V-PBGD for bilevel problems with lower-level constraints}
% \vspace{0.2cm}
\begin{algorithmic}[1]
\STATE Select $(x_1,y_1)\in \Zc=\Cc \times \Uc$, step sizes $\alpha,\beta$, penalty constant $\gamma$, iteration numbers $T_k$ and $K$.

\FOR{$k=1$ {\bfseries to} $K$}
\STATE Obtain the auxiliary variable $\hat{y}_k=\omega_{T_k+1}^{(k)}$ by running $T_k$ steps of inner projected GD update \eqref{eq:omega constrained ll}.

\STATE Use $\hat{y}_k$ to approximate $\nabla v(x_k)$ via $\nabla_x g(x_k,\hat{y}_k)$ and update  $(x_k,y_k)$ following \eqref{eq:xy nonconvex update} with $\Zc=\Cc\times \Uc$.
% \FOR{$t=1$ {\bfseries to} $T_k$}
% \SATE Update $\omega_{t}^{(k)}$ following \eqref{eq:omega noconvex update}.
% \ENDFOR
\ENDFOR
% \STATE Output $\{x_m,y_m\}$ with $m \in \arg\min_{k\in\{1,...,K\}}\|(x_{k+1},y_{k+1})-(x_k,y_k)\|$.
\end{algorithmic}
\label{alg:V-PBGD+}
\end{algorithm}

\begin{theorem}\label{the:nonconvex xy convergence+}
Consider V-PBGD with lower-level constraint (Algorithm \ref{alg:V-PBGD+}). Suppose Assumption \ref{asp:Lipschitz continuity}, Assumption \ref{asp:in theorem}, and either (i)\&(ii) or (i)\&(iii) in Assumption \ref{cond:cp} hold.
With a prescribed $\delta\!>\!0$, select
\begin{align}
&\alpha = (L_f + \gamma (L_g + L_v))^{-1},~\beta = L_g^{-1},~\gamma\text{ chosen by Proposition \ref{pro:CP solution}},~T_k =\Omega(\log (\alpha\gamma k)).\nonumber
\end{align}
i) With $C_f\!=\!\inf_{(x,y)\in\Zc} f(x,y)$, it holds that
\begin{equation}
    \frac{1}{K}\sum_{k=1}^K \|G_\gamma(x_k,y_k)\|^2 \leq \frac{8\big(F_\gamma(x_1,y_1)-C_f\big)}{\alpha K}+\frac{3 L_g^2 \mu C_g}{K}.\nonumber
\end{equation}
ii) Suppose $\lim_{k\rightarrow\infty}(x_k,y_k)=(x^*,y^*)$, then $(x^*,y^*)$ is a stationary point of $\mathcal{CP}_{\gamma p}$. If $(x^*,y^*)$ is a local/global solution of $\mathcal{CP}_{\gamma p}$, then it is a local/global solution of $\mathcal{CP}_{\epsilon_\gamma}$ with some $\epsilon_\gamma\leq\delta$.
\end{theorem}
\begin{proof}
\textbf{Convergence of $\omega$.}
     Given any $x\in\Cc$, by the $1/\mu$-quadratic-growth of $g(x,\cdot)$ and \citep[Corrolary 3.6]{drusvyatskiy2018error}, there exists some constant $\Bar{\mu}$ such that the proximal-error-bound inequality holds. Thus under the either condition of Proposition \ref{pro:CP solution}, there exists $\Bar{\mu}>0$ such that $1/\Bar{\mu}$-proximal-error-bound condition holds for $g(x,\cdot)$.
     This along with the Lipschitz-smoothness of $g(x,\cdot)$ implies the proximal PL condition by \citep[Appendix G]{karimi2016linear}. 

     We state the proximal PL condition below. Defining 
      \begin{align}\label{eq:idkproxpl}
     \mathcal{D}(\omega;x)\coloneqq-\frac{2}{\beta} \min_{\omega' \in \Xc}\left\{\ip{\nabla_\omega g(x,\omega)}{\omega'-\omega}+\frac{1}{2\beta}\|\omega'-\omega\|^2\right\}
          \end{align}
     there exists some constant $\Tilde{\mu}>0$ such that
     \begin{align}\label{eq:idk99}
         \tilde{\mu}\mathcal{D}(\omega;x) \geq  (g(x,\omega)-v(x)),~~\forall \omega\in\Uc\text{ and }x\in\Cc.
     \end{align}

We omit index $k$ since the proof holds for any $k$. 
By the Lipschitz gradient of $g(x,\cdot)$, we have
\begin{align}\label{eq.app.pf.c9}
    g(x,\omega_{t+1})  &\leq g(x,\omega_t) +\ip{\gr_y g(x,\omega_t)}{\omega_{t+1}-\omega_t}+\frac{L_g }{2}\|\omega_{t+1}-\omega_t\|^2 \nonumber\\
    &=g(x,\omega_t)-\frac{\beta}{2}\mathcal{D}(\omega_t;x) 
\end{align}
where in the last equality we have used Lemma \ref{lem:support 1} that
$$\omega_{t+1}=\arg\min_{\omega\in\Uc}~\ip{\nabla_y g(x,\omega_t)}{\omega-\omega_t}+\frac{1}{2\beta}\|\omega-\omega_t\|^2.$$
Using \eqref{eq:idk99} in \eqref{eq.app.pf.c9} yields
\begin{align}
     g(x,\omega_{t+1}) - v(x) \leq (1-\frac{\beta }{2\tilde{\mu}})(g(x,\omega_t)-v(x)).\nonumber
\end{align}
Repeatedly applying the last inequality for $t=1,...,T$ yields
$$g(x,\omega_{T+1}) - v(x) \leq (1-\frac{\beta }{2\tilde{\mu}})^{T}(g(x,\omega_1)-v(x)).$$
This along with the $1/\mu$-quadratic-growth property of $g(x,\cdot)$ yields
\begin{align}\label{eq:idk143}
    d^2_{\Sc(x)}(\omega_{T+1})\leq \mu(1-\frac{\beta }{2\tilde{\mu}})^{T}(g(x,\omega_1)-v(x))\leq \mu(1-\frac{\beta }{2\tilde{\mu}})^{T}C_g,~\forall x\in\Cc
\end{align}
where $C_g=\max_{x\in\Cc,y\in\Uc}(g(x,y)-v(x))$ is a constant.

\noindent\textbf{Convergence of $(x,y)$.} The proof is similar to that of Theorem \ref{the:nonconvex xy convergence}. We write the only step that is different here.
In deriving \eqref{eq:error nonconvex}, instead we have
\begin{align}
     \norm{\gr F_\gamma(z_k)-\hat{\gr}F_\gamma (z_k;\hat{y}_k)}^2
    \leq \gamma^2 L_g^2 d^2_{\Sc(x_k)}(\hat{y}_k) &\leq \gamma^2  L_g^2 \mu C_g (1-\frac{\beta }{2\tilde{\mu}})^{T_k} \quad\text{by }\eqref{eq:idk143} \nonumber\\
    &\leq   \frac{L_g^2 \mu C_g}{4\alpha^2 k^2}
\end{align}
where the last inequality requires $T_k \geq -2\log_{c_\beta}(2\alpha \gamma k)~\text{with}~c_\beta=1-\frac{\beta}{2\Tilde{\mu}}$.

Then \eqref{eq:bb01} is replaced with
\begin{align}
     \norm{\gr F_\gamma(z_k)-\hat{\gr}F_\gamma (z_k;\hat{y}_k)}^2
    \leq \frac{1}{4\alpha}\|z_{k+1}-z_k\|^2+\frac{L_g^2 \mu C_g}{4\alpha k^2}.
\end{align}
Result i) in this theorem then follows from the rest of the proof of i) in Theorem \ref{the:nonconvex xy convergence}. Result ii) in this theorem follows similarly from the proof of ii) in Theorem \ref{the:nonconvex xy convergence} under Proposition \ref{pro:CP solution}. \qed
\end{proof}

Theorem \ref{the:nonconvex xy convergence+} implies an iteration complexity of $\widetilde{\mathcal{O}}(\gamma\epsilon^{-1})$ to find an  $\epsilon$-stationary-point of $\mathcal{CP}_{\gamma p}$. This recovers the iteration complexity of the projected GD method \citep{nesterov2013gradient} with a smoothness constant of $\Theta(\gamma)$.
If we choose $\delta=\epsilon$, then the iteration complexity is $\widetilde{\mathcal{O}}(\epsilon^{-1.5})$ under Condition (b) with $\gamma={\cal O}(\delta^{-0.5})={\cal O}(\epsilon^{-0.5})$ or $\widetilde{\mathcal{O}}(\epsilon^{-2})$ under Condition (a) in Proposition \ref{pro:CP solution} with $\gamma={\cal O}(\epsilon^{-1})$.

\section{Solving Bilevel Problems via Nonsmooth Penalization}\label{sec:nonsmooth penalization}
In this section, we consider the bilevel problem with unconstrained lower-level problem $\mathcal{UP}$ defined in Section \ref{sec:NP}. In the previous discussion, the key obstacle that prevents PBGD from achieving the optimal complexity of ${\cal O}(\epsilon^{-1})$ is the escalating penalty constant $\gamma$. This arises from the fact that the penalized problem $\mathcal{UP}_{\gamma p}$ can only approximate $\mathcal{UP}$ within an $\epsilon$ error. As a possible solution to this issue, we introduce an alternative penalty function in this section. 

\subsection{Penalty reformulation}
Different from the penalty functions in \eqref{eq:penalty-all}, consider the following penalty function, which is not necessarily a square-distance bound function 
\begin{center}
\begin{tcolorbox}[width=0.6\textwidth]
\vspace{-0.4cm}
\begin{align}
\label{eq:nonsmooth-penalty}
  p(x,y)\!=\!\|\nabla_y g(x,y)\|.
\end{align}
\end{tcolorbox}
\end{center}
Then we can define the penalized problem of the bilevel problem $\mathcal{UP}$ as
\begin{align}\label{opt-penalty}
\mathcal{UP}_{\gamma p}:~~~~\min_{x\in\Cc,y}\tilde F_\gamma(x,y)\coloneqq~~ f(x,y)+\gamma \|\nabla_y g(x,y)\|
\end{align}
where we reuse the notation $\mathcal{UP}_{\gamma p}$ defined earlier, but here $p(x,y)$ is specified as \eqref{eq:nonsmooth-penalty}.
The advantage of employing this penalty term, as we will demonstrate next, is that a constant penalty parameter $\gamma={\cal O}(1)$ is able to ensure the equivalence between the penalized problem $\mathcal{UP}_{\gamma p}$ and the original bilevel problem $\mathcal{UP}$. However, a drawback of this approach is the resultant nonsmoothness.
 
Following Section \ref{sec:NP}, we make the PL assumption of $g(x,\cdot)$. Benefiting from the penalty function defined in \eqref{eq:nonsmooth-penalty}, we have the following exact penalty theorem for bilevel optimization. 
\begin{proposition}[Relation on local/global solutions]\label{exact-penalty-nonsmooth}
Suppose Assumptions \ref{asp:Lipschitz continuity} and \ref{def:pl} hold with parameters $L$ and $\mu$, and $g(x,\cdot)$ is Lipschitz-smooth. For any $\gamma> L\sqrt{\mu}$, the following arguments hold 
\begin{enumerate}[label=(\roman*)]
\item The global solutions of $\mathcal{UP}_{\gamma p}$ are the global solutions of the original bilevel problem $\mathcal{UP}$. 
\item If $\|\nabla_y g(x,y)\|$ is convex in $y$, then the local solutions of $\mathcal{UP}_{\gamma p}$ are the local solutions of $\mathcal{UP}$.
\end{enumerate}
\end{proposition}
The proof of Proposition \ref{exact-penalty-nonsmooth} is deferred in Appendix \ref{app.pf.prop6}. 
Compared to the relation between $\mathcal{UP}_{\gamma p}$ and $\mathcal{UP}_{\epsilon}$ in Proposition \ref{pro:NP solution}, the penalized problem $\mathcal{UP}_{\gamma p}$ in \eqref{opt-penalty} with nonsmooth penalization is equivalent to the original problem $\mathcal{UP}$ rather than $\mathcal{UP}_\epsilon$ so that the penalty parameter $\gamma$ would not depend on $\epsilon$. For the local relation in (ii) , the assumption that $\|\nabla_y g(x,y)\|$ is convex can be satisfied in cases when the lower-level objective function $g$ is quadratic, e.g., $g(x,y)=x^\top y+y^\top A y$ where $A\in\reals^{|d_y|\times|d_y|}$.

\subsection{The Prox-linear algorithm}

Solving the penalized problem \eqref{opt-penalty} is not easy since it involves a nonsmooth penalty term $\|\nabla_y g(x,y)\|$, originating from the nonsmoothness of the Euclidean norm $\|\cdot\|$. While it is possible to smooth this term using the Moreau envelope or zeroth-order approximation, these methods can result in slowing down the convergence rate due to smoothing errors. In fact, the gradient norm square in \eqref{eq:gry g} could also be considered as a smooth approximation of \eqref{eq:nonsmooth-penalty}. Given the unique characteristics of the Euclidean norm $\|\cdot\|$, we can solve \eqref{opt-penalty} by the Prox-linear algorithm. 

The Prox-linear algorithm \citep{drusvyatskiy2019efficiency} is a recently developed algorithm tailored for solving the composite nonsmooth problem in the following form
\begin{align}
\min_{z\in\mathcal{Z}}~~ c_1(z)+c_3(c_2(z))
\end{align}
where $\mathcal{Z}$ is closed and convex set, $c_1:\mathbb{R}^m\rightarrow \mathbb{R}$ and $c_2:\mathbb{R}^m\rightarrow \mathbb{R}^n$ are Lipschitz smooth and $c_3:\mathbb{R}^n\rightarrow \mathbb{R}$ is convex but nonsmooth. In the context of the penalized bilevel problem $\mathcal{UP}_{\gamma p}$ in \eqref{opt-penalty}, we concatenate $(x,y)$ as $z=(x,y)$, and choose $c_1$ as  $f(\cdot)$, $c_2$ as $\nabla_y g(\cdot)$ and $c_3$ as the Euclidean norm $\gamma\|\cdot\|$. 

At each iteration $k$, we linearize $f(z)$ and $\nabla_y g(z)$ and add a regularization to define a surrogate function of $\tilde F_\gamma(z)$ at $z^k=(x^k,y^k)$, given by 
 \begin{align}\label{proxlinear-inner}
\ell(z;z^k):=f(z^k)+\nabla f(z^k)^\top(z-z^k)+\gamma\left\|\nabla_y g(z^k)+\nabla(\nabla_y g(z^k))^\top(z-z^k)\right\|+\frac{1}{2\lambda}\|z-z^k\|^2
\end{align}
where $\nabla(\nabla_y g(z^k)):=\left[\begin{array}{cc}\nabla_{xy} g(x^k,y^k)\\\nabla_{yy} g(x^k,y^k)\end{array}\right]$ . Clearly, the surrogate function $\ell(z;z^k)$ is strongly convex on $z$. Then the Prox-linear method solves the subproblem
\begin{align}
z^{k+1}=\argmin_{z\in \Zc}\ell(z;z^k).\label{opt-sub}
\end{align}
We adopt the inexact version of the Prox-linear method \citep{drusvyatskiy2019efficiency}, which solves each subproblem at a certain accuracy. Given a tolerance $\delta$, a point $z\in \Zc$ is said to be an $\delta$ approximate solution of a function $\ell:\mathbb{R}^d\rightarrow\mathbb{R}$ if it is $\delta$-global-optimum in terms of Definition \ref{def_approx_global}.  Therefore, the inexact Prox-linear based bilevel algorithm is summarized in Algorithm \ref{alg:inexact}.

\begin{algorithm}[t]
\caption{Penalty-based Prox-linear (PBPL) method}
\begin{algorithmic}[1]
\STATE Initialization $z^0=\{x^0,y^0\}$, $t>0$ and a sequence of $\{\delta_k\}_{k=0}^{K-1}$
\FOR{$k=0$ {\bfseries to} $K-1$}
\STATE Set $z^{k+1}=(x^{k+1},y^{k+1})$ be an $\delta_k$-approximate solution of the subproblem in \eqref{opt-sub} with $z^k=(x^k,y^k)$. 
\ENDFOR
\end{algorithmic}
\label{alg:inexact}
\end{algorithm}

\subsection{Convergence analysis of PBPL}
To analyze the convergence of PBPL, we first introduce an assumption parallel to Assumption \ref{asp:in theorem}. This assumption adds the smoothness assumption of $\nabla_y g(x,y)$, which is also standard in the convergence analysis of the gradient-based bilevel optimization methods \citep{chen2021tighter,grazzi2020iteration}.
\begin{assumption}[Smoothness]\label{asp:in theorem2}
There exist   constants $L_f$, $L_g$ and $L_{g,2}$ such that $f(x,y)$,   $g(x,y)$  and $\nabla_y g(x,y)$ are respectively $L_f$-Lipschitz-smooth, $L_g$-Lipschitz-smooth and $L_{g,2}$-Lipschitz-smooth in $(x,y)$.
\end{assumption}

The commonly used stationary metric in nonsmooth analysis \citep{drusvyatskiy2019efficiency} is the prox-gradient mapping, which is defined as 
\begin{align*}
\mathcal{G}_\lambda(x,y):=\lambda^{-1}\left((x,y)-\argmin_{(x^\prime, y^\prime)\in\mathcal{Z}} \ell((x^\prime, y^\prime) ; (x,y))\right).
\end{align*}
Then the convergence rate of the PBPL algorithm is stated in the following theorem. 

\begin{theorem}\label{thm:inexact-ProxL}
Suppose Assumption \ref{asp:Lipschitz continuity},\ref{def:pl} and \ref{asp:in theorem2} hold and $\lambda \leq\frac{1}{L_f+\gamma L{g,2}}$. Defining $C_f\!=\!\inf_{(x,y)\in\mathcal{Z}} f(x,y)$, the iterates $(x^k,y^k)$ generated by the PBPL algorithm in Algorithm \ref{alg:inexact} satisfy
\begin{align*}
\min _{k=0, \ldots, K-1}\left\|\mathcal{G}_\lambda(x^k,y^k)\right\|^2 \leq \frac{2 \lambda^{-1}\left(\tilde F_\gamma(x^0,y^0)-C_f+\sum_{k=0}^{K-1} \delta_k\right)}{K}.
\end{align*}
\end{theorem}
\begin{proof}
Before we establish the convergence, we first prove a useful inequality as follows.

Since $\|\cdot\|$ is $1$-Lipschitz, we have
\begin{align*}
&~~~~~~~~|\tilde F_\gamma(z)-\ell(z ; w)+\frac{1}{2\lambda}\|z-w\|^2\||\\
&=
\Big|\tilde F_\gamma(z)-\left(f(w)+\nabla f(w)^\top(z-w)+\gamma\|\nabla_y g(w)+\nabla(\nabla_y g(w))^\top(z-w)\|\right)\Big| \\
&\leq \|f(z)-(f(w)+\nabla f(w)^\top(z-w))\|+\gamma\|\nabla_y g(z)-(\nabla_y g(w)+\nabla(\nabla_y g(w))^\top(z-w))\|\\
&\leq \frac{L_f+\gamma L_{g,2}}{2}\|z-w\|^2
\end{align*}
where the first equality is according to the definition, and the last inequality comes from the smoothness of $f$ and $\nabla_y g$. As a result, we have
\begin{align*}
-\frac{L_f+\gamma L_{g,2}}{2}\|z-w\|^2\leq \tilde F_\gamma(z)-\ell(z ; w)+\frac{1}{2\lambda}\|z-w\|^2 \leq \frac{L_f+\gamma L_{g,2}}{2}\|z-w\|^2. 
\end{align*}
Rearranging terms yields 
\begin{align}\label{eq:difference-surrogate}
-\frac{L_f+\gamma L_{g,2}+\lambda^{-1}}{2}\|z-w\|^2\leq \tilde F_\gamma(z)-\ell(z ; w) \leq \frac{L_f+\gamma L_{g,2}-\lambda^{-1}}{2}\|z-w\|^2.
\end{align}
Now we are ready to establish the convergence.
Let us denote $z=(x,y)$ and $z^k_*=\argmin_{z\in\mathcal{Z}}\ell\left(z ; z_k\right)$. Then we have $\mathcal{G}_\lambda\left(z_k\right)=\lambda^{-1}\left(z^k-z^k_*\right)$. Considering $\ell\left(\cdot ; z_k\right)$ is $(1 / t)$- strongly convex, we have
\begin{align*}
\tilde F_\gamma(z^k)=\ell(z^k ; z^k) \geq \ell\left(z^k_* ; z^k\right)+\frac{\lambda}{2}\left\|\mathcal{G}_\lambda(z^k)\right\|^2 \geq \ell\left(z^{k+1} ; z^k\right)-\delta_{k}+\frac{\lambda}{2}\left\|\mathcal{G}_\lambda(z^k)\right\|^2 .
\end{align*}
Then inequality \eqref{eq:difference-surrogate} with $\lambda^{-1} \geq L_f+\gamma L_{g,2}$ implies that $\ell(z ; z^k)\geq\tilde F_\gamma(z)$ and thus,
\begin{align}\label{diff-step}
\tilde F_\gamma(z^k)\geq \tilde F_\gamma(z^{k+1})-\delta_{k}+\frac{\lambda}{2}\left\|\mathcal{G}_\lambda(z^k)\right\|^2 .
\end{align}
Telescoping \eqref{diff-step} yields
\begin{align*}
\min _{k=0, \ldots, K-1}\left\|\mathcal{G}_\lambda\left(x_k\right)\right\|^2 \leq \frac{1}{N} \sum_{k=0}^{K-1}\left\|\mathcal{G}_\lambda(z^k)\right\|^2 & \leq \frac{2 \lambda^{-1}\left(\sum_{k=0}^{K-1} \tilde F_\gamma(z_k)-\tilde F_\gamma(z_{k+1})+\sum_{k=0}^{K-1} \delta_{k}\right)}{K} \\
& \stackrel{(a)}{\leq} \frac{2 \lambda^{-1}\left(\tilde F_\gamma(x^0,y^0)-C_f+\sum_{k=0}^{K-1} \delta_k\right)}{K},
\end{align*}
where (a) holds because
$\tilde F_\gamma(z)\geq F(z)$ implies $\inf_{z\in\mathcal{Z}} \tilde F_\gamma(z)\geq \inf_{z\in\mathcal{Z}} F(z)=C_f$. \qed
\end{proof}

\noindent\emph{Discussion on Theorem \ref{thm:inexact-ProxL}.} If the errors $\left\{\delta_k\right\}_{k=0}^{\infty}$ are summable (e.g. $\delta_k \sim \frac{1}{k^{q}}$ with $q>1$), one has
$$
\min _{k=0, \ldots, K-1}\left\|\mathcal{G}_\lambda(x^k,y^k)\right\|^2 = {\cal O}\left(\frac{1}{K}\right).
$$
The overall convergence rate of PBPL is achieved by the product of ${\cal O}\left(\frac{1}{K}\right)$ and the complexity of the subroutine in solving the subproblem \eqref{opt-sub}. If the subroutine is linearly convergent, then to achieve an $\epsilon$ stationary point, PBPL requires $\tilde{\cal O}\left(\epsilon^{-1}\right)$ iterations. Given that the subproblem \eqref{opt-sub} incorporates a differentiable quadratic term alongside an $\ell_1$ norm of matrix multiplication (representing the nonsmooth term) with respect to $z$, when the lower-level problem has a special structure, the proximal operator of the nonsmooth component has the closed-form expression. Consequently, one can employ the proximal gradient descent algorithm to solve \eqref{opt-sub} efficiently; see e.g., \citep{beck2009fast}. However, developing efficient solvers for \eqref{opt-sub} is beyond the scope of this paper. 

% As the subproblem in \eqref{opt-sub} is a composite of quadratic and linear problem, it is likely to be solved by ADMM with linear convergence rate \citep{deng2016global}. 

\section{Simulations}\label{sec:simulation}
In this section, we first verify our main theoretical results on V-PBGD and PBPL in the toy problems and then compare the PBGD\footnote{The code is available on GitHub (\href{https://github.com/hanshen95/penalized-bilevel-gradient-descent}{https://github.com/hanshen95/penalized-bilevel-gradient-descent}).} algorithm with several other baselines on the data hyper-cleaning task.

\subsection{Numerical verification}

We first verify the results on V-PBGD by considering the following non-convex bilevel problem:
\begin{align}\label{eq:toy_nc}
    &\min_{x,y}~~f(x,y)=\frac{\cos(4y+2)}{1+e^{2-4x}}+\frac{1}{2}\ln((4x-2)^2+1)\\
   & ~{\rm s.t.}~~~~~x \in[0,3],~y \in \arg\min_{y\in \reals} g(x,y)=(y+x)^2+x\sin^2(y+x)\nonumber
\end{align}
which is an instance of the lower-level unconstrained bilevel problem $\mathcal{UP}$ studied in Section \ref{sec:NP}.
It can be verified that the assumptions in Propositions \ref{pro:NP global solution}, \ref{pro:NP solution} and Theorem \ref{the:nonconvex xy convergence} are satisfied.

\begin{figure*}[t]
    \centering
    \begin{subfigure}[t]{0.49\textwidth}
        \centering
        \includegraphics[width=0.493\textwidth]{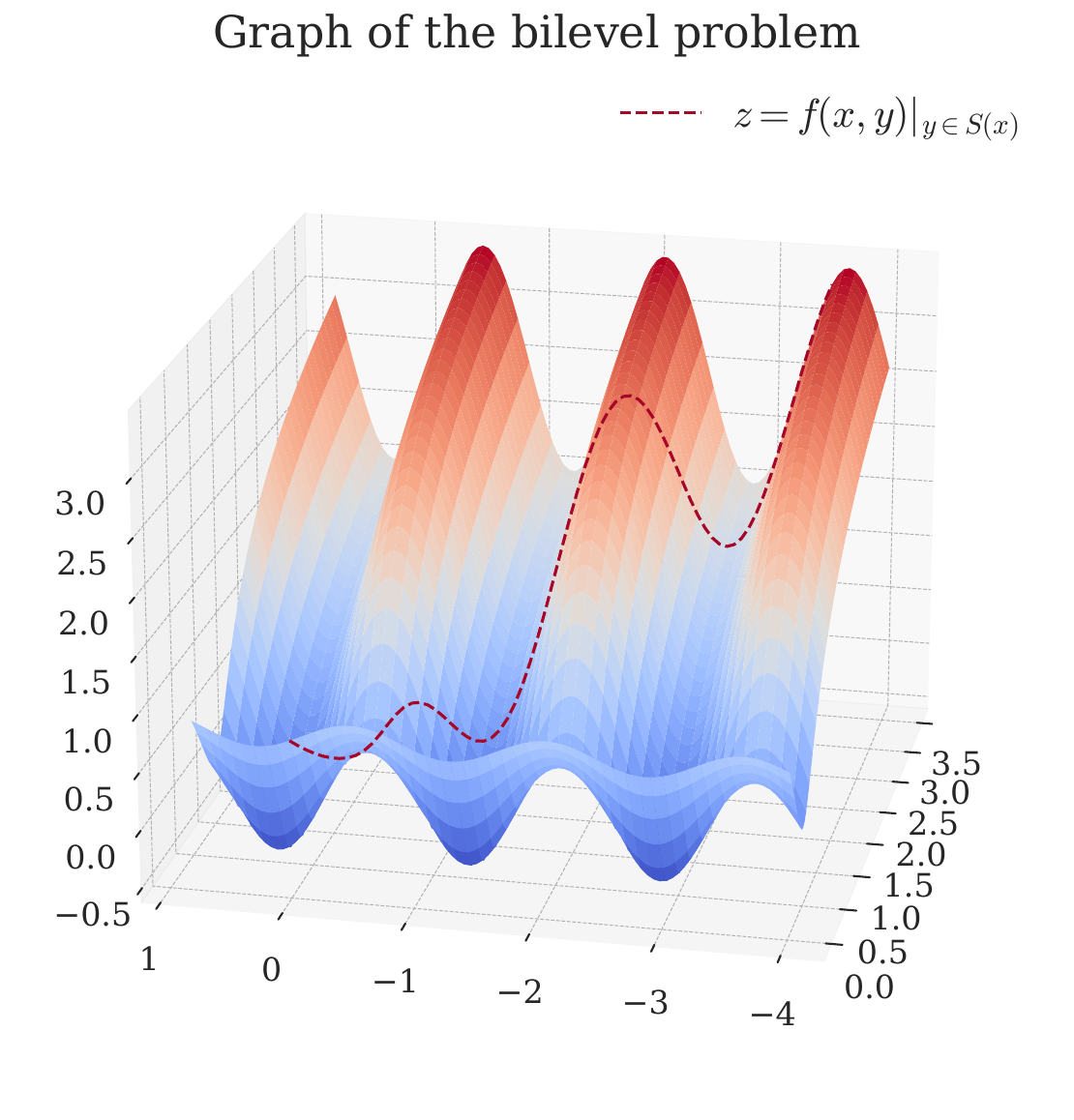}
    \includegraphics[width=0.493\textwidth]{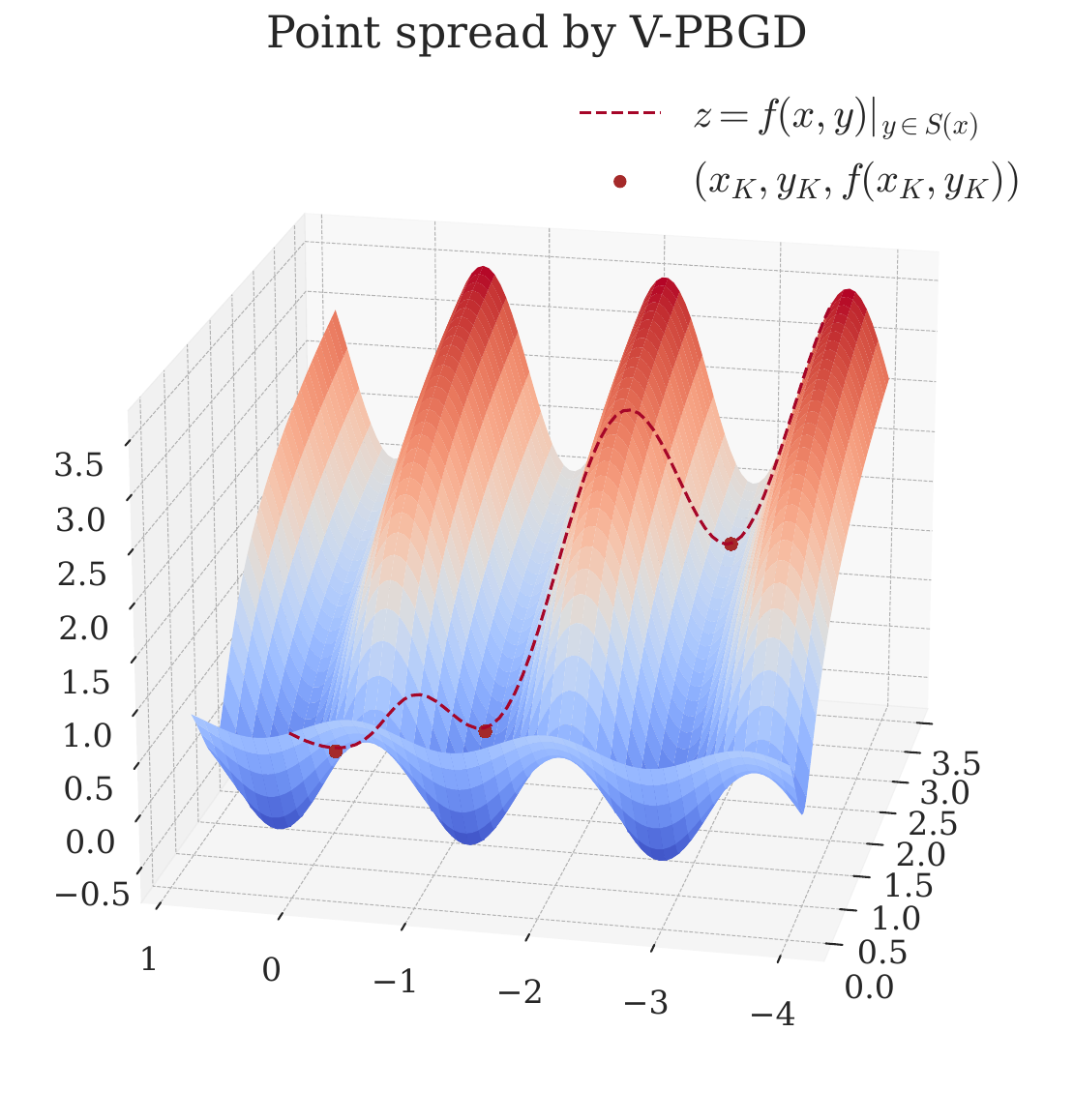}
        \caption{Left figure: the graph of $z\!=\!f(x,y)$ and $z\!=\!f(x,y)|_{y\in\Sc(x)}$; right figure: the last iterates generated by $1000$ runs of V-PBGD with random initial points.} \label{fig:name1}
    \end{subfigure}
    \hfill
    \begin{subfigure}[t]{0.49\textwidth}
        \centering
        \includegraphics[width=0.47\linewidth]{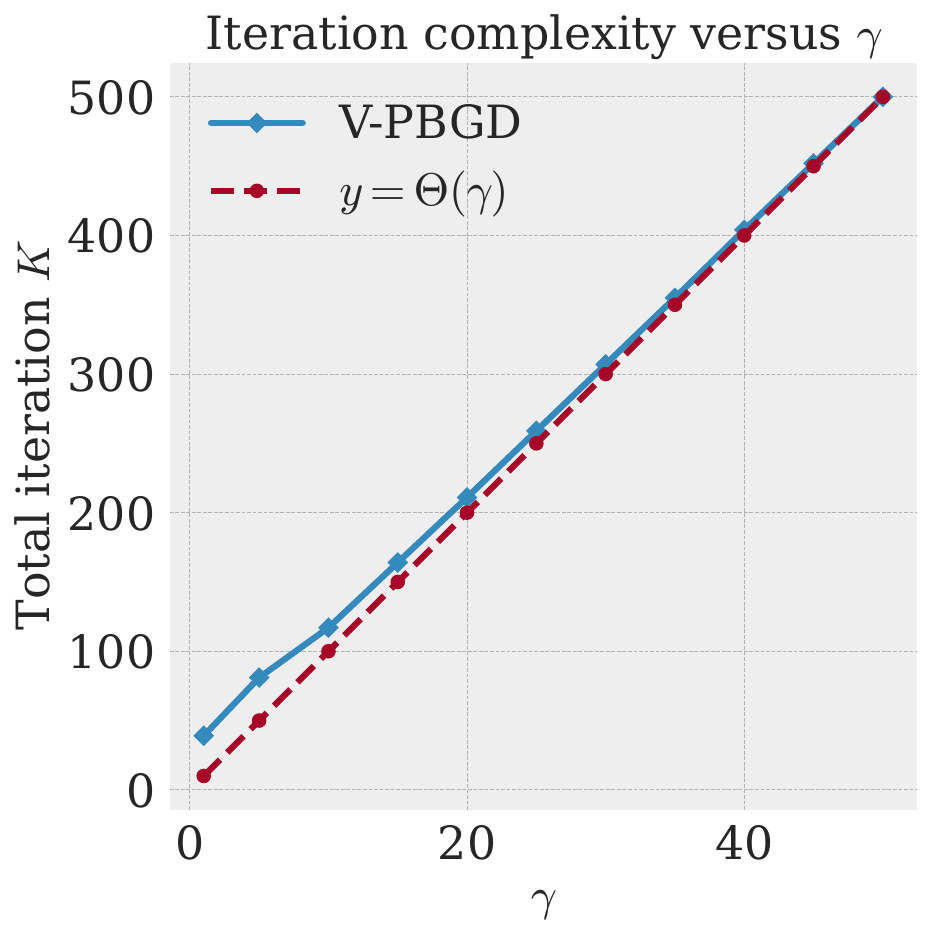}
        \includegraphics[width=0.495\linewidth]{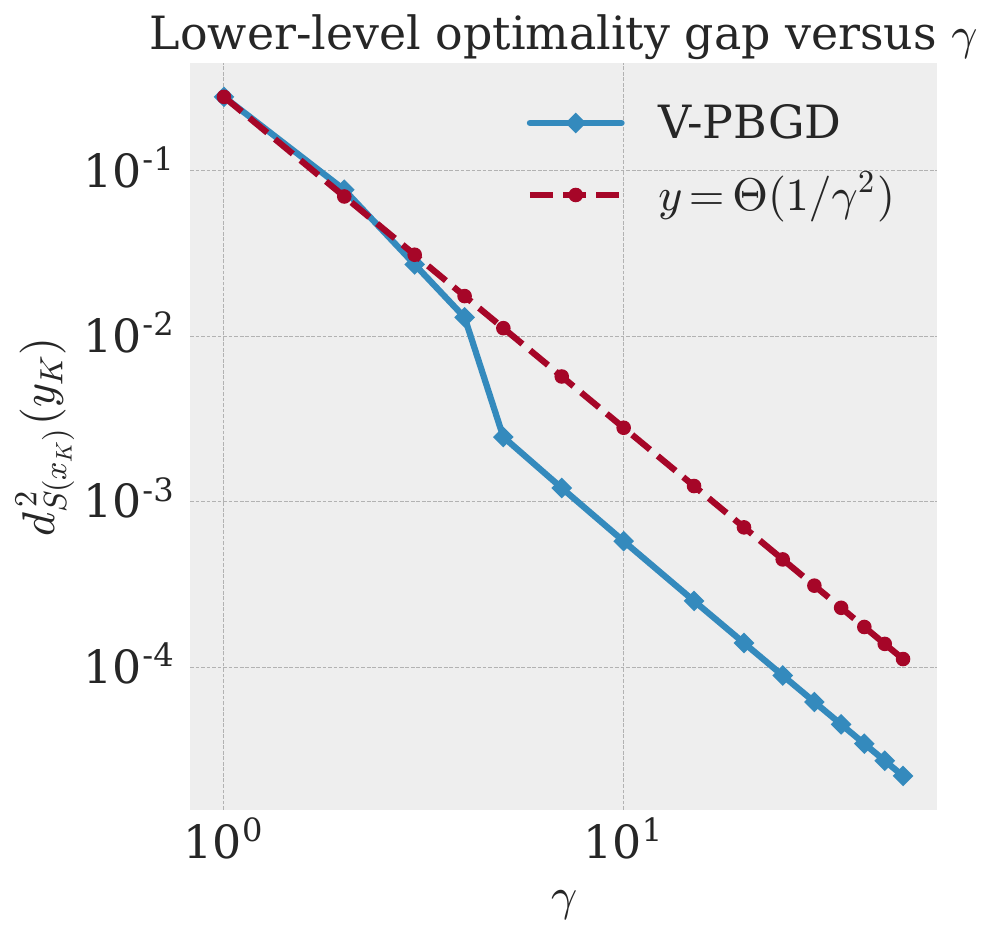}
        \caption{Test of V-PBGD with different constant $\gamma$. The figure is generated by running V-PBGD (Algorithm \ref{alg:V-PBGD}) for $K$ steps such that $\|G_\gamma(x_K,y_K)\|^2 \leq 10^{-4}$. } \label{fig:name2}
    \end{subfigure}
    \caption{Test of V-PBGD in the bilevel problem \eqref{eq:toy_nc}.}
    % \vspace*{-0.1cm}
    \label{fig:name1name2}
\end{figure*}

We plot the graph of $z=f(x,y)$ in Figure \ref{fig:name1} (left). Notice that given any $x\in[0,3]$, we have $\Sc(x)=\{-x\}$. Thus the the bilevel problem in \eqref{eq:toy_nc} can be reduced to $\min_{x\in\Cc} f(x,y)|_{y=-x}$. We plot the single-level objective function $z=f(x,y)|_{y=-x}$ in Figure \ref{fig:name1} (left) as the intersected line of the surface $z=f(x,y)$ and the plane $y=-x$. We then run V-PBGD with $\gamma=10$ for $1000$ random initial points $(x_1,y_1)$ and plot the last iterates in Figure \ref{fig:name1} (right). It can be observed that V-PBGD consistently finds the local solutions of the bilevel problem \eqref{eq:toy_nc}. 
Then we test the impact of $\gamma$ on the performance of V-PBGD, and report the results in Figure \ref{fig:name2}. From Figure \ref{fig:name2}, the iteration complexity is $\Theta(\gamma)$, while the lower-level accuracy is $\Theta(1/\gamma^2)$,   consistent with Theorem \ref{the:nonconvex xy convergence}. 

\begin{figure*}[t]
    \centering
    \begin{subfigure}[t]{0.37\textwidth}
        \centering
    \includegraphics[width=0.85\textwidth]{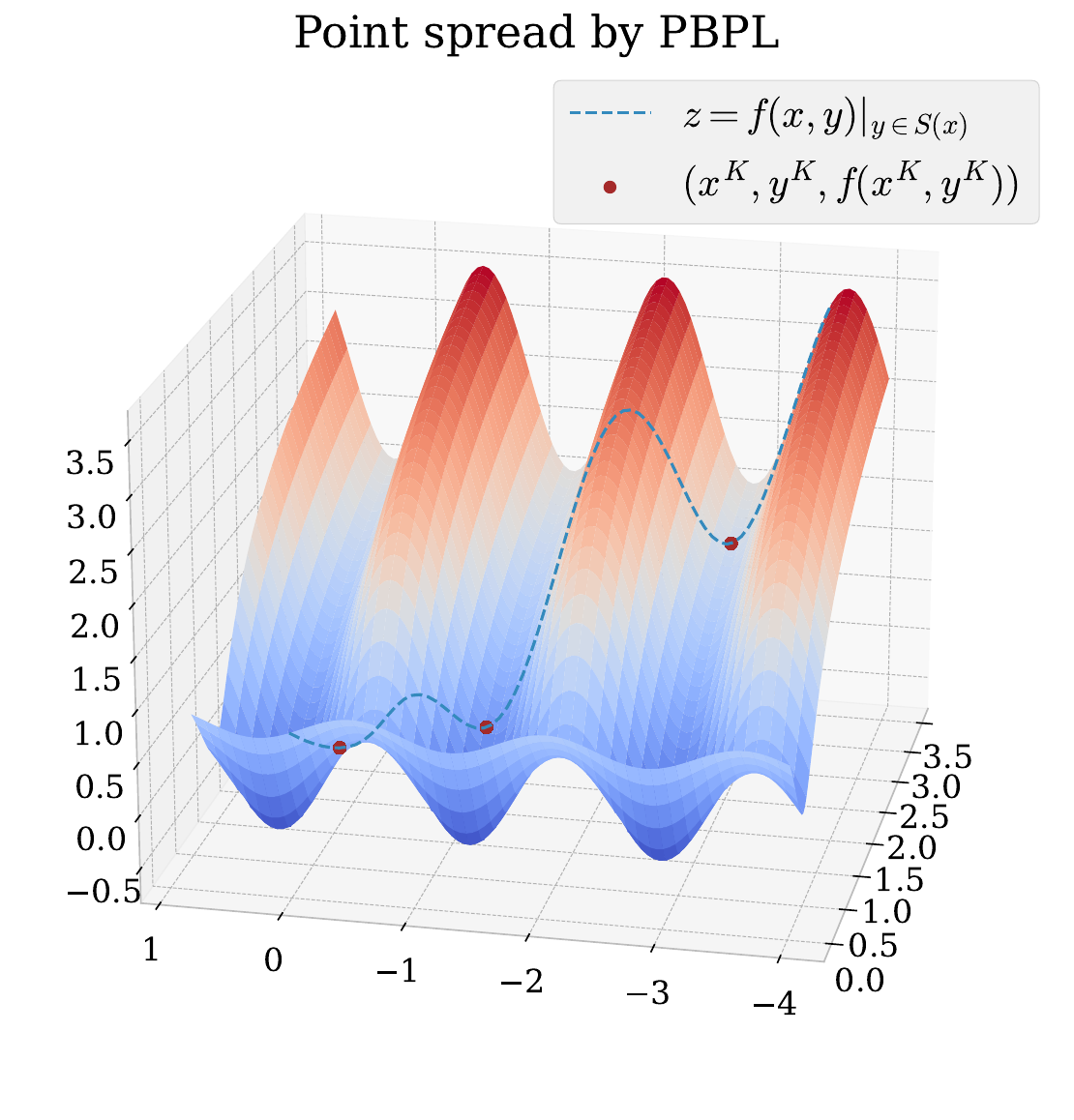}
        \caption{The last iterates generated by $1000$ runs of PBPL with random initial points.}\label{fig:name1pbpl}
    \end{subfigure}
    \hfill
    \begin{subfigure}[t]{0.59\textwidth}
        \centering
        \includegraphics[width=0.45\linewidth]{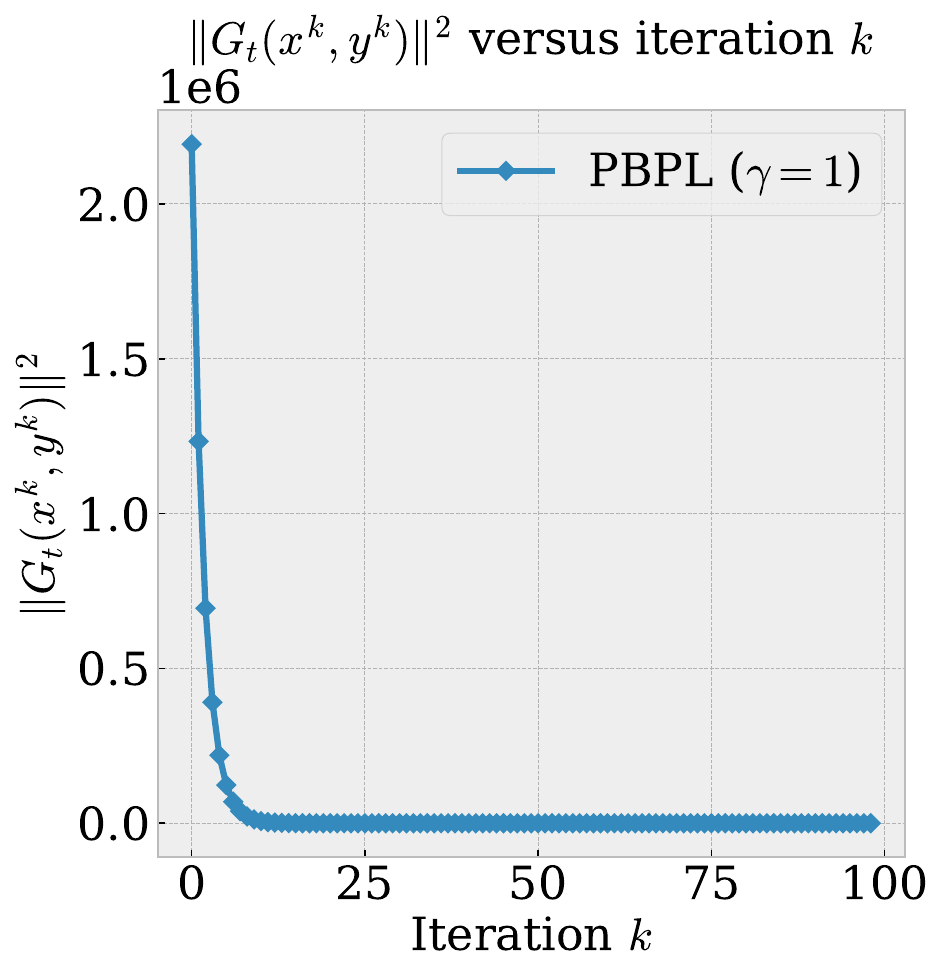}
        \includegraphics[width=0.496\linewidth]{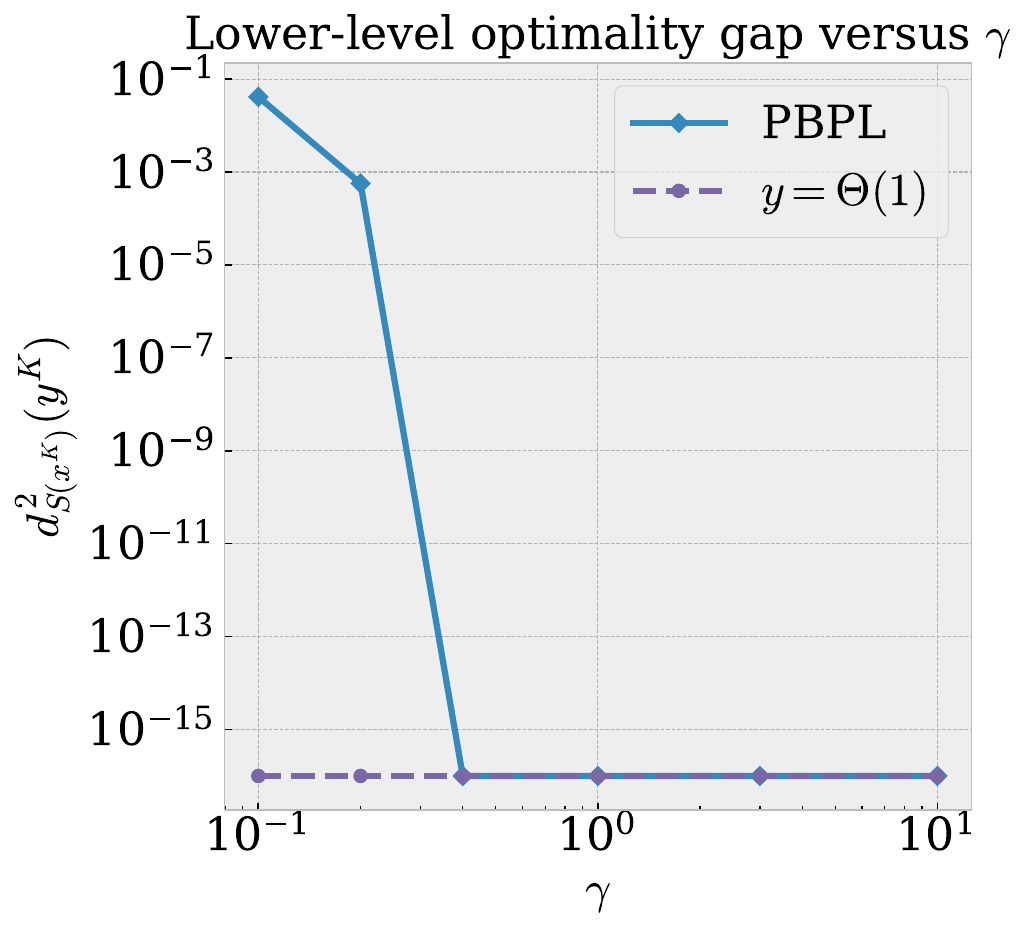}
        \caption{{\color{black}The figure is generated by running PBPL for $K$ steps such that $\|\mathcal{G}_t(x^K,y^K)\| \leq 10^{-4}$. In the log-log plot, we truncate the $y$-value below $10^{-16}$ and treat it as zero.}} \label{fig:name2pbpl}
    \end{subfigure}
    \caption{Test of PBPL in the bilevel problem \eqref{eq:toy_nc_pbpl}.}
    % \vspace*{-0.1cm}
    \label{fig:name1name2pbpl}
\end{figure*}

{\color{black}
Next we verify the PBPL algorithm introduced in Section \ref{sec:nonsmooth penalization} by considering the following problem:
\begin{align}\label{eq:toy_nc_pbpl}
    &\min_{x,y}~~f(x,y)=\frac{\cos(4y+2)}{1+e^{2-4x}}+\frac{1}{2}\ln((4x-2)^2+1)\\
   & ~{\rm s.t.}~~~~~x\in[0,3],~~y \in \arg\min_{y\in \reals} g(x,y)=(y+x)^2.\nonumber
\end{align}
It is straightforward to check that the assumptions in Proposition \ref{exact-penalty-nonsmooth} and Theorem \ref{thm:inexact-ProxL} are satisfied.

Similar to the previous case, we plot the graph of $z=f(x,y)$ and the single-level objective function $z=f(x,y)|_{y=-x}$ in Figure \ref{fig:name1pbpl} as the intersected line of the surface $z=f(x,y)$ and the plane $y=-x$.
Observing from Figure \ref{fig:name2pbpl} (left), PBPL successfully converges in a few iterations.
Thus we can run PBPL with $\gamma=1$ for $1000$ random initial points $(x^1,y^1)$ and plot the last iterates in Figure \ref{fig:name1pbpl}. It can be seen that PBPL converges to the local solutions of  \eqref{eq:toy_nc_pbpl} starting from the sampled initial points.

We also test the impact of $\gamma$ on the performance of PBPL and report the results in Figure \ref{fig:name2pbpl} (right). It can be observed from Figure \ref{fig:name2pbpl} (right) that to reach a zero lower-level optimality gap, we only need a penalty constant of $\gamma = \Theta(1)$ ($\gamma \geq 0.3$ in this example), which is consistent with Proposition \ref{exact-penalty-nonsmooth}. 
Since $\gamma$ needs only to be larger than a constant threshold for exact lower-level accuracy, we only plot $\gamma=1$ in Figure \ref{fig:name2pbpl} (left), and other choices of $\gamma$ will lead to similar convergence behavior.
% Due to this reason Since the penalty is exact in this case, the step size in this simulation can be chosen as a constant independent of the lower-level target accuracy. 
While PBPL works well in this toy example, it should be noted that it may be hard to implement PBPL when the subproblem \eqref{proxlinear-inner} does not have a closed-form solution. 
}

\subsection{Data hyper-cleaning}
In this section, we test PBGD in the data hyper-cleaning task \citep{franceschi2017forward,shaban2019truncated}. In this task, one is given a set of polluted training data {\color{black} $\mathcal{D}_{\rm tr}=\{d_{\rm tr}^i,l_{\rm tr}^i\}_{i=1}^N$ where $d_{\rm tr}^i$ is the data and $l_{\rm tr}^i$ is the label}, along with a set of clean validation {\color{black} $\mathcal{D}_{\rm val}=\{d_{\rm val}^i,l_{\rm val}^i\}_{i=1}^M$} and clean test data. The goal is to train a data cleaner that assigns smaller weights to the polluted data to improve the generalization in unseen test data. {\color{black} Denote the data cleaner's parameter as $x$, the model parameter as $y$. Then the data hyper-cleaning task can be formulated as a bilevel optimization problem:
\begin{align}
    &\min_{x,y} -\frac{1}{M}\sum_{i=1}^M\log \mathbb{P}(l_{\rm val}^i ~| d_{\rm val}^i; y ) \nonumber\\
    &{\rm s.t.}~\|x\|\leq B,~y\in\argmin_{y\in\reals^{d_y}}-\frac{1}{N}\sum_{i=1}^N \omega^i(x)\log \mathbb{P}(l_{\rm tr}^i ~| d_{\rm tr}^i; y)+\frac{\eta}{2}\|y\|^2
\end{align}
where $\eta$ is the regularization constant; $B$ is a constant and the constraint $\|x\|\leq B$ avoids exploding weight parameters; $\mathbb{P}(l ~|d; y)$ is the probability output of class $l$ under input $d$ and the model parameter $y$; $\omega^i(x)$ is the weight for $i$-th training data under parameter $x$. In this experiment, we use sigmoid parameterizaiton for $\omega$, i.e., we have $x=(x_1,...,x_i,...,x_N)\in\reals^N$ and $\omega^i(x)=\frac{1}{1+\exp(-x_i)}$ for any $i$. This bilevel optimization problem is non-convex with a constraint for the upper-level variable.}

 \begin{table*}[bt]
 \small
 \centering
\begin{tabular}{ |p{2cm}||p{2.2cm}|p{2cm}|p{2.2cm}|p{2cm}| }
 \hline
 \multirow{2}{*}{Method} & \multicolumn{2}{|c|}{Linear model} & \multicolumn{2}{|c|}{2-layer MLP} \\
  \cline{2-5}
  & Test accuracy & F1 score & Test accuracy & F1 score \\
 \hline
 RHG & $87.64 \pm 0.19$ & $89.71 \pm 0.25$ & $87.50\pm 0.23$ & $89.41\pm 0.21$ \\
 \hline
 T-RHG & $87.63 \pm 0.19$ & $89.04 \pm 0.24$ & $87.48 \pm 0.22$ & $89.20 \pm 0.21$ \\
 \hline
 BOME & $87.09\pm 0.14$ & $89.83\pm 0.18$ & $87.42\pm 0.16$ & $89.26\pm 0.17$ \\
 \hline
 \textbf{G-PBGD} & $90.09\pm 0.12$ & $90.82\pm 0.19$ & $92.17\pm 0.09$ & $90.73\pm 0.27$\\
 \hline
 IAPTT-GM & $\textbf{90.44}\pm 0.14$  & $\textbf{91.89}\pm 0.15$ & $91.72\pm 0.11$ & $91.82\pm 0.19$ \\
 \hline
 \textbf{V-PBGD} & $\textbf{90.48}\pm 0.13$ & $\textbf{91.99}\pm 0.14$ & $\textbf{94.58} \pm 0.08$ & $\textbf{93.16}\pm 0.15$\\
 \hline
\end{tabular}
\vspace{-0.1cm}
\caption{Comparison of the solution quality among different algorithms. The results are averaged over $20$ runs and $\pm$ is followed by an estimated margin of error under $95\%$ confidence. F1 score measures the quality of the data cleaner \citep{franceschi2017forward}. 
% Test acc. resports the test accuracy achieved by $y$ during training.
% \vspace{0.1cm}
}
\label{tab:name1}\vspace{0.2cm}
\end{table*}

\begin{table*}[t!]
 \small
 \centering
\begin{tabular}{ |p{4.2cm}||p{1.cm}|p{1.2cm}|p{1.cm}|p{1.6cm}|p{1.8cm}|p{1.6cm}| }
 \hline
   & RHG & T-RHG &  BOME &  \textbf{G-PBGD} &IAPTT-GM & \textbf{V-PBGD} \\
  \hline
  GPU memory (MB) linear & 1369 & 1367  & 1149 &1149 &  $1237$ & $1149$ \\
  \hline
  GPU memory (MB) MLP & 7997 & 7757 & 1201 &  1235 & 2613  &$1199$ \\
  \hline
 Runtime (sec.) linear& 73.21 & 32.28 & 5.92 & 7.72 & 693.65 &9.12\\
 \hline
 Runtime (sec.) MLP& 94.78 & 54.96  & 39.78 & 185.08 & 1310.63 &207.53\\
 \hline
\end{tabular}
\caption{Comparison of GPU memory and the runtime to reach the highest test accuracy over 20 runs.
\vspace{-0.1cm}}
\label{tab:name2}
% \vspace{-0.1cm}
\end{table*}

We evaluate the performance of our algorithm in terms of speed, memory usage and solution quality in comparison with several competitive baseline algorithms including the IAPTT-GM \citep{liu2021towards}, BOME \citep{ye2022bome}, RHG \citep{franceschi2017forward} and T-RHG \citep{shaban2019truncated}. In addition to V-PBGD, we also test G-PBGD which is a special case of PBGD with the lower-level gradient norm penalty $p(x,y)=\|\nabla_y g(x,y)\|^2$. 
The method proposed by \citep{mehra2021penalty} can be viewed as G-PBGD.
Adopting the settings in \citep{franceschi2017forward,liu2021towards,shaban2019truncated}, we randomly split the MNIST data-set into a training data set of size $5000$, a validation set of size $5000$ and a test set of size $10000$; and pollute $50\%$ of the training data with uniformly drawn labels. Then we run the algorithms with a linear model and an MLP network.

We report the solution quality in Table \ref{tab:name1}. It can be observed that both PBGD algorithms achieve competitive performance, and V-PBGD achieves the best performance among the baselines. 
We also evaluate PBGD in terms of convergence speed and memory usage, which is reported in Table \ref{tab:name2}. It can be observed that PBGD does not have a steep increase in memory consumption or runtime as compared to the ITD baselines, indicating PBGD is potentially more scalable.

{\color{black}To this end, we have conducted experiments on bilevel optimization problems with an unconstrained lower-level variable. It is also of interest to perform experiments on bilevel problems with constraints on $y$; for example, we could consider $y$ lying within a closed interval in the verification experiments. Additionally, we can explore Stackelberg games where $y$ represents a tabular policy in reinforcement learning (see e.g., \citep{shen2024principled}), making the constraint set the probability simplex.}
 
\section{Conclusions}
In this work, we studied a challenging class of bilevel optimization problems, with possibly nonconvex and constrained lower-level problems, through the lens of the penalty method. We proved that the solutions of the penalized problem approximately solve the original bilevel problem under certain generic conditions verifiable with commonly made assumptions. To solve the penalized problem, we proposed the penalty-based bilevel GD method and established its finite-time convergence under unconstrained and constrained lower-level problems. For some of these settings, we further proposed the fully first-order and the stochastic versions of the penalty-based bilevel GD method. Experiments verified the effectiveness of the proposed algorithm.

\bibliography{bob}
\bibliographystyle{spbasic} 

% \clearpage
\appendix
 
\section{Omitted Proof in Section \ref{sec:generic}}

\subsection{Proof of Theorem \ref{the:BPgamr_epsilon_local_relax}} \label{app.subsec.a2}
In this section, we give the proof of a stronger version of Theorem \ref{the:BPgamr_epsilon_local_relax} in the sense of weaker assumptions.

We first define a new class of functions as follows.
\begin{definition}[Restricted $\alpha$-sublinearity]\label{def:onepointconvex}
Let $\Xc\subseteq \reals^{d}$. We say a function $\ell:\Xc\mapsto \reals$ is restricted  $\alpha$-sublinear on $x\in \Xc$ if there exists $\alpha\in[0,1]$ and $x^*\in \Xc$ which is the projection of $x$ onto the minimum point set of $\ell$ such that the following inequality holds.
$$\ell\big((1-\alpha) x + \alpha x^*\big) \leq (1-\alpha) \ell(x) + \alpha \ell(x^*) .$$
\end{definition}
Suppose $\ell$ is a continuous convex or more generally a star-convex function \citep[Definition 1]{nesterov2006cubic} defined on a closed convex set $\Xc$ and $\ell$ has a non-empty minimum point set, then $\ell$ is restricted $\alpha$-sublinear for any $\alpha\in[0,1]$ on every $x\in\Xc$.

Now we are ready to give the stronger version of Theorem \ref{the:BPgamr_epsilon_local_relax}.
\begin{theorem}[Stronger version of Theorem \ref{the:BPgamr_epsilon_local_relax}]\label{the:stronger generic 2}
Assume $p(x,\cdot)$ is continuous given any $x \in \Cc$ and $p(x,y)$ is $\rho$-squared-distance-bound function.
Given $\gamma>0$, let $(x_\gamma,y_\gamma)$ be a local solution of $\mathcal{BP}_{\gamma p}$ on $\Nc((x_\gamma,y_\gamma),r)$.
Assume $f(x_\gamma,\cdot)$ is $L$-Lipschitz-continuous on $\Nc(y_\gamma,r)$. 

Assume either one of the following is true:
\begin{enumerate}[label=(\roman*)]
 \item There exists $\Bar{y}\in\Nc(y_\gamma,r)$ such that $\Bar{y}\in \Uc(x_\gamma)$ and $p(x_\gamma,\Bar{y})\leq \epsilon$ for some $\epsilon\geq 0$. Define $\Bar{\epsilon}_\gamma=\frac{L^2 \rho}{\gamma^2}\!+\!2\epsilon$.
    \item The set $\Uc(x_\gamma)$ is convex and $p(x_\gamma,\cdot)$ is restricted $\alpha$-sublinear on $y_\gamma$ with some $\alpha \in(0, \min\{\frac{r}{d_{\Sc(x_\gamma)}(y_\gamma)},1\}]$. Define $\Bar{\epsilon}_\gamma=\frac{L^2 \rho}{\gamma^2}$.
\end{enumerate}
Then $(x_\gamma,y_\gamma)$ is a local solution of the following approximate problem of $\mathcal{BP}$ with $0\leq\epsilon_\gamma \leq \Bar{\epsilon}_\gamma$.
\begin{align}\label{bBP_epsilon_local_}
    \min_{x,y}f(x,y)~~~{\rm s.t.}~~~&x\in \Cc,~y\in\Uc(x)\\
    &p(x,y)\leq \epsilon_\gamma. \nonumber
\end{align}
\end{theorem}
The above theorem is stronger than Theorem \ref{the:BPgamr_epsilon_local_relax} in the sense that the condition (ii) in   above theorem is weaker than (ii) in Theorem \ref{the:BPgamr_epsilon_local_relax} since the continuity and convexity of $p(x_\gamma,\cdot)$ implies the restricted $\alpha$-sublinearity of $p(x_\gamma,\cdot)$ on $y_\gamma$ with any $\alpha \in [0,1]$.
\begin{proof}
We prove the theorem for two cases separately. 
\textcolor{black}{For each case, we will first prove $(x_\gamma, y_\gamma)$ is feasible for the problem $\mathcal{BP}_{\bar{\epsilon}_\gamma}$ by deriving the upper bound for $p(x_\gamma,y_\gamma)$. Under the feasibility of $(x_\gamma, y_\gamma)$, we can immediately show that $(x_\gamma, y_\gamma)$ is a local solution by using the condition that it solves $\mathcal{BP}_{\gamma p}$ locally.}

\textbf{Proof of Case (i).} Assume the conditions in Case (i) are true.
For  $\delta \geq 0$, define 
\begin{align}
    \Sc_\delta (x) \coloneqq \{y\in \Uc(x):p(x,y)\leq \delta\},~x\in \Cc.\nonumber
\end{align}
Since $p(x,y)=0$ if and only if (iff) $y\in\Sc(x)=\arg\min_{y\in\Uc(x)}g(x,y)$, it follows that $\Sc(x)=\{y\in\Uc(x):p(x,y)=0\}$.
Then $\Sc_\delta(x) \supseteq \Sc(x)$, and thus $\Sc_\delta(x) \neq \emptyset$. Moreover, $\Sc_\delta(x)$ is closed by continuity of $p(x,\cdot)$ and closeness of $\Uc(x)$ for $x \in \Cc$.

Since $(x_\gamma,y_\gamma)$ is a local solution of $\mathcal{BP}_{\gamma p}$ on $\Nc((x_\gamma,y_\gamma),r)$, it holds for any $(x,y)\in\Nc((x_\gamma,y_\gamma),r)$ that is feasible for $\mathcal{BP}_{\gamma p}$ that
\begin{align}\label{eq:idk33}
    f(x_\gamma,y_\gamma)+\gamma p(x_\gamma,y_\gamma) \leq f(x,y)+\gamma p(x,y).
\end{align}
Since $\Sc_\epsilon(x_\gamma)$ is closed and non-empty, we can find $y_x \in \arg\min_{y'\in\Sc_\epsilon(x_\gamma)} \|y'-y_\gamma\|$.
Since $\Bar{y}\in\Nc(y_\gamma,r) \cap \Sc_\epsilon(x_\gamma)$, we have
 $\norm{y_x-y_\gamma}\leq \norm{\Bar{y}-y_\gamma}\leq r$. This indicates $y_x \in \Nc(y_\gamma,r)$ and $(x_\gamma,y_x) \in \Nc((x_\gamma,y_\gamma),r)$.
Moreover, since $y_x \in \Uc(x_\gamma)$, $(x_\gamma,y_x)$ is feasible for $\mathcal{BP}_{\gamma p}$.  This allows to choose $(x,y)=(x_\gamma,y_x)$ in \eqref{eq:idk33}, leading to
$$f(x_\gamma,y_\gamma)+\gamma p(x_\gamma,y_\gamma) \leq f(x_\gamma,y_x) + \gamma \epsilon\quad \text{since }y_x \in \Sc_\epsilon(x_\gamma). $$

By Lipschitz continuity of $f(x_\gamma,\cdot)$ on $\Nc(y_\gamma,r)$, we further have
\begin{align}\label{eq:idk921}
    \gamma p(x_\gamma,y_\gamma) -L\|y_x-y_\gamma\| -\gamma \epsilon &\leq 0.
\end{align}
Since $\Sc_\epsilon (x_\gamma) \supseteq \Sc(x_\gamma)$, we have $\|y_x-y_\gamma\|=d_{\Sc_\epsilon(x_\gamma)}(y_\gamma) \leq d_{\Sc(x_\gamma)}(y_\gamma) \leq  \sqrt{\rho p(x_\gamma,y_\gamma)}$. 

Plugging this into \eqref{eq:idk921} yields
\begin{align}
    \gamma p(x_\gamma,y_\gamma) -L\sqrt{\rho p(x_\gamma,y_\gamma)} -\gamma \epsilon &\leq 0 \nonumber
\end{align}
which implies $p(x_\gamma,y_\gamma) \leq \Bar{\epsilon}_\gamma=\frac{L^2 \rho}{\gamma^2}+2\epsilon$.
Let $\epsilon_\gamma = p(x_\gamma,y_\gamma)$, then $\epsilon_\gamma \leq \Bar{\epsilon}_\gamma$ and $(x_\gamma,y_\gamma)$ is feasible for problem \eqref{bBP_epsilon_local_}. By \eqref{eq:idk33}, it holds for any $(x,y)\in\Nc((x_\gamma,y_\gamma),r)$ that are feasible for problem \eqref{bBP_epsilon_local_} that
\begin{align}
    f(x_\gamma,y_\gamma) - f(x,y) \leq \gamma (p(x,y)-\epsilon_\gamma)\leq 0.\nonumber
\end{align}
This and the fact that $(x_\gamma,y_\gamma)$ is feasible for \eqref{bBP_epsilon_local_} imply $(x_\gamma,y_\gamma)$ is a local solution of \eqref{bBP_epsilon_local_}.

\textbf{Proof of Case (ii).} Assume the conditions in Case (ii) are true. 
Since $\Sc(x_\gamma)$ is closed and non-empty, we can find $y_x$ such that $y_x\in\arg\min_{y\in\Sc(x_\gamma)}\|y-y_\gamma\|$. Let $\Bar{y}=(1-\alpha) y_\gamma+\alpha y_x$. Since $0<\alpha\leq \min\{r/\|y_\gamma-y_x\|,1\}$, we know $\Bar{y}\in\Nc(y_\gamma,r)$ and $(x_\gamma,\Bar{y})\in\Nc((x_\gamma,y_\gamma),r)$. Moreover, since $\Uc(x_\gamma)$ is convex, we have $\Bar{y}\in \Uc(x_\gamma)$ and $(x_\gamma,\Bar{y})$ is feasible for $\mathcal{BP}_{\gamma p}$.

Since $(x_\gamma,y_\gamma)$ is a local solution of $\mathcal{BP}_{\gamma p}$ on $\Nc((x_\gamma,y_\gamma),r)$, we have
\begin{align}\label{idkkk1}
    f(x_\gamma,y_\gamma)+\gamma p(x_\gamma,y_\gamma) &\leq f(x_\gamma,\Bar{y})+\gamma p(x_\gamma,\Bar{y}).
\end{align}
Since $p(x_\gamma,y)\geq0$ and $p(x_\gamma,y)=0$ iff $d_{\Sc(x_\gamma)}(y)=0$, we know the minimum point set of $p(x_\gamma,\cdot)$ is $\Sc(x_\gamma)$. Then by the restricted $\alpha$-sublinearity of $p(x_\gamma,\cdot)$ on $y_\gamma$, we have
$$p(x_\gamma,\Bar{y}) \leq \alpha p(x_\gamma,y_x) + (1-\alpha) p(x_\gamma,y_\gamma)=(1-\alpha) p(x_\gamma,y_\gamma).$$
Substituting the above inequality into \eqref{idkkk1} yields
\begin{align*}
     f(x_\gamma,y_\gamma)+\gamma p(x_\gamma,y_\gamma)
    &\leq  f(x_\gamma,\Bar{y})+\gamma  (1-\alpha) p(x_\gamma,y_\gamma).
\end{align*}
Re-arranging the above inequality and using the Lipschitz continuity of $f(x_\gamma,\cdot)$ on $\Nc(y_\gamma,r)$ yield
\begin{align}\label{eq:them7.ii}
    \gamma \alpha p(x_\gamma,y_\gamma) \leq  L \alpha d_{\Sc(x_\gamma)}(y_\gamma)\Rightarrow \gamma \alpha p(x_\gamma,y_\gamma)\leq L \alpha  \sqrt{\rho p(x_\gamma,y_\gamma)}
\end{align}
which implies $p(x_\gamma,y_\gamma) \leq \Bar{\epsilon}_\gamma=\frac{L^2 \rho}{\gamma^2}$.
Let $\epsilon_\gamma = p(x_\gamma,y_\gamma)$, then $\epsilon_\gamma \leq \Bar{\epsilon}_\gamma$ and $(x_\gamma,y_\gamma)$ is feasible for problem \eqref{bBP_epsilon_local_}.
Since $(x_\gamma,y_\gamma)$ is a local solution of $\mathcal{BP}_{\gamma p}$ on $\Nc((x_\gamma,y_\gamma),r)$, it holds for any $(x,y)\in\Nc((x_\gamma,y_\gamma),r)$ that is feasible for $\mathcal{BP}_{\gamma p}$ that
\begin{align}
    f(x_\gamma,y_\gamma)+\gamma p(x_\gamma,y_\gamma) \leq f(x,y)+\gamma p(x,y).\nonumber
\end{align}
Based on the above inequality, it holds for any $(x,y)\in\Nc((x_\gamma,y_\gamma),r)$ that are feasible for \eqref{bBP_epsilon_local_} that
\begin{align}
    f(x_\gamma,y_\gamma) - f(x,y) \leq \gamma (p(x,y)-\epsilon_\gamma)\leq 0.\nonumber
\end{align}
This and the fact that $(x_\gamma,y_\gamma)$ is feasible for \eqref{bBP_epsilon_local_} imply $(x_\gamma,y_\gamma)$ is a local solution of \eqref{bBP_epsilon_local_}. \qed
\end{proof}

\section{Omitted Proof in Section \ref{sec:NP}}
\subsection{Proof of Lemma \ref{lem:pl penalty}}\label{app.subsec.b1}
% \begin{proof}
\textbf{Proof of Case (i).} Assume (i) in this lemma holds.
By the definition of $v(x)$, it is clear that $g(x,y)-v(x) \geq 0$ for any $x\in \Cc$ and $y\in\reals^{d_y}$.
Since $\Sc(x)$ is closed, $y\in\Sc(x)$ iff $d_{\Sc(x)}(y)=0$. Then by the definition of $\Sc(x)$, it holds for any $x\in \Cc$ and $y\in\reals^{d_y}$ that
\begin{align}
    g(x,y)-v(x)=0~\text{iff}~y\in \Sc(x) \Rightarrow g(x,y)-v(x)=0~\text{iff}~d_{S(x)}(y)=0. \nonumber
\end{align}
It then suffices to check whether $g(x,y)-v(x)$ is an upper-bound of $d_{S(x)}(y)$. By $\frac{1}{\mu}$-PL condition of $g(x,\cdot)$ and \citep[Theorem 2]{karimi2016linear}, $g(x,\cdot)$ satisfies the $\frac{1}{\mu}$-quadratic-growth condition, and thus for any $x\in \Cc$ and $y\in\reals^{d_y}$, it holds that
\begin{align}\label{eq:qg}
    g(x,y)-v(x) \geq \frac{1}{\mu} d_{S(x)}^2(y).
\end{align}
This completes the proof.

\textbf{Proof of Case (ii).} Assume (ii) in this lemma holds.
We consider when $g(x,\cdot)$ satisfies PL condition given any $x\in \Cc$. By the PL inequality, it is clear that $\|\nabla_y g(x,y)\|^2=0$ is equivalent to $g(x,y)=\min_{y \in \reals^{d_y}} g(x,y)$ given any $x \in \Cc$, thus $\|\nabla_y g(x,y)\|^2=0$ iff $d_{\Sc(x)}(y)=0$ for any $x \in \Cc$.

By $\frac{1}{\sqrt{\mu}}$-PL condition of $g(x,\cdot)$, we have $\|\nabla_y g(x,y)\|^2 \geq \frac{1}{\sqrt{\mu}} (g(x,y)-v(x))$.
By \eqref{eq:qg}, we have $g(x,y)-v(x) \geq \frac{1}{\sqrt{\mu}} d_{S(x)}^2(y)$. Thus it holds that $$ \|\nabla_y g(x,y)\|^2 \geq \frac{1}{\mu} d_{S(x)}^2(y),$$
which completes the proof.
% \end{proof}
\subsection{Support lemma for Theorem \ref{the:nonconvex xy convergence}}
\begin{lemma}\label{lem:support 1}
    Let $\Zc\subseteq \reals^d$ be a closed convex set. Given any $z\in\Zc$, $q\in\reals^d$ and $\alpha>0$, it holds that
    $$\Proj_{\Zc}\big( z-\alpha q\big)=\arg\min_{z'\in\Zc}\ip{q}{z'}+\frac{1}{2\alpha}\|z-z'\|^2.$$
\end{lemma}
\begin{proof}
Given $z\in\reals^d$, define $z^*=\arg\min_{z' \in\reals^d} E(z')$ where
\begin{equation}
    E(z')\coloneqq \ip{q}{z'-z}+\frac{1}{2\alpha}\|z'-z\|^2.
\end{equation}
% $E(z)\coloneqq \ip{\nabla f(z_k)}{z-z_k}+\frac{\rho}{2}\|z-z_k\|^2$ 
By the optimality condition, it follows $z^*=z - \alpha q$.
% \begin{align}
%     z^*=z_k - \frac{1}{\rho}\nabla f(z_k).
% \end{align}
% \SP{The first term follows from the definition of $E(z)$ and the second one follows from the definition of $z$ and (9). Make the comment explicit. }
For any $z'\in\reals^{d}$, it follows that
\begin{align}\label{eq:idk279}
    E(z')-E(z^*)
    &=\ip{q}{z'-z}+\frac{1}{2\alpha}\|z'-z\|^2-\ip{q}{-\alpha q}-\frac{\alpha}{2 }\norm{q}^2\nonumber\\
    &= \frac{1}{2\alpha}\|z'-z\|^2 + \ip{q}{z'-z}+ \frac{\alpha}{2}\|q\|^2 \nonumber\\
    &= \frac{1}{2\alpha}\|z'-(z - \alpha q)\|^2 .
\end{align}
Then we have
\begin{align}
    \arg\min_{z'\in\Zc}\ip{q}{z'}+\frac{1}{2\alpha}\|z-z'\|^2&=\arg\min_{z'\in \Zc}E(z')\nonumber\\
    % &=\arg\min_{z'\in \Zc}E(z')-E(z^*) \nonumber\\
    & =\arg\min_{z'\in\Zc}\|z'-(z - \alpha q)\|^2~~\text{ by \eqref{eq:idk279}}\nonumber\\
    &=\Proj_{\Zc}\big(z - \alpha q \big).
\end{align}
This proves the result. \qed
\end{proof}

\subsection{Proof of Proposition \ref{pro:SP}}\label{app.subsec.b3} 
\begin{proof}
\textbf{Proof of Case (a).}
If we choose the penalty function as $p(x,y)=\|\nabla_y g(x,y)\|^2$ in  \eqref{eq:gry g}, the $\delta$-stationary point condition of the penalized problem $\mathcal{UP}_{\gamma p}$ is 
\begin{align*}
&\|\nabla_y f(x,y)+\gamma\nabla_{yy}g(x,y)\nabla_y g(x,y)\|\leq\delta\\
&\|\nabla_x f(x,y)+\gamma\nabla_{xy}g(x,y)\nabla_y g(x,y)\|\leq\delta.
\end{align*}
By choosing $w=\gamma\nabla_y g(x,y)$, the stationary conditions of $\mathcal{UP}$ in \eqref{kkt1-x-m} and \eqref{kkt1-y-m} hold. When $\gamma=\Omega(\delta^{-0.5})$ which is large enough, the feasibility condition in \eqref{kkt1-low-m} also holds because $\|\nabla_y f(x,y)\|\leq L$ from Assumption \ref{asp:Lipschitz continuity} and $\|\nabla_{yy} g(x,y)\|\leq L_g$ from Condition (i) in Lemma \ref{lem:pl penalty}. Therefore, for Case (a), when $p(x,y)=\|\nabla_y g(x,y)\|^2$, the stationary points of the penalized problem $\mathcal{UP}_{\gamma p}$ imply the the $\epsilon$-stationary points of $\mathcal{UP}$ without additional assumptions. 
\vspace{0.2cm}

\noindent\textbf{Proof of Case (b).}
Instead, if we choose the penalty function as $p(x,y)=g(x,y)-v(x)$ in \eqref{eq:g-v}, the $\delta$-stationary point condition of the penalized problem $\mathcal{UP}_{\gamma p}$ is 
\begin{subequations}\label{sub:stationary_penalty}
    \begin{align}
&\|\nabla_y f(x,y)+\gamma\nabla_y g(x,y)\|\leq\delta\label{eq1}\\
&\|\nabla_x f(x,y)+\gamma(\nabla_x g(x,y)-\nabla_x g(x,y^*(x)))\|\leq\delta\label{eq2}
\end{align}
\end{subequations}
with $\forall y^*(x)\in\argmin_y g(x,y)$. 
First, for a large enough $\gamma=\Omega(\delta^{-0.5})$, \eqref{eq1} and Assumption \ref{asp:Lipschitz continuity} give 
\begin{align*}
\gamma\|\nabla_y g(x,y)\|\leq \|\nabla_y f(x,y)+\gamma\nabla_y g(x,y)\|+\|\nabla_y f(x,y)\|\leq L+\delta. 
\end{align*}
Then dividing both sides by $\gamma=\Omega(\delta^{-0.5})$ yields the feasibility condition in \eqref{kkt1-low-m} through 
\begin{align*}
\|\nabla_y g(x,y)\|\leq L/\gamma+\delta/\gamma = {\cal O}(\delta^{0.5})\leq{\cal O}(\delta).
\end{align*}
Second, by condition (i) in Lemma \ref{lem:pl penalty} and its equivalent error bound condition, $\|\nabla_y g(x,y)\|\leq\delta$ gives  
\begin{equation}\label{eq.pf.eb-Taylor}
\|y-y^*(x)\|\leq \delta\quad {\rm for~~ some }\quad y^*(x)\in\argmin_y g(x,y).
\end{equation}
Therefore, using Taylor expansion and the bound \eqref{eq.pf.eb-Taylor}, we have 
\begin{align}\label{eq.pf.Taylor1}
\nabla_y g(x,y)=\nabla_y g(x,y)-\nabla_y g(x,y^*(x))&=\nabla_{yy} g(x,y^*(x))(y-y^*(x))+{\cal O}(\|y-y^*(x)\|^2)\nonumber\\
&=\nabla_{yy} g(x,y^*(x))(y-y^*(x))+{\cal O}(\delta^2)
\end{align}
and likewise, we have
\begin{align}\label{eq.pf.Taylor2}
\nabla_x g(x,y)-\nabla_x g(x,y^*(x))&=\nabla_{xy} g(x,y^*(x))(y-y^*(x))+{\cal O}(\|y-y^*(x)\|^2)\nonumber\\
&=\nabla_{xy} g(x,y^*(x))(y-y^*(x))+{\cal O}(\delta^2).
\end{align}
% where $\nabla=[\nabla_{x};\nabla_{y}]$ and $\nabla_{y}\nabla=[\nabla_{xy};\nabla_{yy}]$ denote the concatenated derivative matrix operator. 
As a result, combining the two equations \eqref{eq.pf.Taylor1}-\eqref{eq.pf.Taylor2} with \eqref{sub:stationary_penalty}, we have 
\begin{subequations}\label{sub:stationary_penalty_re}
\begin{align}
&\|\nabla_y f(x,y)+\gamma\nabla_{yy} g(x,y^*(x))(y-y^*(x))\|\leq\delta+{\cal O}(\gamma\delta^2)={\cal O}(\delta)\label{eq11}\\
&\|\nabla_x f(x,y)+\gamma\nabla_{xy} g(x,y^*(x))(y-y^*(x))\|\leq\delta+{\cal O}(\gamma\delta^2)={\cal O}(\delta)\label{eq22}
\end{align}
\end{subequations}
where the last equality is because $\gamma=\Omega(\delta^{-0.5})$ and $\delta\leq 1$. If we choose $w=\gamma(y-y^*(x))$ which is bounded since $\|w\|\leq \gamma\|y-y^*(x)\|\leq\delta^{0.5}\leq 1$, then \eqref{sub:stationary_penalty_re} only differs from \eqref{kkt1-x-m} and \eqref{kkt1-y-m} by the evaluation point $y^*(x)$ rather than $y$. 

We next prove the stationary condition in \eqref{kkt1-x-m} as the stationary condition \eqref{kkt1-y-m} can be obtained similarly. 
For the LHS of \eqref{kkt1-x-m}, using the Cauchy-Schwartz inequality, we have 
\begin{align*}
\|\nabla_x f(x,y)+\nabla_{xy}g(x,y)w\|&\leq\|\nabla_x f(x,y)+\nabla_{xy}g(x,y^*(x))w\|+\|w\|\|\nabla_{xy}g(x,y)-\nabla_{xy}g(x,y^*(x))\|\\
&\leq \|\nabla_x f(x,y)+\gamma\nabla_{xy} g(x,y^*(x))(y-y^*(x))\|+L_{g,2}\|w\|\|y-y^*(x)\| \leq{\cal O}(\delta)
\end{align*}
where the second inequality uses the smoothness assumption of $\nabla_y g(x,y)$ and the choice of $w$ as $w=\gamma(y-y^*(x))$, and the third inequality uses $\|w\|\leq 1, \|y-y^*(x)\|={\cal O}(\delta)$ and \eqref{eq22}. Therefore, we have shown the stationary condition in \eqref{kkt1-x-m}. Likewise, the stationary condition in  \eqref{kkt1-y-m} also holds
\begin{subequations}\label{sub:stationary_penalty_re2}
\begin{align}
&\|\nabla_x f(x,y)+\nabla_{xy} g(x,y)w\|\leq{\cal O}(\delta)\\
&\|\nabla_y f(x,y)+\nabla_{yy} g(x,y)w\|\leq{\cal O}(\delta)
\end{align}
\end{subequations}
which completes the proof. 
\end{proof}
 
\section{Proof of Proposition \ref{pro:generalized danskin}}\label{app.subsec.c2}
In this section, we prove a more general version of Proposition \ref{pro:generalized danskin}. 
To introduce this general version, we first prove the following lemma on the Lipschitz-continuity of the solution set $\Sc(x)$.

\begin{lemma}[Lipschitz-continuity of $\Sc(x)$]\label{lem:sx lipschitz}
Assume $g(x,y)$ is $L_g$-Lipschitz-smooth with some $L_g\!>\!0$. Assume either one of the following is true:
\begin{enumerate}[label=(\alph*)]
    \item Condition (ii) in Assumption \ref{cond:cp} holds. Let $L_S=L_g \Bar{\mu}$.
    \item Conditions (i) and (iii) in Assumption \ref{cond:cp} hold. Let $L_S=L_g(\mu\!+\!1)(L_g\!+\!1)$.
\end{enumerate}
Then given any $x_1,x_2\in\Cc$, for any $y_1\in\Sc(x_1)$ there exists $y_2\in\Sc(x_2)$ such that
$$\|y_1-y_2\| \leq L_S\|x_1-x_2\|.$$
\end{lemma}
\begin{proof}
\textbf{Proof of Case (a).} 
Given $x$, define the projected gradient of $g(x,\cdot)$ at point $y$ as
$$G(y;x)=\frac{1}{\beta}\big(y-\Proj_{\Uc}\big(y-\beta\nabla_y g(x,y)\big)\big).$$
By the assumption, the proximal-error-bound inequality holds, that is
$$\Bar{\mu}\|G(y;x)\|^2 \geq d_{\Sc(x)}^2(y),~\forall y\in\Uc\text{ and }x\in\Cc.$$

Therefore, given   $x_1,x_2\in\Cc$, we have for any $y_1\in\Sc(x_1)$ there exists $y_2\in\Sc(x_2)$ such that
\begin{align}
    \|y_1-y_2\| &\leq \Bar{\mu}\|G(y_1;x_2)-G(y_1;x_1)\|^2 \quad\text{since}~~ G(y_1;x_1)=0 \nonumber\\
    &= \frac{\Bar{\mu}}{\beta}\big\|\Proj_{\Uc}\big(y_1-\beta\nabla_y g(x_2,y_1)\big)-\Proj_{\Uc}\big(y_1-\beta\nabla_y g(x_1,y_1)\big)\big\|\nonumber\\
    &\leq \Bar{\mu}\|\nabla g(x_1,y_1) - \nabla g(x_2,y_1)\| \leq L_g\Bar{\mu} \|x_1-x_2\|.
\end{align}
This completes the proof for Case (a).
    
\textbf{Proof of Case (b).} 
    By the $1/\mu$-quadratic-growth of $g(x,\cdot)$ and \citep[Corrolary 3.6]{drusvyatskiy2018error}, the proximal-error-bound inequality holds, that is
    $$(\mu+1)(L_g+1)\|G(y;x)\|^2 \geq d_{\Sc(x)}^2(y),~\forall y\in\Uc\text{ and } x\in\Cc$$
    where we set $\beta=1$ to simplify the constant.
    The result then directly follows from Case (a). \qed
\end{proof}
 
Define $g_2(x,y)=g(x,y)+g_1(y)$, where $g_1$ is convex and possibly non-smooth. Define $\Sc_2(x)=\arg\min_{y\in\reals^{d_y}}g_2(x,y)$. Next we prove the following more general version of Proposition \ref{pro:generalized danskin}.
\begin{proposition}[General version of Proposition \ref{pro:generalized danskin}]\label{pro:general generalized danskin}
    Assume there exists constant $L_g$ such that $g(x,y)$ is $L_g$-Lipschitz-smooth. Assume given any $x_1$ and $x_2$, for any $y_1\in\Sc_2(x_1)$ there exists $y_2\in\Sc_2(x_2)$ such that
$$\|y_1-y_2\| \leq L_S\|x_1-x_2\|.$$
    Then $v_2(x)\coloneqq\min_{y\in\reals^{d_y}} g_2(x,y)$ is differentiable with the gradient
\begin{align*}
    \nabla v_2(x)=\nabla_x g(x,y^*), ~~\forall y^*\in \Sc_2(x).
\end{align*}
Moreover, $v_2(x)$ is $L_v$-Lipschitz-smooth with $L_v \coloneqq L_g(1\!+\!L_S)$. 
\end{proposition}
Given any $x\in\Cc$, choose $g_1$ such that $g_1(y)=0,\forall y \in \Uc$ and $g_1(y)=\infty$ elsewhere gives $v_2(x)=\min_{y\in\reals^{d_y}} g(x,y)+g_1(y)=\min_{y\in\Uc} g(x,y)=v(x)$ and $\Sc_2(x)=\Sc(x)$. Then Proposition \ref{pro:generalized danskin} follows from Proposition \ref{pro:general generalized danskin} with Lemma \ref{lem:sx lipschitz}.

\begin{proof}[Proof of Proposition \ref{pro:general generalized danskin}]
\blue{We will proceed the proof by the following sketch. The major goal is to establish $v_2(x+r d)-v_2(x)=r\ip{\nabla_{x} g(x,y^*)}{d}+{\cal O}(r^2)$ for any radius $r\geq 0$ and direction $d\in\mathbb{R}^{d_x}$. We will first expand $v_2(x+r d)-v_2(x)$ by the smooth difference $g(x+rd,y^*(r))-g(x,y^*)$ and the convex difference $g_1(y^*(r))-g_1(y^*)$, and then leverage the smoothness and convexity property to bound those differences respectively. Finally, the desired outcome is obtained by the first-order optimality condition of the lower-level problem. }

For any $x$, we can choose any $y^*\in \Sc_2(x)$. Then by the assumption, for any $r>0$ and any unit direction $d$, one can find $y^*(r)\in \Sc_2(x+rd)$ such that
\begin{align*}
    \|y^*(r)-y^*\|\leq L_S\|rd\|=L_S r. 
\end{align*}
In this way, we can expand the difference of $v_2(x+rd)$ and $v_2(x)$ as
\begin{align}
    v_2(x+rd)-v_2(x)&=g_2(x+rd,y^*(r))-g_2(x,y^*)\nonumber\\
    &=\underbrace{g(x+rd,y^*(r))-g(x,y^*)}_{\eqref{smooth-1} \& \eqref{smooth-2}}+\underbrace{g_1(y^*(r))-g_1(y^*)}_{\eqref{convex}}\label{sum}
\end{align}
which will be bounded subsequently. 
First, according to the Lipschitz smoothness of $g$, we have
\begin{subequations}
\begin{align}
    &~~~~~g(x+rd,y^*(r))-g(x,y^*)\nonumber\\
    &\leq r\ip{\nabla_{x} g(x,y^*)}{d}+\ip{\nabla_y g(x,y^*)}{y^*(r)-y^*}+L_g(1+L_S^2)r^2\nonumber\\
    &\leq r\ip{\nabla_{x} g(x,y^*)}{d}+\ip{\nabla_y g(x+rd,y^*(r))}{y^*(r)-y^*}+L_g(1+L_S^2)r^2\nonumber\\
    &~~~~+\|\nabla_y g_2(x+rd,y^*(r))-\nabla_y g(x,y^*)\|\|y^*(r)-y^*\|\nonumber\\
    &\leq r\ip{\nabla_{x} g(x,y^*)}{d}+\ip{\nabla_y g(x+rd,y^*(r))}{y^*(r)-y^*}+L_g(1+L_S+2L_S^2)r^2\label{smooth-1}
    \end{align}
    and 
    \begin{align}
g(x+rd,y^*(r))-g(x,y^*)\geq r\ip{\nabla_{x} g(x,y^*)}{d}+\ip{\nabla_y g(x,y^*)}{y^*(r)-y^*}-L_g(1+L_S^2)r^2.\label{smooth-2}
    \end{align}
\end{subequations}
By the definition of the sub-gradient of the convex function $g_1$, we have for any $p_1\in\partial g_1(y)$ and $p_2\in\partial g_1(y^*(r))$, it holds that
\begin{align}\label{convex}
    \ip{p_1} {y^*(r)-y^*}\leq g_1(y^*(r))-g_1(y^*)\leq \ip{p_2}{y^*(r)-y^*}.
\end{align}
Moreover, the first order necessary optimality condition of $y^*$ and $y^*(r)$ yields
\begin{align*}
    0\in \nabla_y g(x,y^*)+\partial g_1(y^*) ~~~\text{ and }~~~0\in \nabla_y g(x+rd,y^*(r))+\partial g_1(y^*(r)). 
\end{align*}
As a result, there exists $p_1\in\partial g_1(y)$ and $p_2\in\partial g_1(y^*(r))$ such that 
\begin{align}\label{v12}
    0= \nabla_y g(x,y^*)+p_1, ~~~\text{ and }~~~0= \nabla_y g(x+rd,y^*(r))+p_2. 
\end{align}
Choosing $p_1, p_2$ satisfying \eqref{v12} in \eqref{convex}, then substituting \eqref{smooth-1}, \eqref{smooth-2} and \eqref{convex} into \eqref{sum} yields
\begin{align*}
    r\ip{\nabla_{x} g(x,y^*)}{d}-{\cal O}(r^2)\leq v_2(x+rd)-v_2(x)\leq r\ip{\nabla_{x} g(x,y^*)}{d}+{\cal O}(r^2). 
\end{align*}
With the above inequalities, the directional derivative $\nabla v_2(x; d)$ is then
\begin{align*}
    \nabla v_2(x; d)=\lim _{r \rightarrow 0^{+}} \frac{v_2(x+r d)-v_2(x)}{r}=\ip{\nabla_{x} g(x,y^*)}{d}.
\end{align*}
This holds for any $y^*\in\Sc_2(x)$ and $d$. We get $\nabla v_2(x)=\nabla_{x} g(x,y^*),~\forall y^*\in\Sc_2(x)$. 

Given any $x_1,x_2$, by the assumption, we can choose $y_1\in\Sc_2(x_1)$ and $y_2\in\Sc_2(x_2)$ such that
$\|y_1-y_2\| \leq L_S\|x_1-x_2\|.$
Then the Lipschitz-smoothness of $v_2(x)$ follows from
\begin{align*}
    \|\nabla v_2(x_1)-\nabla v_2(x_2)\|&=\|\nabla_x g(x_1,y_1)-\nabla_x g(x_2,y_2)\|\\
    &\leq L_g(\|x-x^\prime\|+\|y_1-y_2\|)\\
    &\leq L_g(1+L_S)\|x-x^\prime\|
\end{align*}
which completes the proof. \qed
\end{proof}

\section{Proof for Proposition \ref{exact-penalty-nonsmooth}}\label{app.pf.prop6}

Similar to the definition of a squared-distance-bound function in Definition \ref{def:SDB}, we introduce the definition of a distance-bound function next. 
\begin{definition}[Distance bound functions]\label{def:DBF}
A function $p:\reals^{d_x}\times \reals^{d_y}\mapsto \reals$ is a $\rho$-distance-bound if there exists $\rho>0$ such that for any $x\in \Cc,y\in \Uc(x)$, the following two conditions are met
\begin{subequations}
    \begin{align}
    &p(x,y)\geq 0,~\rho p(x,y) \geq d_{\Sc(x)}(y) \label{eq.dbf_a}\\
    &p(x,y)=0~\text{ if and only if }~d_{\Sc(x)}(y)=0.
\end{align}
\end{subequations}
\end{definition}

The following two theorems for the general bilevel problem $\mathcal{BP}$ and its penalty reformulation $\mathcal{BP}_{\gamma p}$ are used to prove Proposition \ref{exact-penalty-nonsmooth}.
\begin{theorem}\label{the:BP_gamr}
Assume $p(x,y)$ is an $\rho$-distance-bound function and $f(x,\cdot)$ is $L$-Lipschitz continuous on $\Uc(x)$ for any $x\in \Cc$. Any global solution of $\mathcal{BP}$ is a global solution of $\mathcal{BP}_{\gamma p}$ with any $\gamma \geq \gamma^*=L\rho$. Conversely, given $\epsilon_2\geq0$, let $(x_\gamma,y_\gamma)$ achieve $\epsilon_2$-global-minimum of $\mathcal{BP}_{\gamma p}$ with $\gamma > \gamma^*$. Then $(x_\gamma,y_\gamma)$ is the global solution of the following approximate problem of $\mathcal{BP}$ with  $0\leq\epsilon_\gamma \leq \epsilon_2/(\gamma\!-\!\gamma^*)$, given by
\begin{align}\label{bBP_epsilon_2}
    \min_{x,y}f(x,y)~~~{\rm s.t.}~~~&x\in \Cc,~y\in\Uc(x),\nonumber\\
    &p(x,y)\leq \epsilon_\gamma.
\end{align}
\end{theorem}
\begin{proof}
This proof mostly follows from that of Theorem \ref{the:BP_gamr_epsilon}. We will show the different steps here.
We have
\begin{align}\label{eq:idk222}
    f(x,y)+\gamma^* p(x,y)-f(x,y_x) &\geq -L d_{\Sc(x)}(y) + \gamma^* p(x,y) \nonumber\\
    &\geq -L d_{\Sc(x)}(y) + \frac{\gamma^*}{\rho} d_{\Sc(x)}(y) \nonumber\\
    &=0\quad\text{with }\gamma^*=L\rho.
\end{align}
Since $y_x \in \Sc(x)$ (thus $y_x\in \Uc(x)$) and $x\in \Cc$, $(x,y_x)$ is feasible for $\mathcal{BP}$. Let $f^*$ be the optimal objective value for $\mathcal{BP}$, we know $f(x,y_x) \geq f^*$. This along with \eqref{eq:idk222} indicates
\begin{align}\label{eq:global calmness 2}
    f(x,y)+\gamma^* p(x,y)-f^*\geq 0,~~\forall x\in \Cc,~y\in \Uc(x).
\end{align}
The rest of the proof follows from that of the first half of Theorem \ref{the:BP_gamr_epsilon} with \eqref{eq:global calmness 2} in place of \eqref{eq:global calmness 1}. \qed
\end{proof}

\begin{theorem}\label{the:stronger generic 4}
Assume $p(x,\cdot)$ is continuous given any $x \in \Cc$ and $p(x,y)$ is $\rho$-distance-bound function.
Given $\gamma>0$, let $(x_\gamma,y_\gamma)$ be a local solution of $\mathcal{BP}_{\gamma p}$ on $\Nc((x_\gamma,y_\gamma),r)$.
Assume $f(x_\gamma,\cdot)$ is $L$-Lipschitz-continuous on $\Nc(y_\gamma,r)$, the set $\Uc(x_\gamma)$ is convex, and $p(x_\gamma,\cdot)$ is convex with $\gamma>L\rho$.
Then $(x_\gamma,y_\gamma)$ is a local solution of $\mathcal{BP}$:
\begin{align}\label{bBP_epsilon_local_2}
    \min_{x,y}f(x,y)~\quad{\rm s.t.}\quad&x\in \Cc,~y\in\Uc(x)\\
    &p(x,y)=0. \nonumber
\end{align}
\end{theorem}
\begin{proof}
The proof is similar to that of Theorem \ref{the:stronger generic 2} (ii). We will show the different steps here.

Following the proof of Theorem \ref{the:stronger generic 2} (ii), in \eqref{eq:them7.ii}, we instead use \eqref{eq.dbf_a} and have for any $\alpha\in(0,1)$ that
\begin{align}\label{eq:idk898}
    \gamma \alpha p(x_\gamma,y_\gamma) \leq  L \alpha d_{\Sc(x_\gamma)}(y_\gamma)\Rightarrow \gamma \alpha p(x_\gamma,y_\gamma)\leq L \alpha  \rho p(x_\gamma,y_\gamma).
\end{align}
Assume $p(x_\gamma,y_\gamma)\neq 0$. Since we have chosen $\gamma > L\rho$, we will also have $\alpha p(x_\gamma,y_\gamma)>L \alpha  \rho p(x_\gamma,y_\gamma)$ which contradicts with \eqref{eq:idk898}. This indicates $p(x_\gamma,y_\gamma)= 0$.
Then one can set $\bar{\epsilon}_\gamma = 0$ in the proof immediately after \eqref{eq:them7.ii} and finally follow the same steps to obtain the result. \qed
\end{proof}

 By Lemma \ref{lem:pl penalty} and Definition \ref{def:DBF}, we can show that $\|\nabla_y g(x,y)\|$ is a $\sqrt{\mu}$-distance-bound function. Further notice that $\mathcal{UP}$ and $\mathcal{UP}_{\gamma p}$ are respectively the special cases of $\mathcal{BP}$ and $\mathcal{BP}_{\gamma p}$ with $\Uc(x)=\reals^{d_y}$, then Condition (i) of Proposition \ref{exact-penalty-nonsmooth} directly follows from Theorem \ref{the:BP_gamr} with $\epsilon_2=0$; and Condition (ii) of Proposition \ref{exact-penalty-nonsmooth} directly follows from Theorem \ref{the:stronger generic 4}. 
\end{document}